\documentclass{ieeetj}

\usepackage{cite}
\usepackage{amsmath,amssymb,amsfonts}
\usepackage{algorithmic}
\usepackage{graphicx,color}
\usepackage{textcomp}
\usepackage{xcolor}
\usepackage[colorlinks=true, citecolor=blue, linkcolor=blue, urlcolor=red]{hyperref}
\usepackage{algorithm,algorithmic}
\def\BibTeX{{\rm B\kern-.05em{\sc i\kern-.025em b}\kern-.08em
    T\kern-.1667em\lower.7ex\hbox{E}\kern-.125emX}}
\AtBeginDocument{\definecolor{tmlcncolor}{cmyk}{0.93,0.59,0.15,0.02}\definecolor{NavyBlue}{RGB}{0,86,125}}

\usepackage{todonotes} 
\usepackage{acronym}
\usepackage{siunitx}
\usepackage[capitalize]{cleveref}
\usepackage{placeins}
\usepackage{booktabs}
\usepackage{overpic}
\let\labelindent\relax
\usepackage{enumitem}
\usepackage{needspace}
\usepackage{subfigure}
\usepackage{bbm}

\acrodef{CNN}{Convolutional Neural Network}
\acrodef{CoM}{Center of Mass}
\acrodef{DoF}{Degree of Freedom}
\acrodef{FF}{Feed Forward}
\acrodef{IK}{Inverse Kinematics}
\acrodef{IMU}{Inertial Measurement Unit}
\acrodef{KF}{Kalman Filter}
\acrodef{LUT}{Look-Up Table}
\acrodef{MAE}{Mean Absolute Error}
\acrodef{ML}{Machine Learning}
\acrodef{MLP}{Multilayer Perceptron}
\acrodef{MPC}{Model Predictive Control}
\acrodef{NN}{Neural Network}
\acrodef{PPO}{Proximal Policy Optimization}
\acrodef{PRM}{Probabilistic Roadmaps}
\acrodef{RL}{Reinforcement Learning}
\acrodef{RRT}{Rapidly-exploring Random Trees}
\acrodef{RRT*}{Rapidly-exploring Random Trees - STAR}
\acrodef{ROS}{Robot Operating System}
\acrodef{SM}{State Machine}
\acrodef{TCN}{Temporal Convolution Network}
\acrodef{THO}{Task Hierarchical Optimization}
\acrodef{ZOH}{Zero-Order Hold}
\acrodef{DEM}{Discrete Element Method}

\def\OJlogo{\vspace{-4pt}}
\def\seclogo{\vspace{10pt}}

\def\authorrefmark#1{\ensuremath{^{\textbf{#1}}}}

\renewcommand\thesection{\Roman{section}}

\begin{document}

\markboth{Large Scale Robotic Material Handling}{Spinelli {et al.}}

\title{Large Scale Robotic Material Handling: Learning, Planning, and Control}

% Authors
\author{
    Filippo A. Spinelli\authorrefmark{1}\textsuperscript{\dag}, Yifan Zhai\authorrefmark{1}\textsuperscript{\dag}, Fang Nan\authorrefmark{1}, Pascal Egli\authorrefmark{1}, Julian Nubert\authorrefmark{1}, \\ Thilo Bleumer\authorrefmark{2}, Lukas Miller\authorrefmark{2}, Ferdinand Hofmann\authorrefmark{2}, and Marco Hutter\authorrefmark{1}

% Affiliations
    \affil{Robotic Systems Lab, ETH Zurich, Zurich, Switzerland}
    \affil{Liebherr-Hydraulikbagger GmbH, Kirchdorf an der Iller, Germany}
    
    % Funding and contributions
    \authornote{\textsuperscript{\dag} Filippo A. Spinelli and Yifan Zhai contributed equally to this work. \\
    This work is supported by the NCCR digital fabrication, and the Liebherr-Hydraulikbagger GmbH.}
    \corresp{Corresponding author: Filippo A. Spinelli (email: {\href{mailto:filippo.a.spinelli@gmail.com}{\textcolor{blue}{filippo.a.spinelli@gmail.com}}})
    }
}

\begin{abstract}
Bulk material handling involves the efficient and precise moving of large quantities of materials, a core operation in many industries, including cargo ship unloading, waste sorting, construction, and demolition. These repetitive, labor-intensive, and safety-critical operations are typically performed using large hydraulic material handlers equipped with underactuated grippers. 
In this work, we present a comprehensive framework for the autonomous execution of large-scale material handling tasks. The system integrates specialized modules for environment perception, pile attack point selection, path planning, and motion control. The main contributions of this work are two reinforcement learning-based modules: an attack point planner that selects optimal grasping locations on the material pile to maximize removal efficiency and minimize the number of scoops, and a robust trajectory following controller that addresses the precision and safety challenges associated with underactuated grippers in movement, while utilizing their free-swinging nature to release material through dynamic throwing.
We validate our framework through real-world experiments on a \SI{40}{\tonne} material handler in a representative worksite, focusing on two key tasks: high-throughput bulk pile management and high-precision truck loading. Comparative evaluations against human operators demonstrate the system’s effectiveness in terms of precision, repeatability, and operational safety. To the best of our knowledge, this is the first complete automation of material handling tasks on a full scale.
\end{abstract}


\maketitle

\section{Introduction}

\begin{figure*}
    \centering
    \begin{minipage}{\columnwidth}
        \begin{overpic}[trim={0cm 0cm 0cm 0cm},clip, width=\textwidth]{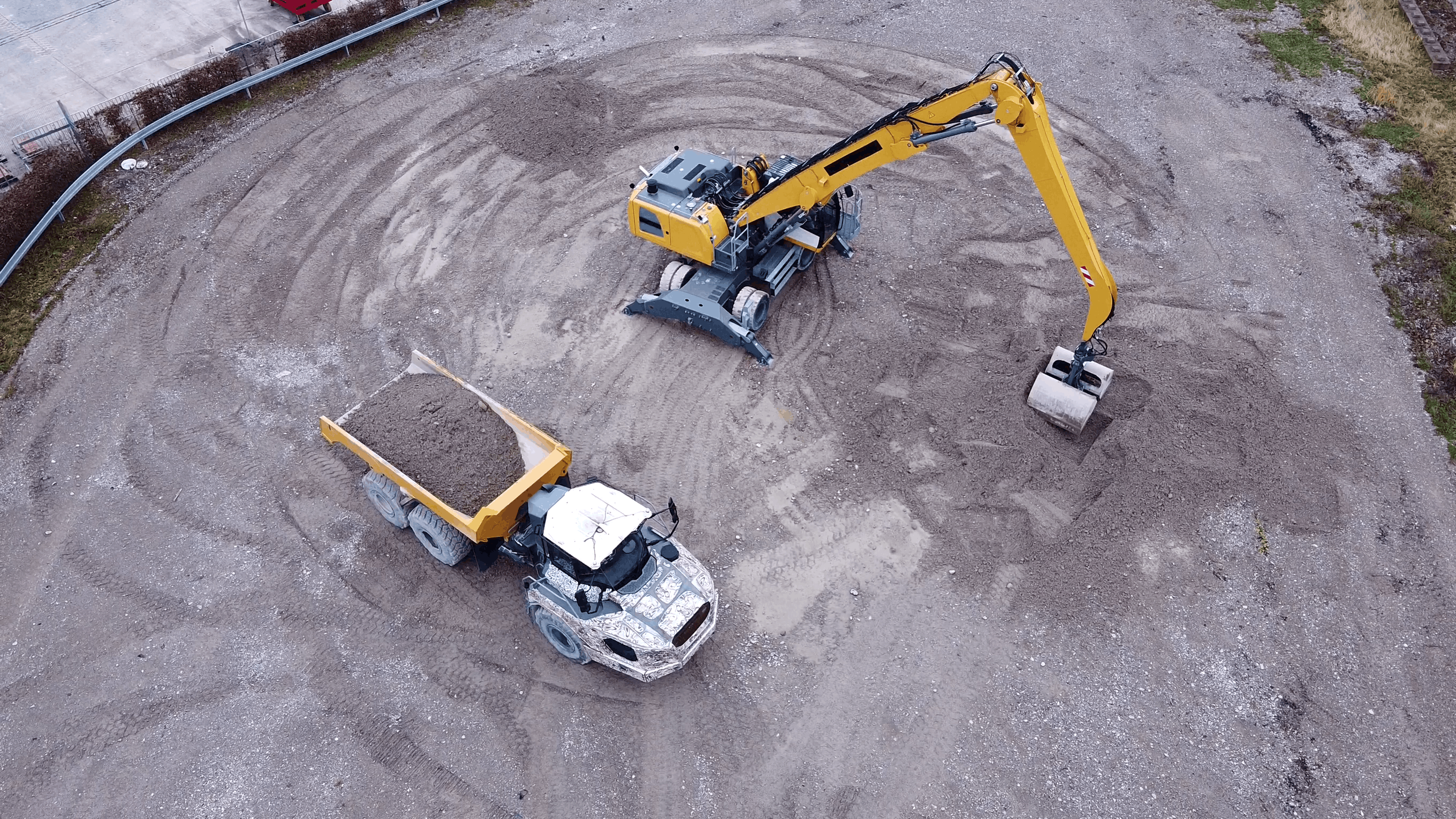}
            \put(5,5){\textcolor{white}{\textbf{A}}}
        \end{overpic}
        \par\vspace{0.5\columnsep}
        \begin{overpic}[trim={0cm 0cm 0cm 0cm},clip, width=\textwidth]{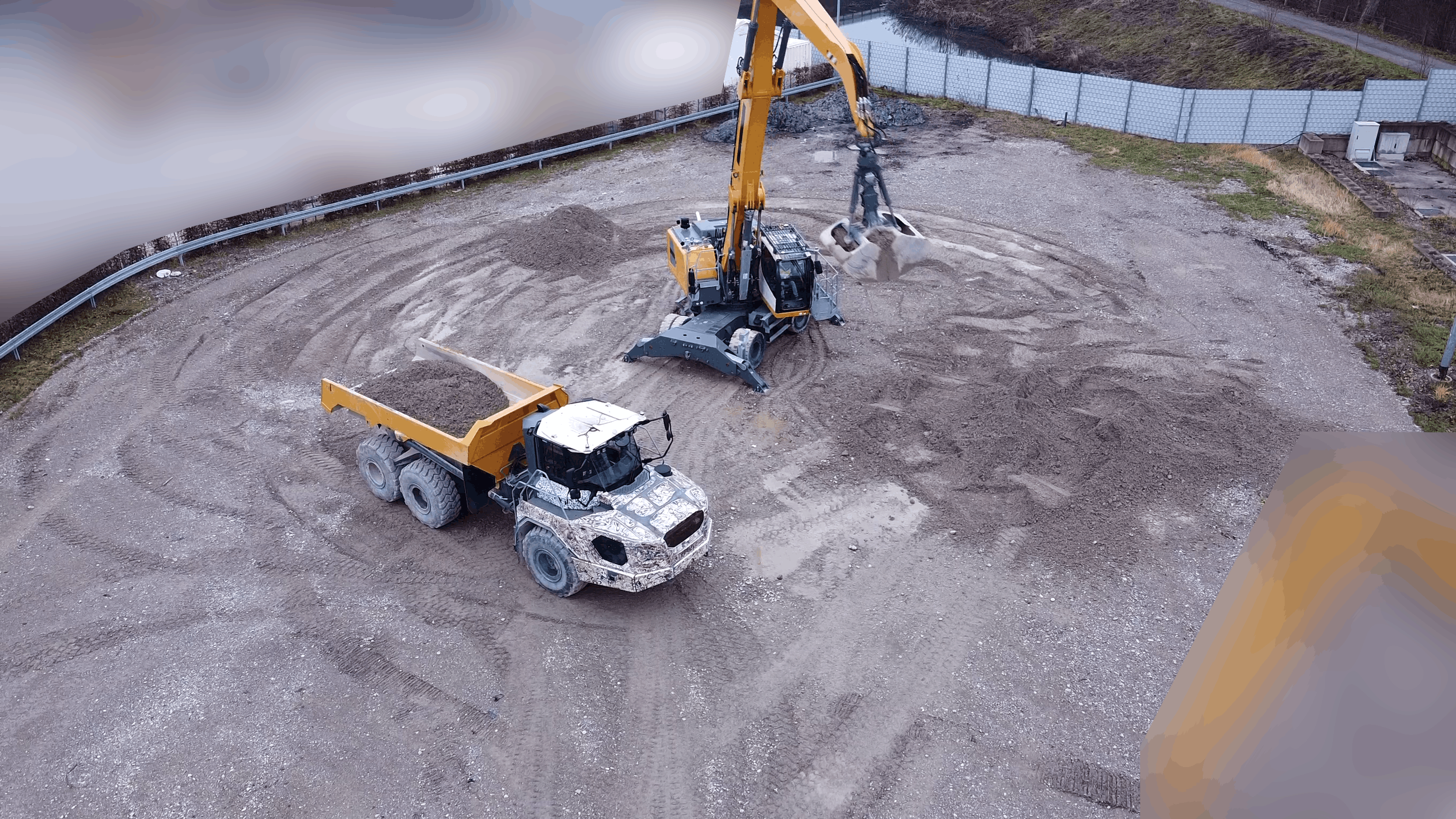}
            \put(5,5){\textcolor{white}{\textbf{C}}}
        \end{overpic}
        \par\vspace{0.5\columnsep}
        \begin{overpic}[trim={0cm 0cm 0cm 0cm},clip, width=\textwidth]{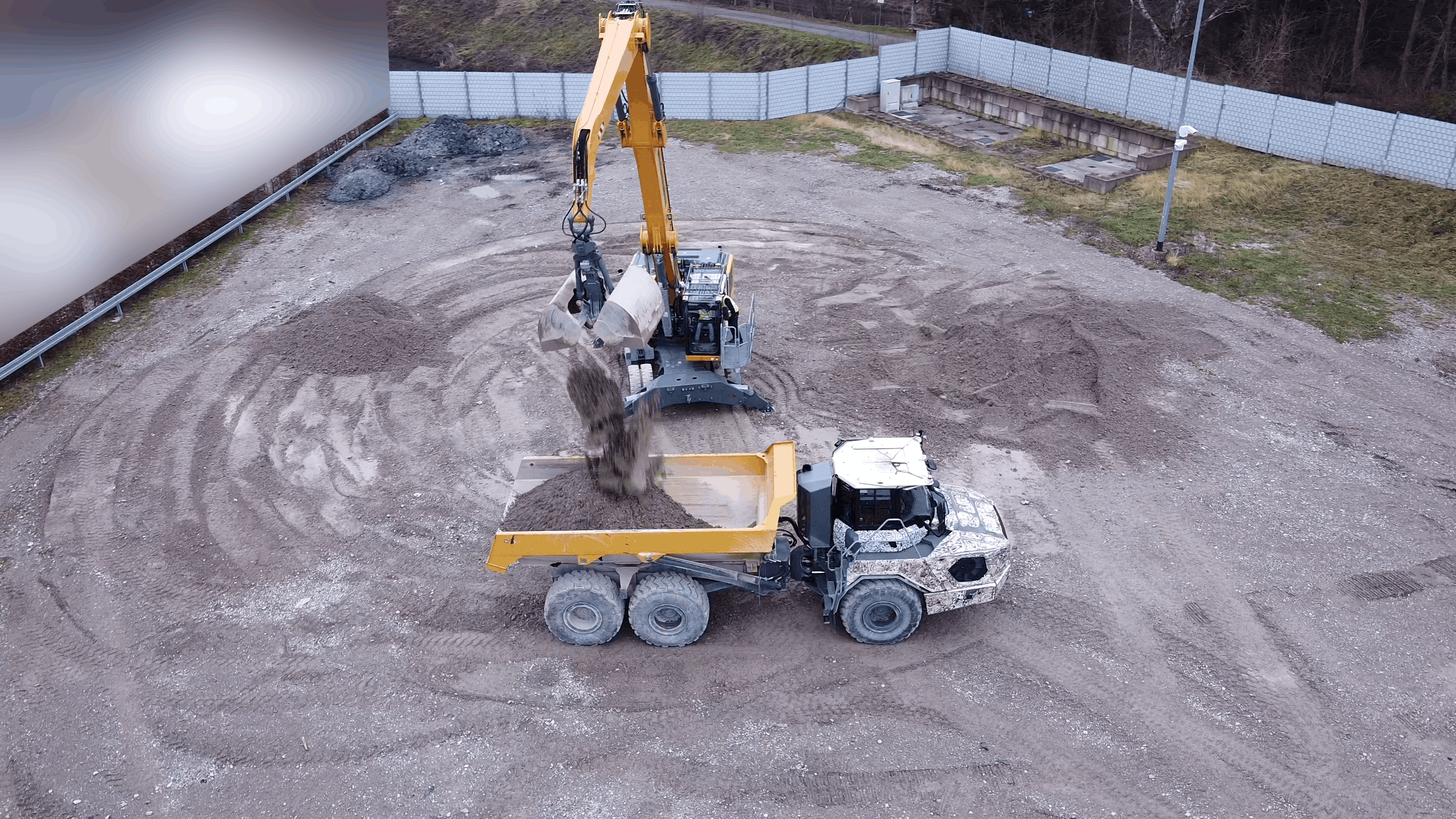}
            \put(5,5){\textcolor{white}{\textbf{E}}}
        \end{overpic}
    \end{minipage}
    % \hspace{-0.6\columnsep}
    \begin{minipage}{\columnwidth}
        \begin{overpic}[trim={11cm 2cm 3cm 6cm},clip, width=\textwidth]{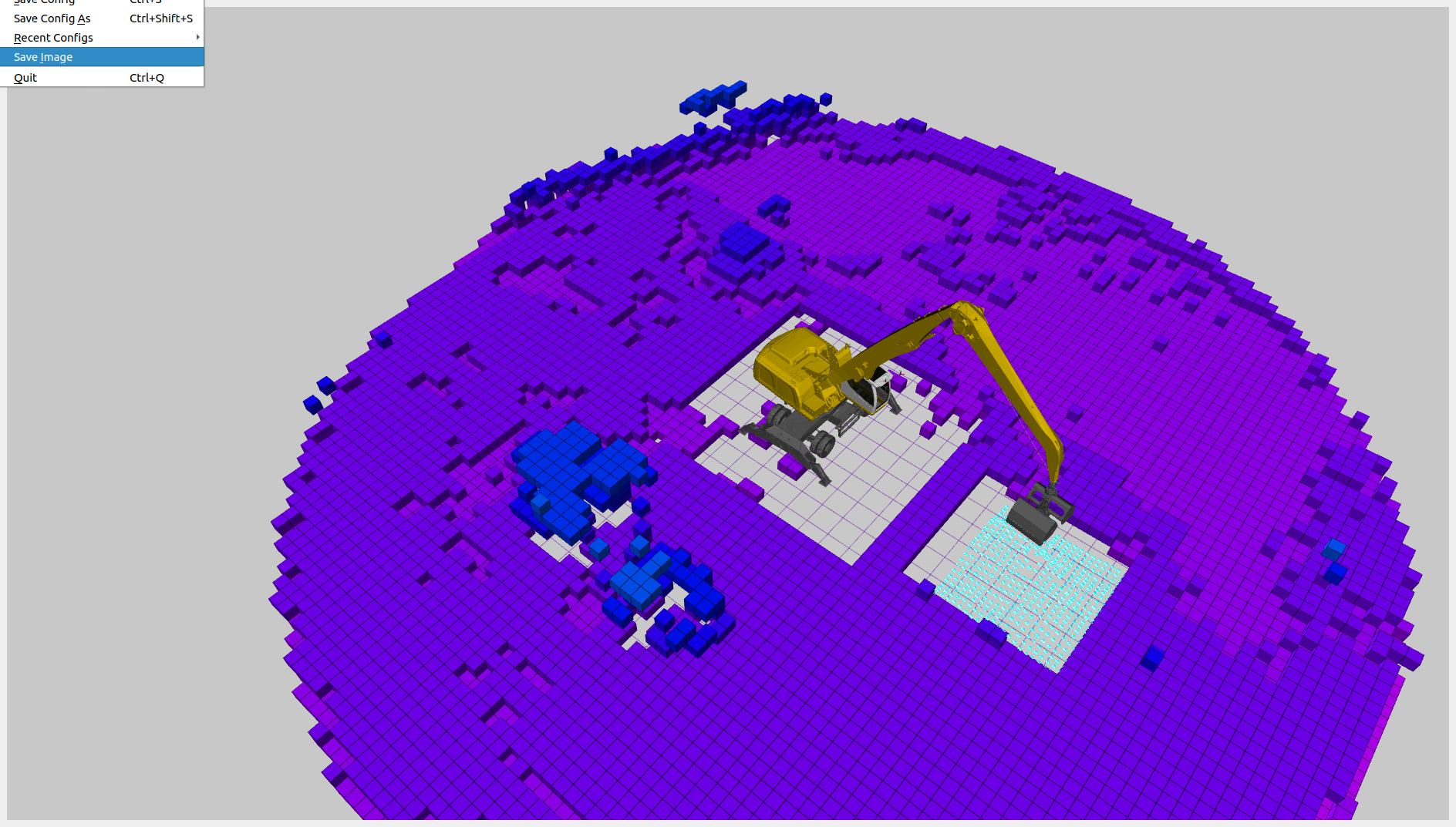}
            \put(5,5){\textcolor{white}{\textbf{B}}}
        \end{overpic}
        \par\vspace{0.5\columnsep}
        \begin{overpic}[trim={12cm 2cm 2cm 6cm},clip, width=\textwidth]{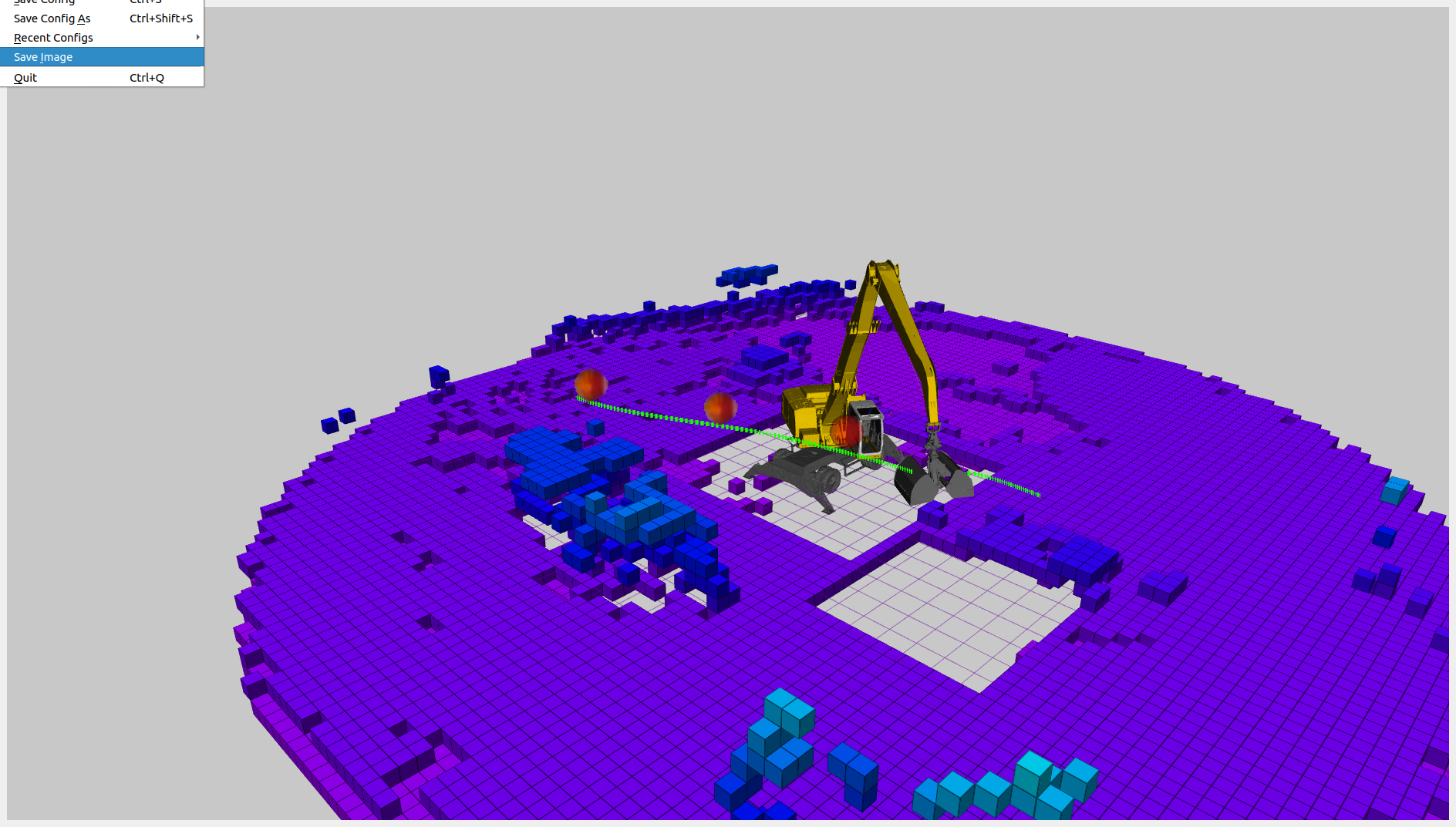}
            \put(5,5){\textcolor{white}{\textbf{D}}}
        \end{overpic}
        \par\vspace{0.5\columnsep}
        \begin{overpic}[trim={8cm 4cm 6cm 4cm},clip, width=\textwidth]{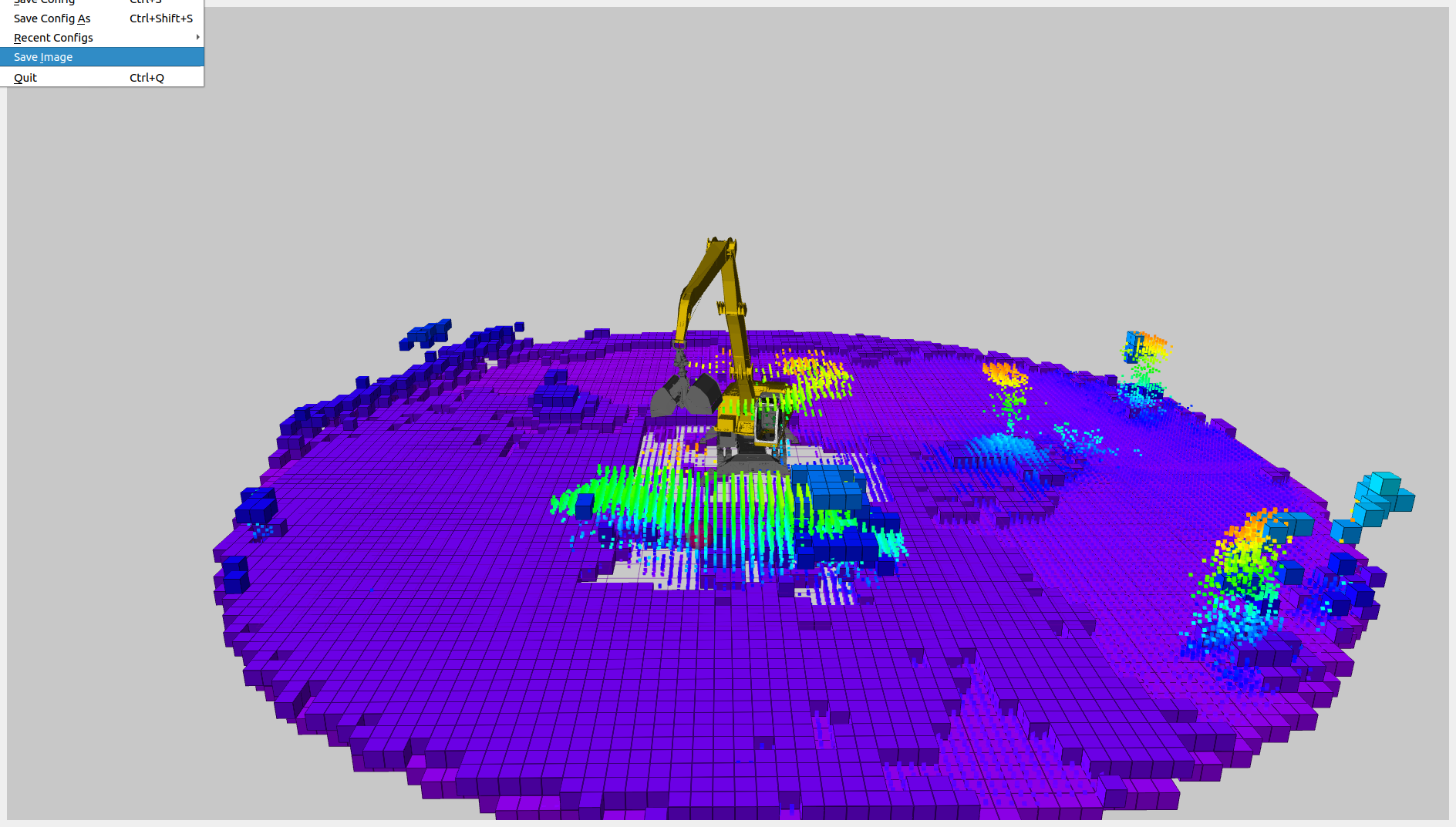}
            \put(5,5){\textcolor{white}{\textbf{F}}}
        \end{overpic}
    \end{minipage}
    \caption{The proposed framework performing the dump truck loading routine. The framework consists of key modules such as the RL attack point planner, the obstacle-avoiding path planner, the RL waypoint-following controller, and the RL throwing controller. This figure demonstrates the orchestration of these modules in a single grasp-and-dump cycle. First, the RL attack point planner proposes the optimal attack point based on LiDAR observation of the pile geometry, shown as light blue dots (Subfigures A and B). Then, the sampling-based path planner proposes a collision-free path, shown as a green dotted line. A sequence of waypoints sampled from the path, represented by red spheres, is tracked by the RL throwing policy (Subfigures C and D). Finally, the RL throwing controller dumps the load at the user-defined location in the the truck bed (Subfigures E, F). In addition to proprioceptive data, the framework uses the LiDAR pointclouds for exteroceptive observations: the cropped pointcloud of the pile is directly used to grasp planning (Subfigure B), while voxels converted from the pointclouds are used to represent obstacles in the environment (Subfigure D, F).}
    \label{fig:dump_truck_drone}
    
\end{figure*}

Robotic manipulation has been widely adopted in various industries, such as manufacturing and logistics~\cite{IFRInternationalFederationOfRoboticsWorldRobotics, ValuatesReportsRoboticManipulators}, to automate repetitive and labor-intensive tasks.
While large-scale hydraulic manipulators pose significant challenges due to their highly nonlinear dynamics \cite{Nan24LearningAdaptive, Egli22GeneralApproach}, they have already been successfully automated for real-world applications such as stone wall building~\cite{Johns23FrameworkRobotic}, free-form trenching~\cite{Jud2019Autonomous}, tree harvesting~\cite{Jelavic22HarveriSmall}, and long-horizon excavation~\cite{Terenzi24AutonomousExcavation}. These works have shown the potential of robotic manipulation to improve efficiency, safety, and sustainability across various industries.

However, the automation of bulk material handling tasks using hydraulic manipulators remains underexplored in research. These tasks are critical in mining, recycling, and construction industries, where materials such as ores, gravel, or sand are handled in large volumes. Typically stored in piles, these materials must be routinely transferred, sorted, and relocated. To perform these operations, hydraulically actuated heavy manipulators known as material handlers are commonly used. While similar to hydraulic excavators, material handlers differ in arm geometry, cylinder placement, and attachments to enhance lifting capacity and material transfer efficiency. In particular, the end-effector tools, usually clamshell grippers, are attached to unactuated joints, which reduces manufacturing costs, facilitates grasping, and enables agile behaviors.
Automating these machines presents three key challenges:
\textit{i)} the complex, granular behavior of bulk materials,
\textit{ii)} the nonlinear dynamics of large hydraulic systems combined with an underactuated gripper, and
\textit{iii)} the need for safe and robust operation in unstructured environments.

First, bulk materials are inherently difficult to model and manipulate. Their behavior differs significantly from that of rigid objects~\cite{han2023survey}, requiring specialized approaches in perception, planning, and interaction. For example, high-throughput material transfer requires a planner capable of perceiving the geometry of the source pile and selecting optimal attack points to maximize the volume collected each grasp.

Second, like other large hydraulic machines, material handlers exhibit complex dynamics and actuation delays that complicate precise control~\cite{Nan24LearningAdaptive}. Reliable automation typically requires data-driven approaches that capture the full range of system behaviors and uncertainties~\cite{Egli22GeneralApproach}. This challenge is compounded by the underactuated kinematics of the free-hanging tool, which passively swings during movement. For human operators, efficient use of these grippers demands coordinated machine motions, a skill developed through experience. Failure to manage this coordination can lead to reduced operational efficiency, decreased accuracy, potential equipment damage, and reduced safety.

Lastly, material handlers are often deployed in unstructured environments such as construction sites and harbors, where unexpected obstacles pose additional risks. Safe operation requires the autonomous system to perceive its surroundings and plan movements accordingly, taking into account the passive motion of the gripper to avoid collisions. Beyond collision-free path planning, automation systems must implement control policies that can precisely track trajectories and stabilize the swinging gripper.

In this work, we present a framework that addresses the three aforementioned challenges of bulk material handling by breaking down tasks into grasp-and-dump cycles, which the system autonomously executes. In each cycle, the system selects an optimal attack point on the pile, plans a collision-free trajectory to the target, tracks it with bounded error, and accurately releases the material at the target location. The sequence of these steps is orchestrated by a state machine. Our autonomous system is capable of executing two representative tasks: (1) transferring bulk material from a source pile to a target location with high throughput, and (2) loading bulk material from a pile onto a dump truck~\cite{stentz1999robotic} with high accuracy, as illustrated in~\cref{fig:dump_truck_drone}. These scenarios reflect typical applications of material handlers and represent some of the most demanding use cases in terms of efficiency, precision, and safety. In both tasks, the material handler’s base and dumping locations are fixed, and the motion is limited to arm only while ensuring collision avoidance with environmental obstacles.
 We evaluate performance based on the average time per grasp-and-dump cycle, the precision of material throwing, the safety margins during motion in unstructured environments, and the overall effectiveness of the system in hydraulic machine control and bulk material manipulation.

\subsection{Related Work}

\subsubsection{Large-Scale Field Robotic Manipulation Systems}

In recent years, there has been a growing effort to automate large-scale industrial machines for autonomous manipulation operations, with a significant amount of research focused on construction machinery. 
Advanced soil manipulation has been demonstrated using robotic excavators, such as the autonomous embankment pipeline of ~\cite{Jud21RoboticEmbankment}, capable of shaping the terrain into complex shapes. The method encompasses mapping (performed with the help of a drone), planning, and control to build a robotic landscaping system that exchanges information in real-time during construction. 
This work mostly relies on proprioception and \ac{THO}-based control to operate on realistic soil conditions. To improve manipulation capabilities, especially with granular material, a vision-driven adaptation has been proposed~\cite{Zhu23FewshotAdaptation}: by incorporating RGB-D image patches directly into a deep Gaussian process prediction model, the robot can visually assess local terrain properties without prior terrain knowledge, and adjust the scooping strategy accordingly.
A further step towards large-scale material manipulation is performed by Zhang et al.~\cite{Zhang21AutonomousExcavator}. They developed an excavator automation system for material handling tasks in unstructured environments, integrating advanced perception, planning, and control methods. The system features hierarchical task and motion planning, blending learning-based techniques with optimization-based methods, ensuring seamless integration between perception and control. Tested on various scenarios such as dump truck loading, waste handling, rock capturing, pile removal, and trenching, it achieves performance comparable to experienced human operators, enhancing efficiency and robustness in real-world applications. 

Equipping the end-effector with an actuated gripper enables the integration of active manipulation capabilities, thereby allowing the system to tackle more complex tasks. 
For instance, dry stone walls were constructed in~\cite{Mascaro20AutomatingConstruction, Johns23FrameworkRobotic} using only in-situ materials. 
The authors proposed a comprehensive framework encompassing rock shape scanning via onboard sensors, placement planning to ensure structural stability, and precise positioning of each rock at its designated location.
Similar abilities extend naturally to forestry applications. Jelavic et al.~\cite{Jelavic22RoboticPrecision} developed a precision tree harvesting pipeline in which a robot autonomously scans the forest, segments individual trees, and grasps those selected by an operator. 
Likewise, La Hera et al.~\cite{LaHera24ExploringFeasibility} introduced an autonomous forwarder capable of navigating clear-cut forest trails, detecting and localizing logs using stereo vision, and transporting them to a collection point via fully automated crane manipulation and vehicle control. 
These contributions collectively underscore the viability of fully autonomous operations in forestry environments.

While the aforementioned perception and planning strategies are broadly applicable to field robotic applications, the control techniques previously discussed cannot be directly applied to our problem domain. 
State-of-the-art material-loading controllers typically assume a fully actuated end-effector and often overlook the complex interaction between the shovel and the soil.
Conversely, although the forest forwarder shares the underactuated kinematic structure common to most material handlers, autonomous pipelines are primarily designed to grasp single, rigid objects and have been tested exclusively in unstructured yet open environments.

\subsubsection{Control of Hydraulic Machinery}
In the automation of large-scale heavy machinery, one of the primary challenges is achieving precise control of the hydraulic actuators.
Some early works in this direction used model-based control~\cite{Ha02RoboticExcavation, Chang02StraightlineMotion, Mattila17SurveyControl}, relying heavily on the availability of an accurate model to address delays and nonlinearities in the system.
In practice, obtaining such a model analytically can be challenging, which limits the applicability of these control methods.
While retrofitting of high-performance hydraulic valves reduces the modeling effort~\cite{Jud2019Autonomous}, their practical use is limited by high costs and restricted oil flow, which typically permits only slow arm motions.
In response, recent studies have increasingly used data-driven techniques in the modeling and control algorithms.
Nurmi et al.~\cite{Nurmi18NeuralNetwork} proposed a deep learning-based method to identify nonlinear velocity \ac{FF} curves for pressure-compensated hydraulic valves.
Park et al.~\cite{Park17OnlineLearning} introduced an online learning framework for the position control of hydraulic excavators using echo-state networks, where a model is trained online from input-output time series data to generate control commands for new trajectory references.

Recently, \ac{RL} has emerged as an alternative to control hydraulic excavation machines.
Egli and Hutter~\cite{Egli20RLBasedHydraulic, Egli22GeneralApproach} proposed a data-driven approach for modeling hydraulic cylinders on excavators.
A trained \ac{NN} model predicts excavator behaviors under arbitrary valve commands, and is used as a simulation environment for an \ac{RL} agent to learn end-effector position or velocity tracking in free space and with weak ground contact.
Dhakate et al.~\cite{Dhakate22AutonomousControl} considered the control of a small forest forwarder with a free-swinging gripper, leveraging similar ideas. Their work captures the mapping between cylinders' displacements and joint variables through a \ac{NN} model.
They then train an \ac{RL} position controller, which commands joint setpoints, simply treating the unactuated tool joint as a disturbance.
Spinelli et al.~\cite{Spinelli24ReinforcementLearning} followed on this approach to train an end-effector tracking controller for material handling machines, demonstrating a successful sim-to-real transfer. They subdivide the modeling into two parts: \textit{i)} a data-driven method is used for modeling the dynamics of the cabin turn (slew) joint, with pressure and inertia as additional features, and \textit{ii)} first-principles modeling is adopted for the rest of the machine. While this work develops a control strategy that accounts for passive tool oscillation minimization, it does not address the complexity of real construction sites, as it operates only in free space and neglects potential collisions with obstacles.

Following the accuracy achieved by these novel methods, Lee et al.~\cite{Lee22PrecisionMotion} propose a model inversion control approach where a graybox model structure is used to reduce the number of parameters to be learned from data.
By decomposing the model into physics-inspired components and arbitrary nonlinearities, model and controller learning could be performed at much higher efficiency.
Nan and Hutter~\cite{Nan24LearningAdaptive} further improved the learning efficiency by proposing an offline pre-training with online adaptation workflow.
An adaptive controller is trained solely with simulated data and can be efficiently tailored to different hardware platforms at runtime.

\subsubsection{Control in the Presence of Passive Joints}
Prior research in the construction domain has seldom addressed the complexities of free-swinging end-effectors, and how to use them for manipulation tasks. However, insights can be drawn from other similar tasks.
For example, the control of slung-load systems with aerial vehicles has received considerable attention, with
both quadrotors~\cite {DeCrousaz15UnifiedMotion} and helicopters~\cite{Oktay13ModelingControl} considered. Proposed solutions involve using either trajectory optimization~\cite{Sreenath13TrajectoryGeneration} or \ac{RL}~\cite{Palunko13ReinforcementLearning} to control both vehicle and load states at the same time.
In the robotic manipulation domain, Zimmermann et al.~\cite{Zimmermann19PuppetMasterRobotic} developed a computational framework for animating robotic string puppets. These are coupled pendulum systems, sharing similar dynamics with a freely swinging gripper. The proposed predictive control, relying on an accurate kinematic and dynamic model, it is however not well suited for construction robots. 
Some promising results have been achieved in the safe control of tower cranes~\cite{Ramli17ControlStrategies, Rauscher21ModelingControl}, more closely related to material handling, although their operational routines usually do not involve any aggressive motions.
Among the others, Zhang et al.~\cite{Zhang23DeepReinforcement} have applied \ac{RL} to tower crane control under varying payload conditions and achieved better performance than linear controllers.

Active control of construction machines with underactuated kinematics has received limited attention, with notable focus on hydraulic forestry cranes. Targeting the free-hanging tool dynamics, Kalmari et al.~\cite{kalmari2014nonlinear} applied nonlinear \ac{MPC} to track predefined paths while damping load oscillations. On-machine experiments confirmed that anti-sway control preserved both tracking accuracy and velocity. Further studies~\cite{paz2025full} integrated hydraulic actuators limitations into the constrained optimization, demonstrating the feasibility of real-time \ac{MPC} in complex hydraulic systems. Such controllers can be coupled with optimal motion planners~\cite{jebellat2024motion} that account for the gripper dynamics in order to operate in realistic scenarios.  
Efforts to reduce computational complexity include research on feed-forward alternatives~\cite{kowsari2024optimal}, though model-based controllers still depend on accurate system identification to achieve optimal performance. As an alternative, learning-based approaches have emerged. For instance, Andersson et al.~\cite{Andersson21ReinforcementLearning} trained an \ac{RL} policy for a forestry forwarder to grasp logs. The emergent behavior was showed capable of exploiting the passive gripper swing to improve efficiency, albeit without hardware validation. Similarly, \cite{vu2025towards} introduced a simulation framework for training \ac{RL} agents on large-scale manipulation tasks, initially targeting log grasping.  
In the context of material handlers, the policy of~\cite{Spinelli24ReinforcementLearning} shaped motion profiles to minimize oscillations, while \cite{Werner24dynamicThrowing} used \ac{RL} to enable dynamic throwing, extending the workspace beyond static pick-and-place. However, their method was limited to single-object scenarios and structured environments.

\subsubsection{Attack Point Planning for Bulk Material Transfer}
Carefully planned end-effector attack points are crucial for efficient material handling regardless of the material-moving machine type. Optimized attack points maximize the volume grasped each time, reducing the total number of cycles required to transfer the bulk material and thereby improving overall efficiency. 

For wheel loaders, the machine configuration constrains candidate attack points to the interface between material piles and the ground, effectively reducing the search space to a one-dimensional optimization problem. As a result, most studies concentrate on directly evaluating these candidate points.
Magnusson et al.~\cite{magnusson2011pileshape} estimated of convexity and sideload metrics by fitting quadric surfaces to the pile. Wang et al.~\cite{wang2022bucketfill} proposed a more direct evaluation approach by predicting the bucket fill ratio through co-simulation, combining a multi-body dynamics model of the machine with a \ac{DEM} simulation of the material. To mitigate the computational cost of high-fidelity simulations such as \ac{DEM}, Aoshima et al.~\cite{aoshima2024worldmodel} trained a world model on simulated loading data to predict the loaded mass, cycle time, and work for a single loading operation, enabling more efficient optimization. Utilizing this world model, \cite{aoshima2025end-to-end} jointly optimized loading performance and transportation efficiency over multiple cycles, including the time and energy costs associated with the driving maneuvers between loading and dumping phases.

For excavators and material handlers, the problem formulation is expressed as 2.5D, as the end-effector can grasp material from any location on the pile surface. Moreover, repositioning the base between successive cycles does not significantly affect time or energy consumption, since both scooping and dumping require only arm movements. In most existing autonomous solutions for these machines, heuristic-based approaches are employed to select attack points. In~\cite{Wang21HierarchicalPlanning}, the trajectory that contains the highest point in the excavation zone is chosen. In~\cite{Jud17PlanningControl}, a simple optimization on the highest points and slope of the soil is used to find the attack point. In~\cite{Terenzi24AutonomousExcavation}, the excavation attack points that result in the most scooped soil volume within the excavation zone are found by Bayesian optimization in real-time. 
In~\cite{zhao2021TaskNet}, a cell excavation model is trained on synthesized data using a rule-based simulation that digs at the highest grid in the local map. A small amount of human demonstrations are then used to fine-tuned the model. Such model has been successfully deployed in a robust system that achieves 24 hours of continuous operation per intervention~~\cite{Zhang21AutonomousExcavator}.
All of the above methods are designed around heuristics, which restrict the general applicability of the planner. Furthermore, these methods rely on robust state estimators to convert LiDAR measurements into soil elevations, necessitating laborious and time-consuming fine-tuning.

\subsection{Contribution}

This work proposes a complete framework for autonomous bulk material handling with a focus on fixed-base manipulation. To the best of our knowledge, this is the first full-scale demonstration of such an autonomous system. While several components are built on established techniques, the key novelties of this work are summarized as follows:

\begin{itemize}
    \item A perceptive \ac{RL} planner that proposes the optimal attack points on the pile ensuring efficient material removal with a minimum number of scoops. 
    
    \item A two-step arm motion planning and control pipeline that combines a sampling-based planner with an \ac{RL} trajectory tracking controller.

    \item An \ac{RL} tracking policy that follows a provided trajectory while maintaining bounded tracking error and reducing oscillations of the free-swinging gripper, enabling safe deployment in unstructured environments.

    \item An \ac{RL} throwing policy that tracks a provided trajectory, and dynamically dumps bulk material at the given target by shaping the gripper swinging motion. It is trained accounting for the granular material nature.
    
    \item An experimental validation of the proposed framework at a representative worksite, supported by ablation studies in simulation. The real-world evaluation focuses on high-throughput material transfer and high-precision truck loading, including benchmarks against human operators.
\end{itemize}
\section{Machine Setup \& Low-Level Control}
\label{sec:hardware-setup}

\begin{figure}
    \centering
    \includegraphics[trim={3.5cm 14cm 5cm 4cm},clip, width=\columnwidth]{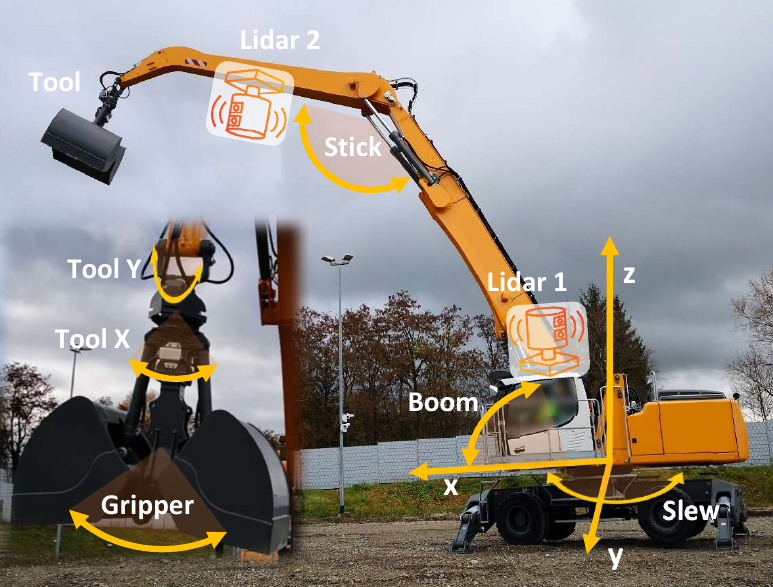}
    \caption{The material handler used in this work has an operational range of about \SI{20}{\meter} and weighs around \SI{40}{\tonne}. It is equipped with a \SI{1.5}{\tonne} grabshell gripper designed for bulk material, with a maximum load capacity of \SI{1.5}{\cubic \meter}. When fully open, it covers an area of approximately $2.0\times1.5 \si{\meter \squared}$. 
    }
    \label{fig:machine_frame}
    % \vspace{-0.5cm}
\end{figure}

\subsection{Machine Overview}
We conduct our experiments on a custom research platform shown in~\cref{fig:machine_frame}. The arm features three hydraulically actuated joints, one \ac{DoF} each: the slew turn joint is actuated with a rotary motor, while the boom and stick joints are actuated with linear pistons. The end-effector tool, a two-jaw clamshell gripper, is attached to the machine's arm via two unactuated joints, allowing it to swing freely. The gripper tool provides two additional \acp{DoF}: shell opening/closing, and yaw rotation around its vertical axis. 
The tasks of interest of this work target fast and repeated material transfer motions, which typically don't require human operators to adjust the gripper orientation, and to drive the machine to different locations. Consequently, we fix the yaw joint at its default position, as shown in~\cref{fig:machine_frame}, and considered the machine base static. As a result, the system operates with a total of six \acp{DoF}, with the two tool \acp{DoF} unactuated.

\subsection{Sensors \& Perception}
The following sensors are installed:
\begin{itemize}
    \item Position encoders on slew, boom, and stick joints, providing position and velocity measurements. 
    \item An \ac{IMU} on the gripper, providing 3D angular velocities, 3D linear accelerations, and 2D angular orientation. 
    \item Pressure sensors on slew, boom, stick, and gripper cylinders, measuring fluid pressure on the piston side and rod side of each cylinder.
    \item Two LiDARs, one mounted on the cabin, the other one on the stick.
\end{itemize}

These sensors enable observation of the machine’s slew, stick, and boom states, as well as the pitch and roll of the tool, with 50~\si{\hertz} update rate. Exteroceptive observations of terrain elevation and obstacles in the workspace are obtained by fusing and filtering the two LiDAR point clouds into a single unified representation.

The current sensor setup presents several limitations.  
First, velocities of the slew, boom, and stick joints are estimated from position data, introducing noise and latency.  
Second, the clamshell gripper’s absolute position is unobservable; only its binary open/closed state can be inferred from pressure readings.  
Third, the gripper’s yaw joint is similarly unobservable due to the absence of a magnetometer in the sensor configuration. As a result, yaw control and shell opening angle are omitted in the current setup.  
Finally, the fused point cloud may be contaminated by falling material, particularly when the loaded gripper is lifted above the pile, requiring dedicated filtering strategies.

\subsection{Low-Level Joint Velocity Controller}\label{sec:arm_vel_controller}

\begin{figure}
    \centering
    \includegraphics[trim={1cm 1cm 0cm 0cm},clip,width=\columnwidth]{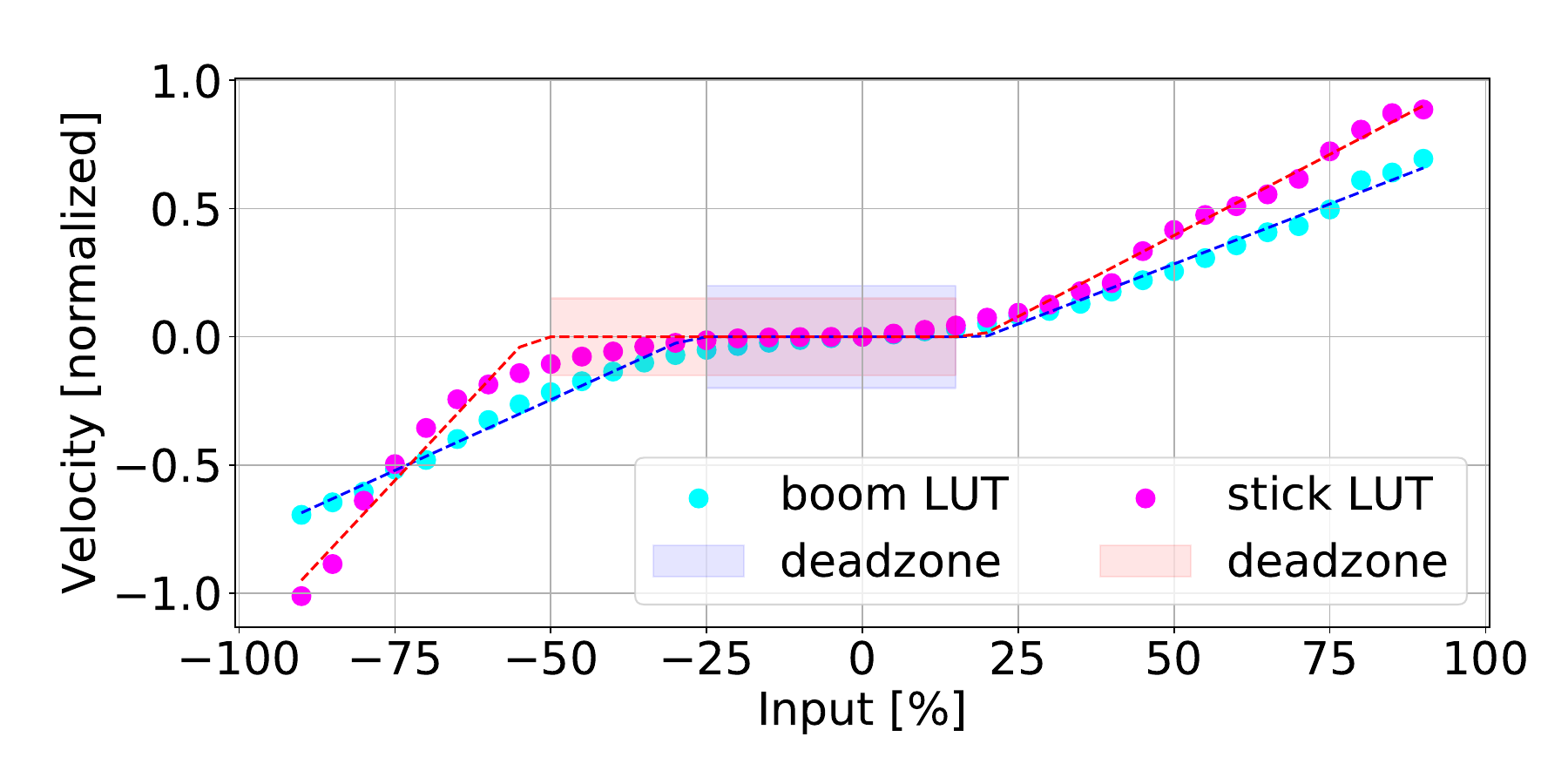}
    \caption{The LUTs are constructed from 37 datapoints each, obtained by measuring the steady-state velocity response of each joint under step excitations. 
    Each value is computed by averaging the final $2\si{\second}$ of recorded velocities for a given input signal, in line with the steady-state assumption. 
    While the boom behavior is relatively symmetric, the stick dynamics exhibit a pronounced asymmetry in the input-velocity relationship.
    }
    \label{fig:LUTs}
\end{figure}

A low-level control module is introduced to convert the arm control outputs into machine joystick commands. 
Specifically, the joint-level controllers convert \SI{10}{\hertz} velocity references $\dot{q}^{\text{~ref}}$ or open/close gripper binary actions to normalized joystick commands $\mathcal{J}$ at \SI{50}{\hertz}, which are sent to the machine's CAN bus to actuate the joints. Due to the differing dynamic characteristics of each joint and limitations in the sensor setup, distinct control strategies are required for individual joints, as summarized in \cref{eq:low_lev_ctrl}.

\begin{equation}\label{eq:low_lev_ctrl}
\begin{aligned}
    \mathcal{J}=
    \begin{cases}
        J_{\text{boom}} = FF_{\text{boom}} + PI_{\text{boom}}(\dot{q}_{\text{~boom}}^{\text{~ref}} - \dot{q}_{\text{boom}} ) \\
        J_{\text{stick}} = FF_{\text{stick}} + PI_{\text{stick}}(\dot{q}_{\text{~stick}}^{\text{~ref}} - \dot{q}_{\text{stick}} ) \\
        J_{\text{slew}} = ZOH(u_{\text{slew}}) \\
        J_{\text{grab}} = \begin{cases}
            J_{\text{grab,open}} & \text{if} \quad u_{\text{gripper}} = \text{open} \\
            J_{\text{grab,close}} & \text{if} \quad u_{\text{gripper}} = \text{close}
        \end{cases}
        \\
    \end{cases}
\end{aligned}
\end{equation}

For the boom and stick joints, we adopt a model-based control approach to track the reference joint velocities, $\dot{q}_{\text{~boom}}^{\text{~ref}}$ and ${\dot{q}_{\text{~stick}}^{\text{~ref}}}$. Following previous works~\cite{Jud2019Autonomous, Egli22SoilAdaptiveExcavation, Werner24dynamicThrowing}, we implement a PI feedback controller augmented with a \ac{FF} term derived from a \ac{LUT} for each joint. 
The \ac{LUT} maps joystick inputs to the steady-state joint velocities, based on empirical step-response experiments. As shown by the \acp{LUT} in \cref{fig:LUTs}, both boom and stick velocity responses follow a regular pattern, close to ideal piece-wise linear functions as described in~\cite{Nurmi18NeuralNetwork}. The inverse mapping allows the \ac{FF} term to directly generate a joystick command that produces the desired steady-state velocity. The PI controller then compensates for transient dynamics and unmodeled disturbances to ensure accurate velocity tracking. An additional safety layer is then implemented, guaranteeing compliance with user-defined joint limits to enforce self-collision avoidance. 

The slew joint, in contrast, does not have an easily identifiable mapping from joystick input to joint velocity, since the velocity response is influenced by additional factors such as configuration-dependent inertia around the rotation axis~\cite{Spinelli24ReinforcementLearning}. To address this, we employ a learning-based dynamics model for the slew joint, which is integrated into our \ac{RL} training environment as described in \cref{sec:sim_dynamics}. Consequently, the RL policy directly outputs joystick commands $u_{\text{slew}}$ for the slew joint. These commands are passed through a \ac{ZOH}, upsampled and then sent to the CAN bus, without the need of low-level velocity tracking controller.

For the gripper joint, lacking an encoder, only a binary open/closed state can be inferred from pressure measurements. Consequently, we operate the gripper in a binary mode, controlled by the binary gripper command $u_{\text{gripper}}$. A constant joystick input, $J_{\text{grab,open}}$ or $J_{\text{grab,close}}$, is sent to the machine to open or close the gripper at a fixed, suitable velocity, depending on $u_{\text{gripper}}$.

\section{Method}

Our material manipulation framework, building on the hardware, sensors, and low-level control described in \cref{sec:hardware-setup}, consists of distinct modules that can be categorized as follows:

\newlist{catlist}{itemize}{1}
\setlist[catlist]{label=--, leftmargin=2.5em, itemsep=0pt, topsep=0pt}

\vspace{0.5em}
\Needspace{3\baselineskip}
\textbf{Planners}
\begin{catlist}
  \item RL attack point planner~(\cref{sec:rl_grasp})
  \item Obstacle-avoiding arm path planner~(\cref{sec:planner})
\end{catlist}

\vspace{0.25em}

\Needspace{4\baselineskip}
\textbf{Controllers}
\begin{catlist}
  \item RL waypoint-following controller~(\cref{sec:waypoint_rl})
  \item RL throwing controller~(\cref{sec:throw_rl})
  \item Grasping controller~(\cref{sec:grasp_control})
\end{catlist}

\vspace{0.25em}

\Needspace{2\baselineskip}
\textbf{Controller Coordination}
\begin{catlist}
  \item High-level state machine~(\cref{sec:state_machine})
\end{catlist}
\vspace{0.5em}

In particular, the RL attack point planner and the RL controllers constitute the primary contributions of this work. 
Due to the unobservability of the gripper shell state, two different control policies are defined: the waypoint-following one moves the empty open gripper to the planned grasping location, the throwing one, assuming a loaded gripper, learns to release the material into a pile.
In \cref{sec:method_mh_pipeline}, an overview of the entire pipeline and the orchestration of these modules is presented, followed by detailed descriptions of each module in the subsequent sections.

\begin{figure*}
    \centering
    \includegraphics[trim={3cm 6.5cm 4cm 2cm},clip, width=2\columnwidth]{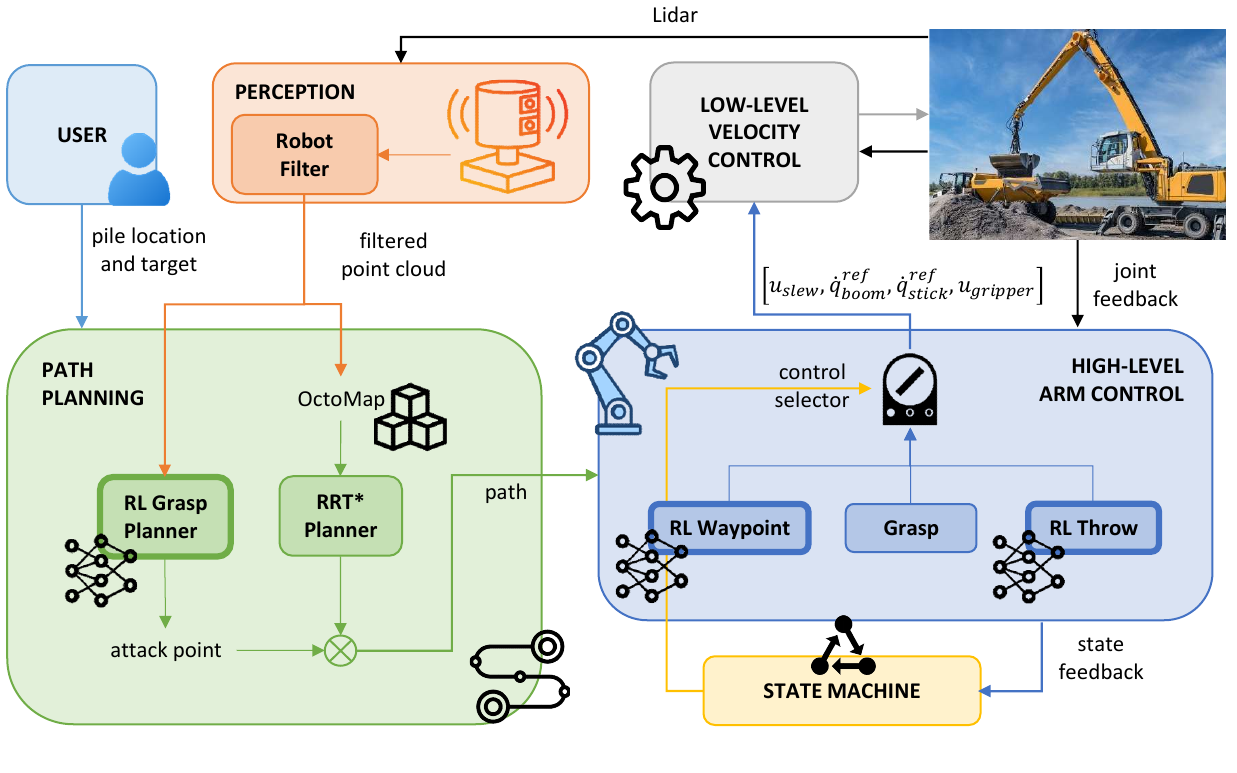}
    \caption{Component diagram of our framework. Given a specified grasping region and dumping target, the pipeline autonomously controls the machine in closed loop. The planning module takes into account the user input and LiDAR point cloud data to generate collision-free paths toward either the optimal grasp point or the designated dumping target. These paths are followed by high-level control policies that guide the end-effector through subsampled waypoints. A state machine manages the active controller, selecting between waypoint following, grasping, and throwing, ensuring that only one controller sends commands to the low-level control module at any time.}  
    \label{fig:ros2_overview}
\end{figure*}

\subsection{Material Handling Pipeline}
\label{sec:method_mh_pipeline}
A graphical overview of the pipeline and the interactions between its modules is shown in \cref{fig:ros2_overview}. The user specifies a bounding box around the source pile and defines a target location for material release. Based on these inputs, the framework autonomously executes material handling through repeated grasp-and-dump cycles.
The detailed step-by-step procedure for a single cycle is as follows:

\begin{enumerate}
  \item \textbf{Approaching source pile with empty gripper}
  \begin{enumerate}
    \item The RL attack point planner proposes an optimal grasping location on the source pile based on the current LiDAR observations, and the sampling-based arm path planner proposes a collision-free trajectory to approach the pile, using the OctoMap representation of the environment built from the LiDAR point cloud. The grasping location is appended to the trajectory. 
    \item The full trajectory is subsampled into discrete waypoints, which are tracked by the RL waypoint-following controller.
  \end{enumerate}
  
  \item \textbf{Grasping Material}
  
    The grasping controller lowers the arm until the gripper makes full contact with the material pile, closes the gripper, and lifts up the arm in preparation for dumping.
  
  \item \textbf{Approaching the target and dump material}
  \begin{enumerate}
    \item The arm path planner plans a collision-free trajectory to the user-supplied dumping location.
    \item If operating in open space, the RL throwing controller can be trained without waypoint-following constraints, thus approaching the target location with the fastest path but without safety guarantees. Otherwise, the collision-free trajectory is subsampled into waypoints, which are tracked by the constrained RL throwing policy.
    \item Upon reaching the target location, the RL throwing policy commands the gripper to open and releases the material.
  \end{enumerate}
\end{enumerate}

The transition between stages and selection of the active controller within each cycle is managed by the state machine. Despite the grasping and dumping piles locations being fixed for the entire task execution, the path planning process is run on the updated map every transition, in order to account for sudden large modifications of the scenario.

\subsection{RL Attack-Point Planner}\label{sec:rl_grasp}
For an efficient material-handling workflow, an optimal sequence of gripper attack points must be selected, with the goal of maximizing the volume of transferred material per scoop and thereby minimizing the total number of required grasp-and-dump cycles. In each cycle, the planner provides the optimal 3D end-effector location for grasping the bulk material. Additionally, the planner has to be robust to noisy measurements from the LiDAR observations.
Therefore, we propose an \ac{RL} attack point planner, explicitly designed to efficiently and robustly guide the removal of bulk material. Our planner operates directly on LiDAR point clouds. A simple interpolation, such as nearest-neighbor or cubic interpolation on a filtered point cloud is sufficient to process the sensor measurements into policy inputs. This is made possible by training the policy under observation noise, making the planner robust to various disturbances during real-world deployment.

\subsubsection{Simulation Environment}\label{sec:attack_sim_env}

\begin{figure}
    \begin{minipage}{0.95\columnwidth}
    \centering
      \includegraphics[trim={0cm 1cm 0cm 3cm}, clip, width=\textwidth]{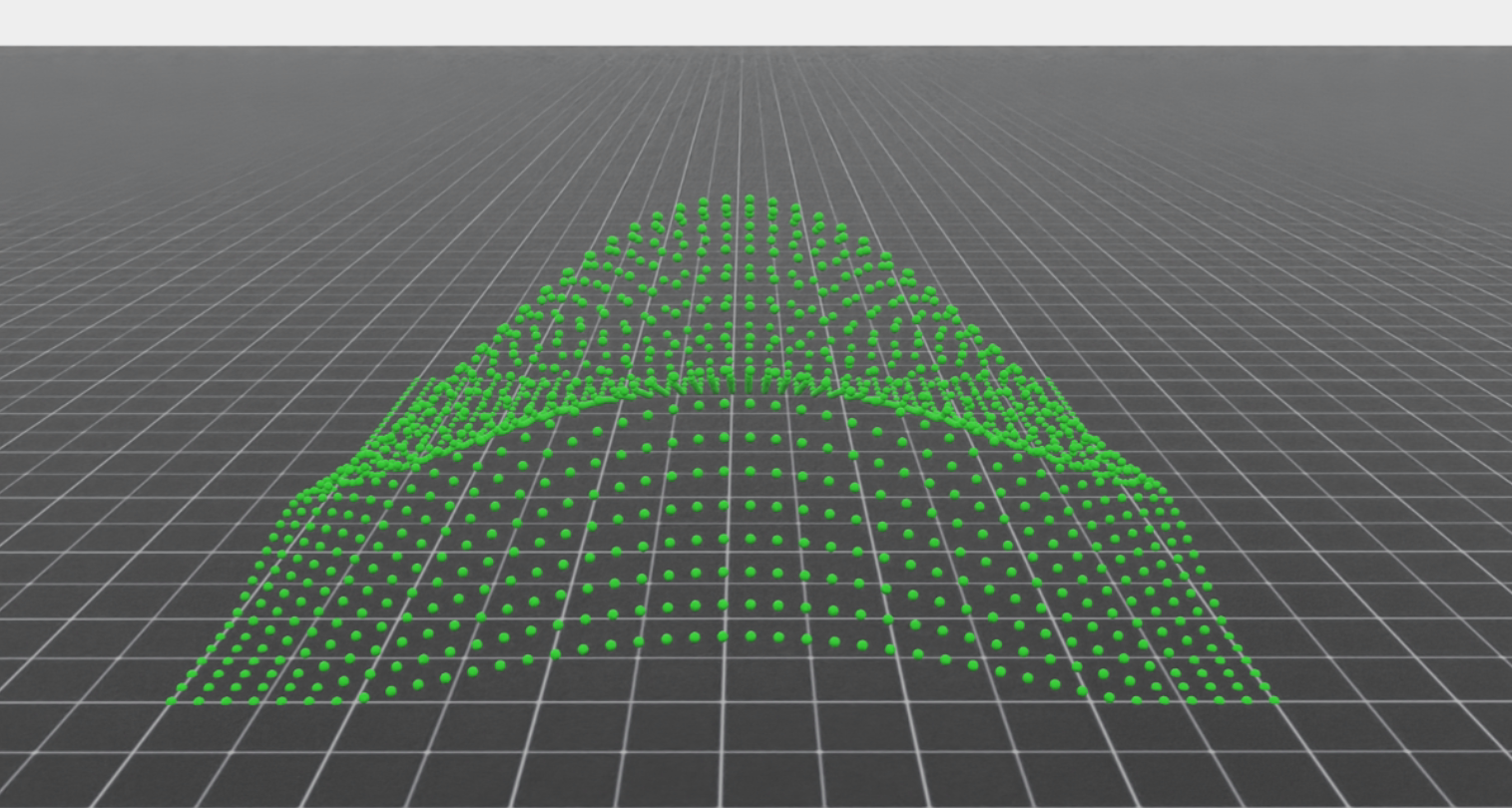}
    \end{minipage}
    
\par\vspace{0.5\columnsep}

    \begin{minipage}{0.95\columnwidth}
    \centering
        \includegraphics[trim={0cm 1cm 0cm 0.8cm}, clip, width=\textwidth]{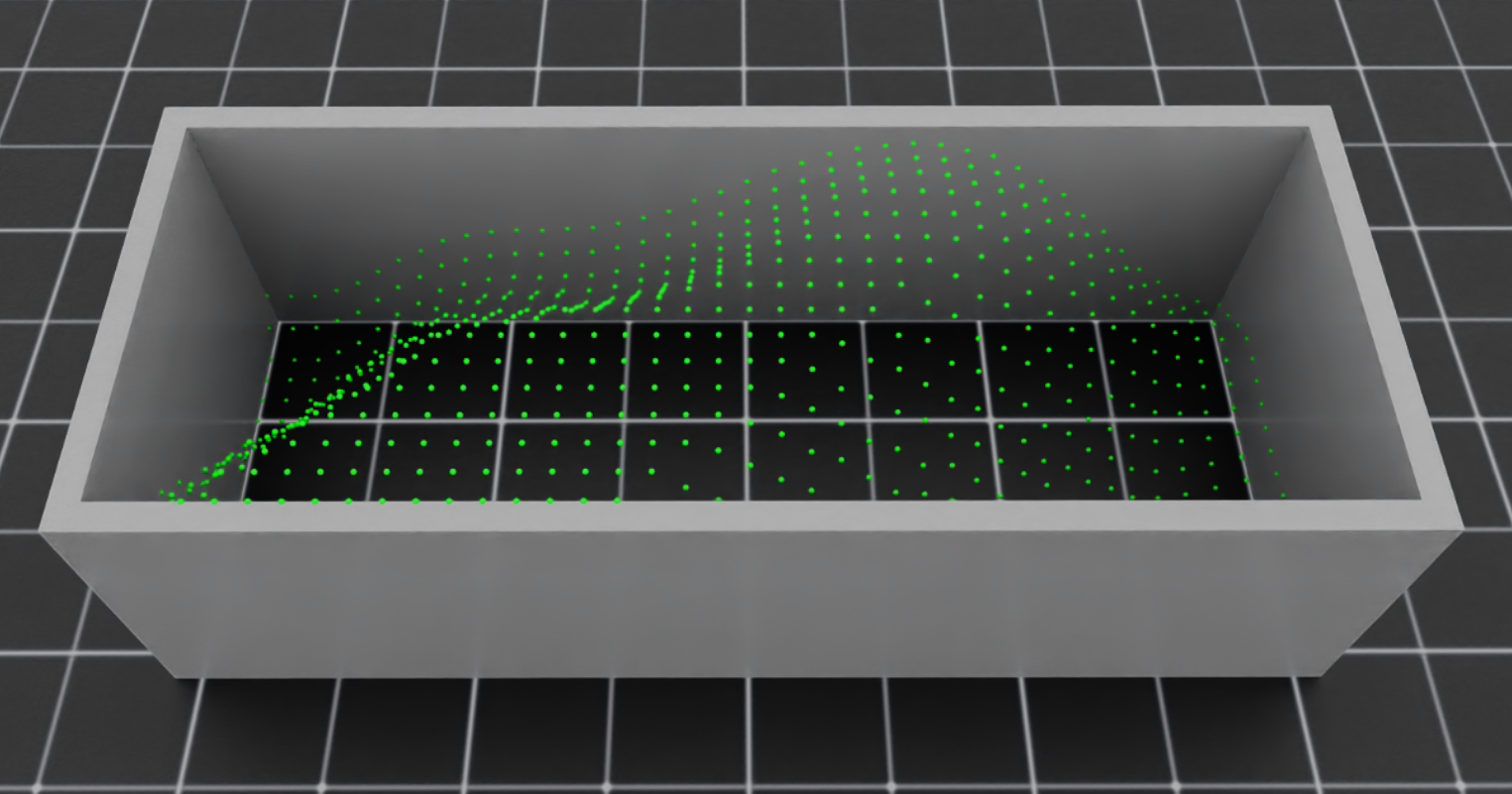}
    \end{minipage}
\caption{Simulated soil for the \ac{RL} attack point planner. Our flexible simulation allows for an easy adaptation to various geometries, such as an unconstrained pile on the ground (top) and bulk material in a container (bottom). The profiles shown are sampled at the beginning of each episode based on 2D Gaussian distributions.}

\label{fig:soil_model}
\end{figure}

\begin{figure}
    \begin{minipage}{0.49\columnwidth}
        \includegraphics[trim={2cm 0cm 2cm 6.25cm}, clip, width=\columnwidth, height=3cm]{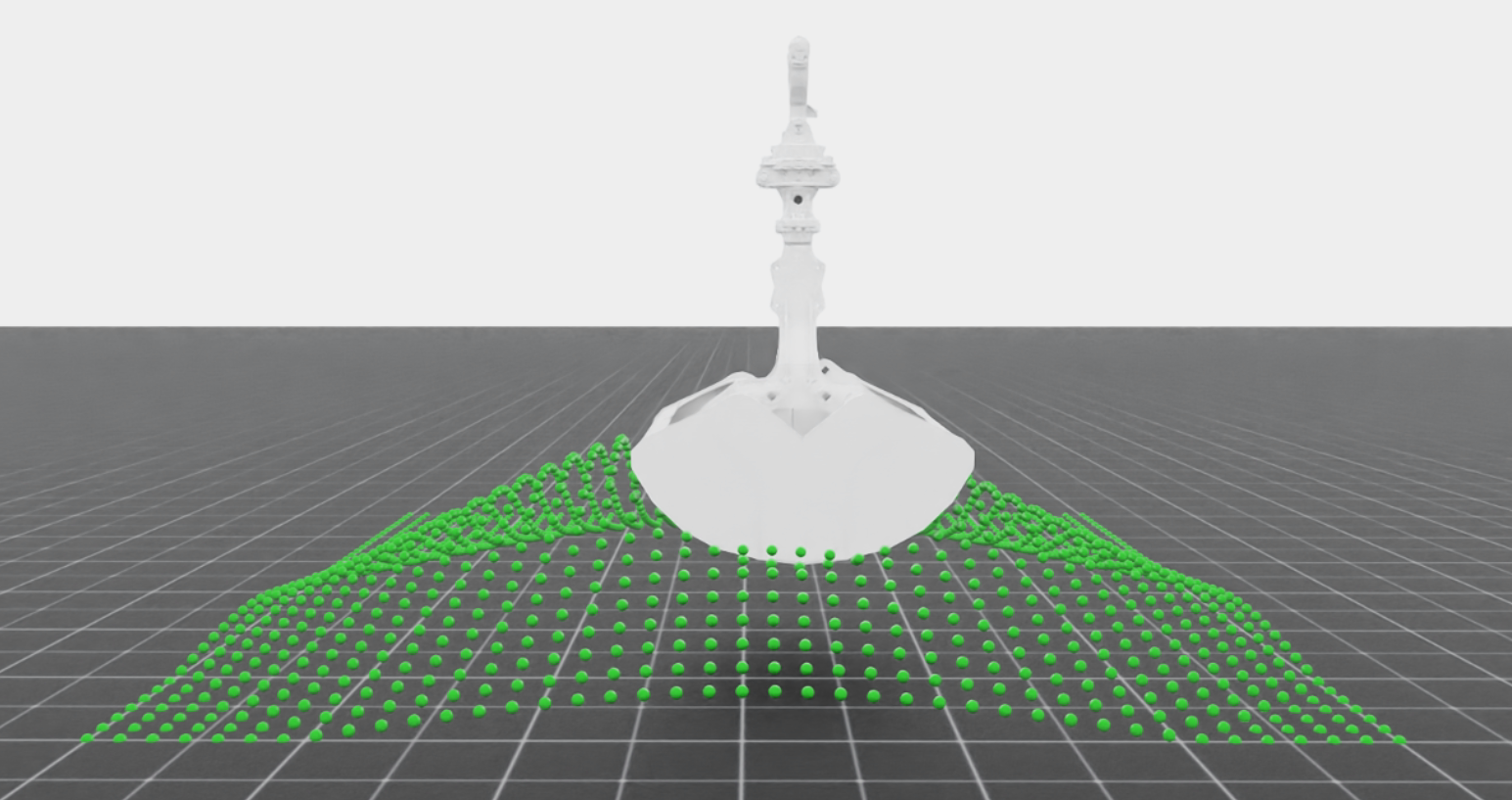}
    \end{minipage}
    \hfill
    \begin{minipage}{0.49\columnwidth}
        \includegraphics[trim={2cm 0cm 2cm 0cm}, clip, width=\columnwidth, height=3cm]{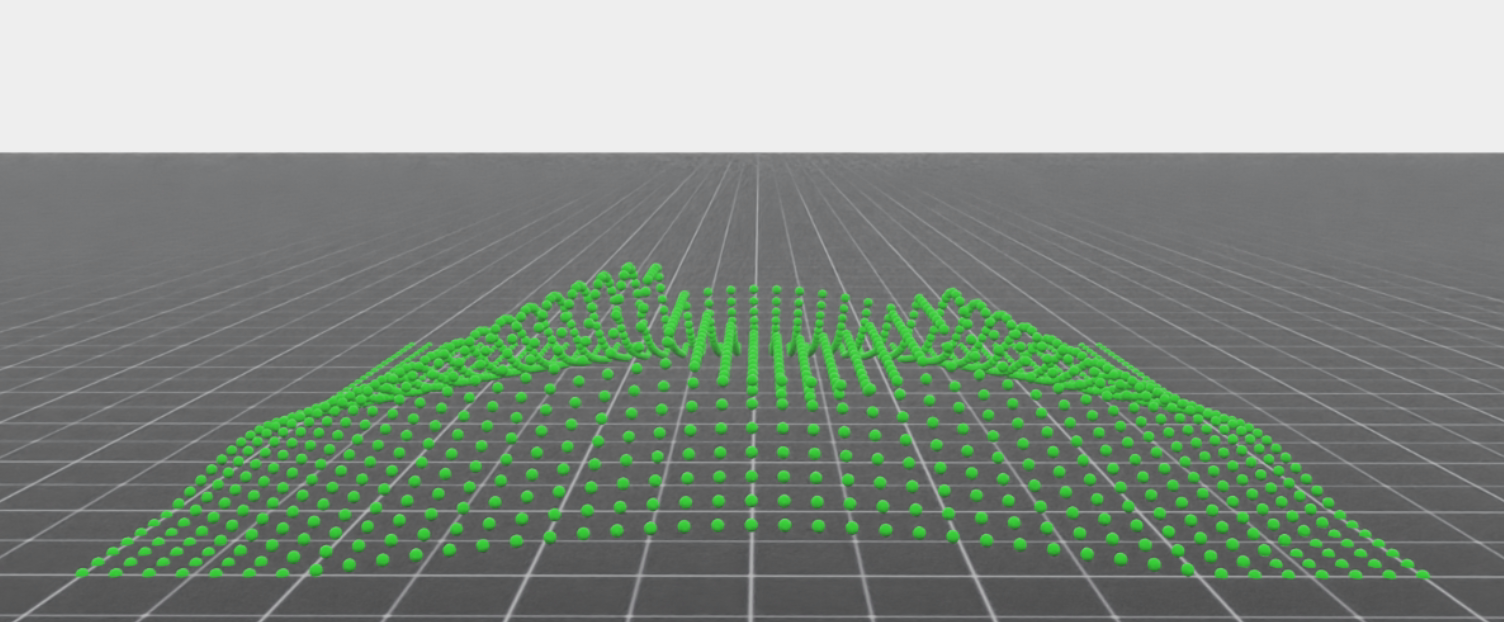}
    \end{minipage}    
    
\caption{Updating the soil profile after scooping in simulation by using the geometry of the closed clamshell bucket.}

\label{fig:scoop_update}
\end{figure}

The 2.5D soil height map is represented by a 2D tensor $\mathbf{h}_{soil}$, as visualized in \cref{fig:soil_model}. Each element $\mathbf{h}_{soil}(i, j)$ represents the height at position $(x, y) = (i \cdot \Delta_{x}, \: j \cdot \Delta_{y})$ on a grid, where $\Delta_{x}$, $\Delta_{y}$ is a user-defined resolution suitable for the task scale. The total size of the grid is also user-defined, which must contain the bulk material to be removed, either in a pile or in an open container. To update $\mathbf{h}_{soil}$ after each scoop, we simulate the footprint left by the scooping operation using the shape of the closed clamshell bucket (\cref{fig:scoop_update}).  

\begin{figure}
    \begin{minipage}{0.49\columnwidth}
        \includegraphics[trim={2cm 0cm 2cm 6.25cm}, clip, width=\columnwidth, height=3cm]{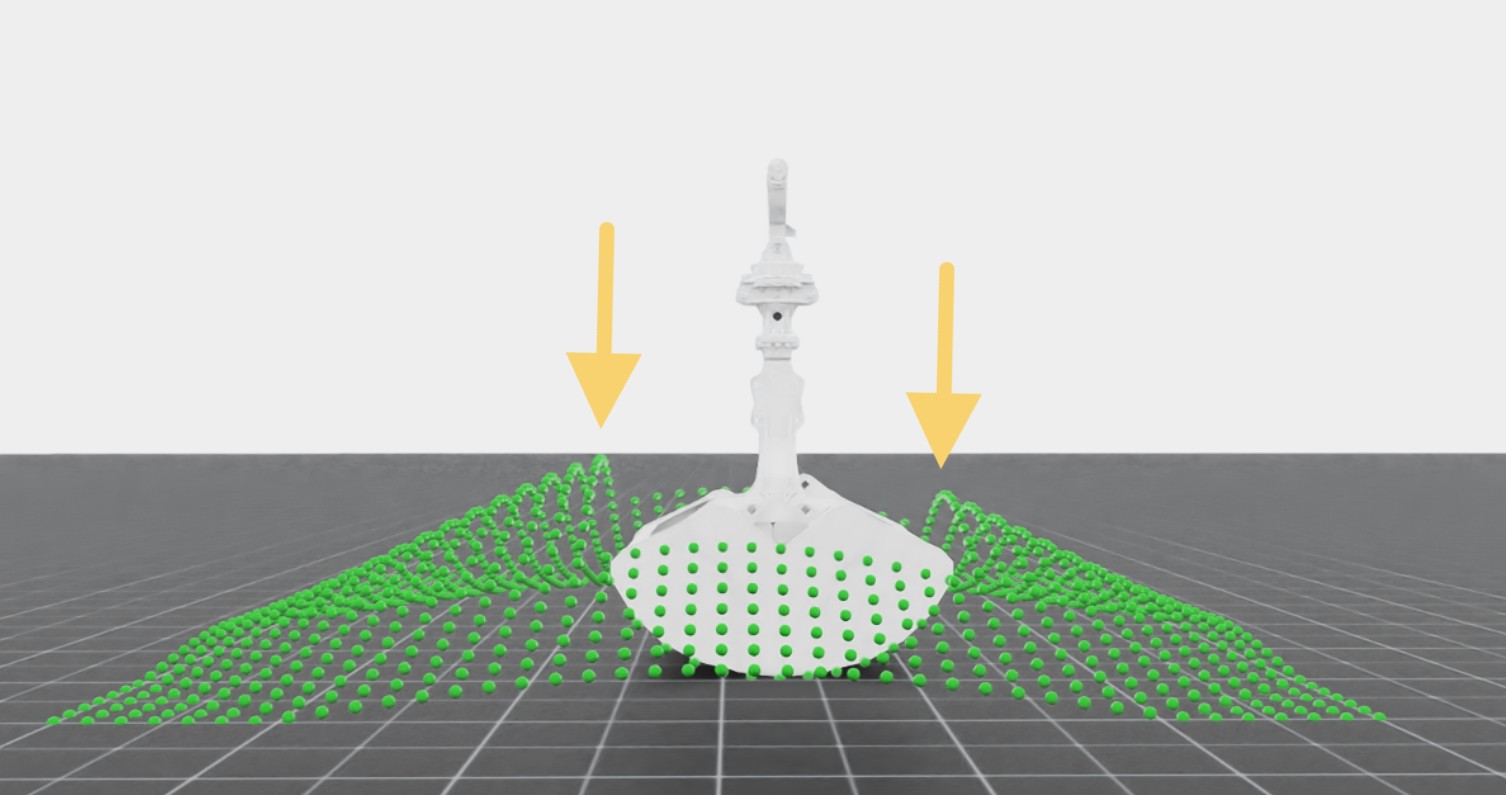}
    \end{minipage}
    \hfill
    \begin{minipage}{0.49\columnwidth}
        \includegraphics[trim={2cm 0cm 2cm 0cm}, clip, width=\columnwidth, height=3cm]{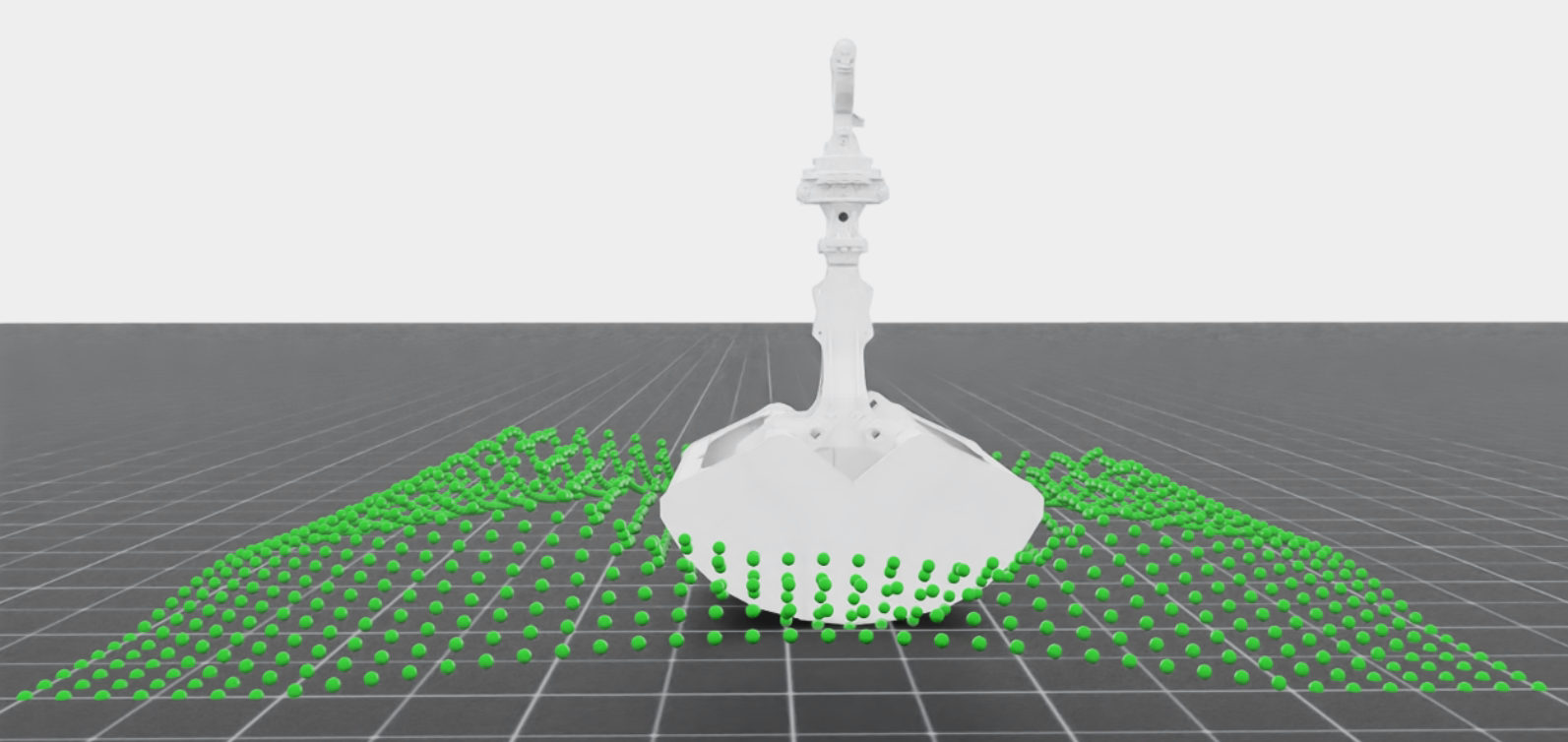}
    \end{minipage}    
    
\caption{Without simulating soil slumping (i.e., the natural collapse of soil under steep slopes), the simulation can become unrealistic, resulting in infinitely steep slopes (left, shown by arrows). A more realistic simulation is achieved by adding the slump update (right) and thus, reducing the sim-to-real gap.}

\label{fig:slump}
\end{figure}

To simulate realistic pile slumps after each scoop, we employ a cellular automaton-based method~\cite{Piegari06CellularAutomaton, Tucker18LatticeGrain}. This algorithm ensures the soil gradient, $\nabla \mathbf{h}_{soil}$, remains below a predefined critical slope, $s_{crit}$.
The simulation iteratively identifies any point $(i, j)$ where the gradient is too steep ($|\nabla \mathbf{h}_{soil}(i, j)| > s_{crit}$) and moves a small volume of material downslope (in the $-\nabla \mathbf{h}_{soil}$ direction). This process repeats until the entire pile is stable (all gradients $\le s_{crit}$) or a maximum number of iterations is reached. The total soil volume is conserved throughout this update, as visualized in \cref{fig:slump}.

The simulation is lightweight and fast. It does not require a physics solver, as all updates are based on kinematics and geometry, and assumes the bucket filling is independent of the soil force acting on the machine. Our hardware experiments validate this assumption. All update operations are parallelized through tensor operations, enabling the simultaneous simulation of multiple environments.

\subsubsection{RL Attack-Point Planner}\label{sec:rl_att_planner}
\paragraph{Task Description}
The goal of the attack point planner is to output 3D attack points $\mathbf{p}=[p^{x}, p^{y}, p^{z}]^\top$ that enable efficient material removal to a specified volume threshold. Unlike mobile wheel loaders, the attack point selection order for a fixed-base handler does not impact the total travel distance, which is constant regardless of the chosen sequence.
Therefore, we disregard transit time between attack and dump targets and optimize for bucket fill ratio and the total number of cycles to clear a target volume. These metrics serve as direct proxies for operational efficiency and energy consumption.

\paragraph{Actions and Observations}
The policy is represented by an \ac{MLP} of the form $\mathbf{p} = \pi_{att}(\mathbf{h}_{soil})$. 
It maps the flattened ${\mathbf{h}_{soil}}$ tensor to a normalized attack point ${\mathbf{p}}$, which is subsequently clamped to the ${\mathbf{h}_{soil}}$ grid boundaries. When using a large workspace and/or a fine grid resolution leads to a large ${\mathbf{h}_{soil}}$ dimension, we apply bi-cubic interpolation on ${\mathbf{h}_{soil}}$ to get coarser observations as policy input. 
We add noise to the observations to account for three noise sources for sim-to-real robustness: \textit{i)} zero-mean Gaussian noise with a standard deviation of \SI{0.1}{\meter}, representing sensor noise; \textit{ii)} drop-out (zeroing) with $10\%$ probability, representing missing points in the LiDAR scans; \textit{iii)} large vertical spikes with $0.5\%$ probability, with distribution $\mathcal{U}(\SI{1}{\meter},\ \SI{3}{\meter})$, representing unfiltered self-occlusions from the machine itself or material dropped from the gripper and detected by the LiDAR during free-fall in the air.

\paragraph{Episode Initialization and Domain Randomization}
Various initialization strategies for $\mathbf{h}_{soil}$ can be applied, depending on the task-specific bulk geometry. When training for the piled material scenario, we sample 2D skew-normal distributions with mean location and shape parameters, resulting in asymmetric piles with a unique peak. For materials in open containers, we sample from Gaussian processes with squared exponential kernels. A sampled $\mathbf{h}_{soil}$ for each environment is visualized in \cref{fig:soil_model}.

\paragraph{Termination Conditions}
\label{sec:rl_grasp_terms}
An episode is terminated successfully when the remaining material volume is reduced below a given threshold; we select \SI{0.15}{\cubic\meter} of soil volume per \si{\square\meter} of grid area. If the policy exceeds the maximum number of steps without terminating successfully, the episode times out. We set the maximum timestep to 150, which is sufficient to complete material removal from the workspace, and provides the policy with an adequate timesteps amount to explore until positive termination.

\paragraph{Reward}
At each timestep $k$, given the action $\mathbf{p}_k = [p^{x}_k, p^{y}_k, p^{z}_k]^\top$, the reward is defined as: 
\begin{equation}\label{eq:grasp_reward}
    \begin{aligned}
        R_k & = r_k^{\text{ssv}} + r_k^{\text{neg\_z}} + r_k^{\text{living}} + r_k^{\text{pos\_term}} + r_k^{\text{neg\_term}}.
    \end{aligned}
\end{equation}

\noindent Specifically, the three main rewards are defined as:
\begin{equation*}
    \begin{aligned}
        & r_k^{\text{ssv}} \propto \exp{(\text{ssv}_k)}, \\ 
        & r_k^{\text{neg\_z}} \propto \begin{cases}
           -||z_{\text{feasible}}-p^{z}_{k}||^2 & \text{if} \quad p^{z}_{k} < z_{\text{feasible}}  \\
            0 & \text{otherwise}
        \end{cases}, \\
        & r_k^{\text{living}} \propto -k\\
    \end{aligned}
\end{equation*}

\noindent where $\text{ssv}_k$ denotes the scooped soil volume at step $k$. To prevent the policy from proposing infeasibly deep attack points, $z_{\text{feasible}}$ is the maximum depth that the gripper can physically reach at the location $(p^{x}_k, p^{y}_k)$ on the pile. The living cost $r_k^{\text{living}}$ increases with episode length, encouraging the policy to complete the task as efficiently as possible.
The terms $r_{k}^{\text{pos\_term}}$ and $r_{k}^{\text{neg\_term}}$ are associated with the previously defined positive and negative termination conditions.

If more tailored strategies are needed, such as starting scooping from a designated location and sequentially clearing material in columns along the $y$-axis before progressing in the positive $x$-direction, the following rewards can be included:
\begin{equation*}
    \begin{aligned}
        & r_0^{\text{init}} \propto -||(p^x_0, p^y_0) - (x_{\text{init}}, y_{\text{init}})||^2,\\ 
        & r_k^{\text{x\_back}} \propto \begin{cases}
           -||p^{x}_{k-1}-p^x_{k}|| & \text{if} \quad p^x_k < p^x_{k-1}  \\
            0 & \text{otherwise}
        \end{cases},
    \end{aligned}
\end{equation*}
\noindent where $r_0^{\text{init}}$ encourages the first attack point to be $(x_{\text{init}}, y_{\text{init}})$, and $r_k^{\text{x\_back}}$ discourages going back in $x$-direction, thus resulting in a policy that plans grasping in $y$-column order. This tailored behavior is essential when grasping from containers and cargo ships, since the material must be removed along the length of the cargo ship as it moves relative to the material handler. This use case is not part of this work's hardware experiments, but has only been evaluated in simulation.

\paragraph{Training Details}
\label{sec:att_pt_training_details}
We train the policy with \ac{PPO}~\cite{Schulman17ProximalPolicy}, with $N_{upd}=10{,}000$ and $N_{env}=4096$. The \ac{MLP} has hidden dimensions $[256, 256, 128, 64]$ with ReLU activation. For the pile setting, we use a $6 \times 6$ \si{\meter} grid with a resolution of \SI{0.15}{\meter}. For the container environment, we use a container of size $4 \times 1.5 \times 1.2$ \si{\meter} with a resolution of \SI{0.05}{\meter}. Different resolutions are chosen based on the clamshell gripper sizes as well as the different required precision.
\subsection{Arm Path Planner}\label{sec:planner}

\begin{figure*}
    \centering
    \begin{minipage}{\columnwidth}
        \includegraphics[trim={5cm 5cm 5cm 0cm},clip, width=\textwidth]{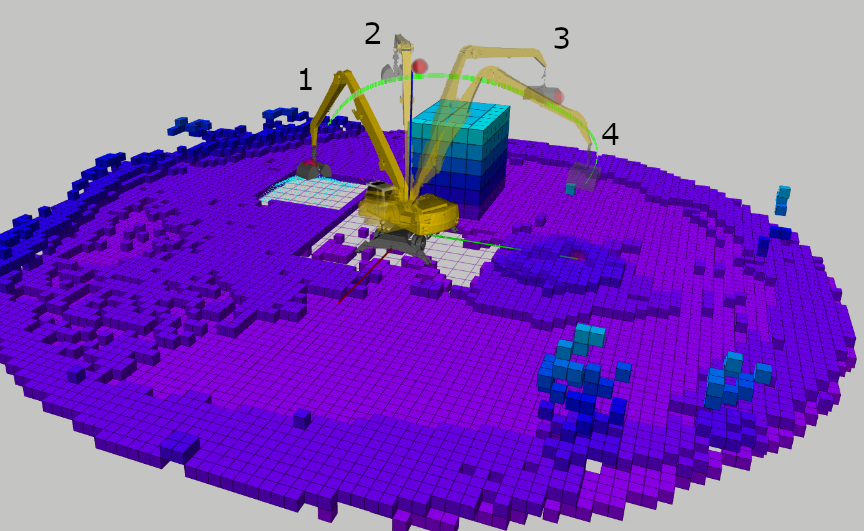}
    \end{minipage}
    \begin{minipage}{\columnwidth}
        \includegraphics[trim={4cm 5cm 6cm 0cm},clip, width=\textwidth]{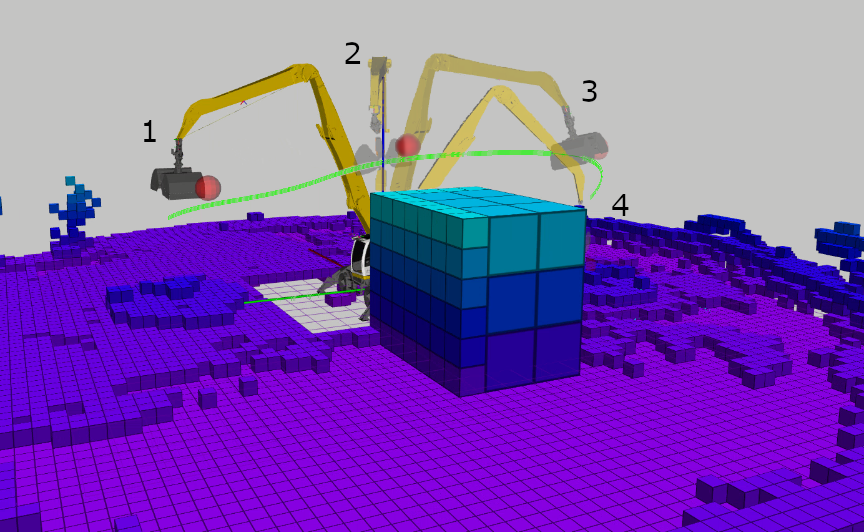}
    \end{minipage}
    
    \caption{Grasping and dumping trajectories generated using the RRT*-based planner. The smoothed B-spline paths are shown in green, and the subsampled tracking waypoints are shown in red. The environment is represented as an OctoMap constructed from the filtered point cloud. To prevent unnecessary collision detection, specific OctoMap blocks are masked: (1) the material pile (left) is removed to allow the trajectory to approach the grasping point, and (2) regions around the machine’s upper carriage are excluded, as self-collisions are already handled through joint-space constraints during planning. Virtual obstacles, such as the virtual wall above, can be added by inserting OctoMap blocks, enabling easy testing of the pipeline or to limit the pipeline's access to restricted zones in the workspace.}
    \label{fig:octomap}
\end{figure*}

We employ a standalone, sampling-based arm path planner to generate collision-free reference trajectories under the assumption of a fixed base. These trajectories are tracked by the \ac{RL} policy, rather than integrating perception and planning into end-to-end \ac{RL} training. Such a modular design simplifies training by reducing task complexity, improves pipeline interpretability for safer execution, and offers greater flexibility for behavior adjustment at deployment time.

To enable environmental perception for the planner, we construct OctoMaps~\cite{Hornung13OctoMapEfficient}, an octree-based data structure for representing 3D occupancy grids (\cref{fig:octomap}), using filtered LiDAR point cloud data. The map resolution is set to \SI{0.5}{\meter} to balance computational efficiency with sufficient spatial detail compared to the machine scale. Using this full 3D representation instead of a 2.5D heightmap is critical for material handling tasks, especially in constrained indoor environments such as recycling facilities, where collisions could take place with overhead structures and ceilings.

To generate the collision-free path, we use the asymptotically optimal \ac{RRT*}~\cite{Karaman11SamplingbasedAlgorithms} algorithm implemented in the OMPL library~\cite{Sucan12OpenMotion}. The planning state space consists of the slew, boom, and stick joints represented in $SO(2)$. Differently from~\cite{jebellat2024motion}, the unactuated gripper joints are excluded from the planning model. Instead, we adopt a simpler solution by inflating the gripper’s collision geometry to account for possible tool oscillations, which are in turn mitigated through \ac{RL} policy training. The planner minimizes joint-space path length, resulting in trajectories that naturally follow cylindrical segments. All sampled paths respect joint limits and avoid self-collisions. In simulation, \ac{RRT*} achieves a success rate above $99\%$ and converges to near-optimal path lengths within \SI{1}{\second} of optimization. For deployment, the optimization time is extended to \SI{2}{\second}. 
 
The resulting path is then smoothed with a B-spline, with continuous collision checking performed to ensure safety with respect to workspace obstacles. This smoothed trajectory is subsequently subsampled into discrete waypoints, which are tracked by the downstream \ac{RL} controllers. The waypoint sampling resolution is a tunable parameter that depends on the trajectory's curvature. Because the \ac{RL} policy interpolates between waypoints during execution, a sufficiently high waypoint density is required to ensure the interpolated trajectory remains close to the original collision-free B-spline. \Cref{fig:octomap} illustrates representative grasping and throwing trajectories executed on the physical system, with a virtual obstacle added to the workspace to demonstrate the system's collision avoidance capabilities.

\subsection{RL Waypoint-Following \& Throwing Controllers}\label{sec:rl_control}
Autonomous control of material handlers involves several challenges, such as delays and nonlinear dynamics of the hydraulic actuators, the coordination required to dampen oscillations of the free-hanging gripper, and the need to maintain safety in unstructured environments. To address these challenges, we use \ac{RL} to train two policies: a waypoint-following controller for approaching the pile attack point with an empty gripper (\cref{sec:waypoint_rl}), and a throwing controller for dumping material at a target location (\cref{sec:throw_rl}). By training separate controllers instead of a unified one for the whole grasp-and-dump cycle, we reduce the training complexity and avoid the need to simulate the intricate interactions between the machine and the material pile, which are instead handled by the grasping controller introduced in \cref{sec:grasp_control}.

The main training objective of the waypoint-following controller is to track the input waypoints, while minimizing the gripper oscillations; the throwing controller has the additional goal of accurately releasing material at the target location. When operating in waypoint-following mode, both policies are subject to \textit{tube constraints}, which encourage the gripper to stay within cylindrical regions, or tubes, connecting consecutive waypoints. These constraints help minimize deviation and provide safety guarantees during operation.
Notably, the throwing controller can also be trained without waypoint inputs, allowing the policy to approach the release point without restrictions when operating in free space.

\subsubsection{Simulation Environment}\label{sec:sim_dynamics}
Unlike the boom and stick joints, the rotational dynamics of the slew joint is difficult to model analytically. To address this, we adopt a hybrid simulation approach introduced in~\cite{Spinelli24ReinforcementLearning}, where the slew joint is modeled using data-driven, learning-based techniques, while the rest of the system relies on first-principles modeling. 

To model the dynamics of the slew joint, we train two \acp{MLP} using real machine data. The first \ac{MLP} $\mathcal{F}_p$ predicts the pressures in the left and right chambers $\mathbf{P} = [P_l, P_r]^\top$ given past inputs and system states:

\begin{equation}
\label{eq:NN_formulation_press}
    \begin{aligned}
        \mathbf{P}_k = \mathcal{F}_p\Big(
        &u_{[k-9:k]}, \mathbf{P}_{[k-10:k-1]}, \omega_{[k-10:k-1]}, I_{z,k-1} \Big),
    \end{aligned}
\end{equation}
where $u$ is the control input, $\omega$ is the angular velocity of the cabin, and $I_z$ is the configuration-dependent inertia around the rotation axis; it is computed from the arm position and the nominal link weights, approximating the tool and its load as a single point mass. The notation $\cdot_{[i:j]}$ denotes the discrete time series of the given quantity from step $i$ to step $j$, inclusive.  
The second \ac{MLP}, $\mathcal{F}_\omega$, using the pressure predictions as an input, predicts the joint velocity:
\begin{equation}
\label{eq:NN_formulation_vel}
    \begin{aligned}
        \omega_{k} = \mathcal{F}_\omega \Big( 
        & u_{[k-9:k]}, \mathbf{P}_{[k-9:k]}, \omega_{[k-10:k-1]}, I_{z,k-1} \Big) .
    \end{aligned}
\end{equation} 
As observed in~\cite{Spinelli24ReinforcementLearning}, this two-stage architecture, which separates pressure prediction from velocity prediction, effectively decouples the hydraulic and mechanical dynamics of the actuator, resulting in better modeling accuracy.

For the other joints, simple analytical models that require minimal machine-specific data are used to simulate their dynamics. For the boom and stick joints, we use a first-order system with delay to approximate the velocity-tracking behavior of the low-level arm controller:
\begin{equation}\label{eq:arm_simulated}
    \dot{q}_{k} = \dot{q}_{k-1} + K\big( \dot{q}^{\text{~ref}}_{k - d} - \dot{q}_{k-1} \big),
\end{equation}
where $\dot{q}$ is the joint velocity, $\dot{q}^{\text{~ref}}$ is the velocity reference, $d$ represents the delay, and $K$ is a time constant identified from the rising time of the low-level joint controller (\cref{sec:arm_vel_controller}).

\begin{figure}
    \centering
    \begin{minipage}{0.5\columnwidth}
    \includegraphics[trim={5.5cm 6cm 8cm 10.5cm},clip, width=\textwidth]{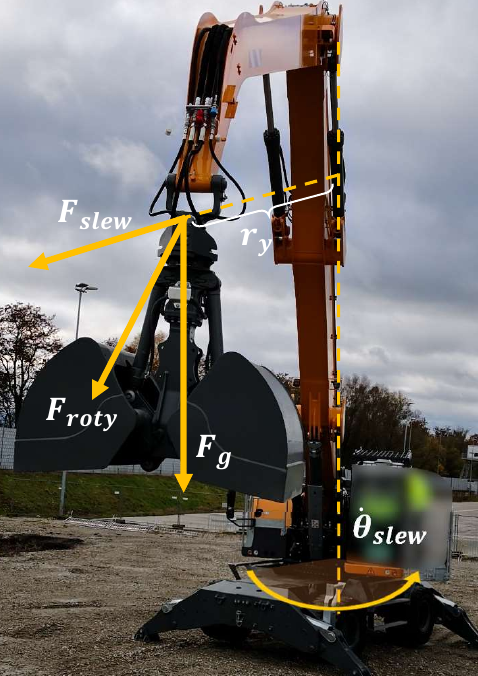}
    \end{minipage}\hfill
    \begin{minipage}{0.5\columnwidth}
    \includegraphics[trim={8cm 8cm 7.5cm 11.2cm},clip,width=.9\textwidth]{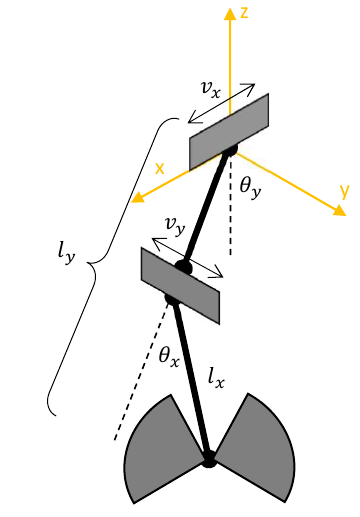}
    \end{minipage}
    \caption{The tool is modeled as a double pendulum with linearly oscillating support. The relevant forces are shown in the left figure, and the approximations for each \ac{DoF} are shown on the right.
    }
    \label{fig:tool_all}
\end{figure}

The gripper oscillation dynamics is modeled as a double pendulum with decoupled $x$ and $y$ rotations. We account for the Euler and centrifugal forces generated by the moving reference frame, but neglect the Coriolis component. 
Additionally, we include a dissipative term via Rayleigh's dissipation function~\cite{Minguzzi14RayleighsDissipation}. 
The motion is mathematically formulated as:
\begin{equation}\label{eq:tool_y}
    \begin{aligned}
    \dot{\theta}_{y,k+1} &= \Bigg(\bigg(\frac{v_{x,k+1}-v_{x,k}}{\Delta t}\cos \theta_{y,k} -\underbrace{g\sin \theta_{y,k}}_{F_g} \\ &- \underbrace{r_y\dot{\theta}_{slew,k}^2}_{F_{slew}}\bigg)/l_y - \underbrace{b_{fy}\dot{\theta}_{y,k}}_\text{dissipation}\Bigg)\Delta t + \dot{\theta}_{y,k},
    \end{aligned}
\end{equation}
\begin{equation}\label{eq:tool_x}
    \begin{aligned}
         \dot{\theta}_{x,k+1} &= \Bigg(\bigg(-\frac{v_{y,k+1}-v_{y,k}}{\Delta t}\cos \theta_{x,k} -\big(\underbrace{g\cos \theta_{y,k}}_{F_g} \\ &+ \underbrace{l_y\dot{\theta}_{y,k}^2}_{F_{roty}}\big)\sin \theta_{x,k}\bigg)/l_x - \underbrace{b_{fx}\dot{\theta}_{x,k}}_\text{dissipation}\Bigg)\Delta t + \dot{\theta}_{x,k} .
    \end{aligned}
\end{equation}
Here, $\theta_{x}$ and $\theta_{y}$ denote the angles between the two unactuated joints and the principal frame axes, $v_{x}$ and $v_{y}$ are the linear velocities of the tool attachment points, $l_{x}$ and $l_{y}$ are the corresponding tool lengths, $r_y$ is the distance between the tool and the slew rotation axis, $g$ is the gravity constant, and $b_{fx}$ and $b_{fy}$ are the dissipative coefficients. An illustration of the model is shown in \cref{fig:tool_all}.

Among the various gripper–material interactions, only the throwing action is required for training the policies, as the grasping from the pile is executed by the grasping controller. The granular behavior of the bulk material during release is approximated using three discrete loads, $L_1$, $L_2$, and $L_3$. Upon receiving an open command, the first load is released after a delay of approximately \SI{0.3}{\second}, based on real-robot measurements. Subsequent loads are spawned at \SI{0.5}{\second} intervals to emulate continuous material flow.  Each load follows a parabolic trajectory, initialized with the gripper’s velocity at release time. A visualization is provided in \cref{fig:throw_sim}.  
While discrete element simulations~\cite{Backman21ContinuousControl, Aoshima24ExaminingSimulationtoreality} offer higher fidelity, we believe their computational cost outweighs the benefits for contact-free throwing. Given the focus of our training pipeline on airborne material release, we prioritize the computational efficiency of an analytical model, and use domain randomization~\cite{Zhao20SimtoRealTransfer} to generalize across different materials behaviors and enhance robustness during deployment on hardware.  
This simplified material model is employed only for the throwing policy training.

\subsubsection{\ac{RL} Waypoint-Following Controller}\label{sec:waypoint_rl}

\paragraph{Task Description}

The waypoint-following controller is trained for two concurrent objectives: \textit{i)} controlling the joints so that the gripper tracks a sequence of waypoints, and \textit{ii)} minimizing gripper oscillations both during motion and upon reaching the final waypoint. The second objective is essential for safe machine operation and to ensure a smooth transition to the subsequent grasping phase.

\begin{figure}
    \centering
    \includegraphics[trim={1cm 1cm 5cm 1cm},clip, width=\columnwidth]{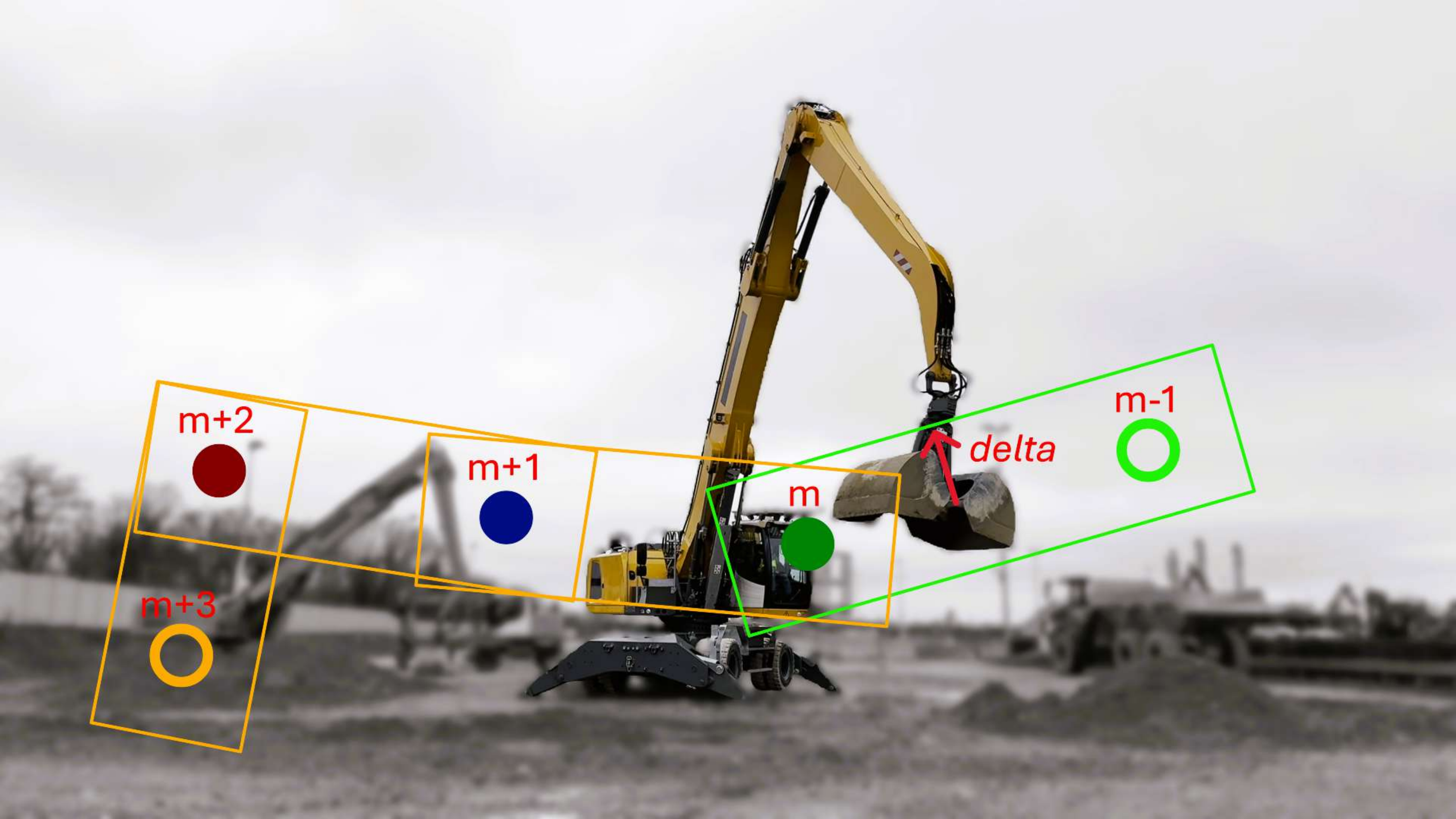}
    \caption{During training, the next three waypoints out of the full sequence of five are provided as observations to the controller. They are visualized in green, blue, and red, respectively. A tube linearly interpolates between each pair of waypoints, and the distance to the closest tube edge is provided as an observation.}
    \label{fig:waypoint_sim}
\end{figure}

Each training episode lasts \SI{15}{\second} and involves navigating through five randomly sampled waypoints, provided to the policy in a receding horizon fashion with a horizon length of three. The policy is trained to satisfy a tube-shaped position constraint. As illustrated in \cref{fig:waypoint_sim}, a cylindrical volume of radius \(r_{\text{tube}}\) connects each waypoint pair, and the gripper’s motion is limited within this region. This encourages shortest-path behavior while maintaining a user-defined safety margin, thereby enhancing predictability during operation. Upon reaching the final waypoint, the controller must additionally minimize overshoot and residual motion, and stabilize the gripper with minimal end-effector position error. 
By tuning \(r_{\text{tube}}\), policies can be trained to exhibit varying trade-offs between speed and safety. Although the radius could be provided as a time-varying input, for simplicity we instead train multiple policies with fixed \(r_{\text{tube}}\) and switch between them based on task requirements.

\paragraph{Actions and Observations}
\label{sec:actobs_waypoint_rl}

The waypoint-following policy is an \ac{MLP} of the form:
\begin{equation}\label{eq:waypoint_policy}
    \begin{aligned}
        \mathbf{u}_{k} & = \Big[u_{\text{slew}, k},\dot{q}^{\text{~ref}}_{\text{~boom}, k},\dot{q}^{\text{~ref}}_{\text{~stick}, k}
        \Big]^\top \\
        & = \pi_{wp} \Big( \mathbf{u}_{[k-H: k-1]}, \dot{\mathbf{q}}_{[k-H: k-1]}, \mathbf{q}_{k-1}, I_{z, k-1}, \mathbf{w}_{[m:m+2]}, \\ & \mathbbm{1}{\{m=M\}}, \delta_{\text{tube}, k-1} \Big), \\
    \end{aligned}
\end{equation}
where
\begin{equation*}
    \begin{aligned}
        &\dot{\mathbf{q}} = \big[\dot{q}_{\text{slew}}, \dot{q}_{\text{boom}}, \dot{q}_{\text{stick}}, \dot{q}_{\text{tool},x}, \dot{q}_{\text{tool},y} \big]^\top, \\
        &\mathbf{q} = \big[q_{\text{boom}}, q_{\text{stick}}, q_{\text{tool},x}, q_{\text{tool},y} \big]^\top, \\
        &\mathbf{w} = [r^{wp}, \phi^{wp}, z^{wp}]^\top.
    \end{aligned}
\end{equation*}

By integrating the slew \ac{NN} model into the simulation, the policy is trained to output direct a slew joystick command $u_{\text{slew}}$. It is constrained to $[-85, 85]\%$ of the actuation range to align with the data distribution from real operators used in training, and thus to generate accurate dynamics in simulation. For the arm joints, the velocity references $\dot{q}^{\text{~ref}}_{\text{~boom}}$ and $\dot{q}^{\text{~ref}}_{\text{~stick}}$ are clipped to $[-0.2, 0.2]$~\si{\radian\per\second}.

The observation vector definition is inspired by previous work~\cite{Egli22GeneralApproach, Nan24LearningAdaptive, Spinelli24ReinforcementLearning}.
A history of actions $\mathbf{u}$ and velocities $\dot{\mathbf{q}}$ of length $H$ is provided for each joint, with different lengths based on the joint simulation principle: five for slew, three for boom and stick, and three for unactuated tool joints. The longer history for the slew joint leverages its more accurate dynamic model, enabling the agent to estimate the state evolution without being affected by a large sim-to-real gap, which increases with history length. For the boom, stick, and tool joints, a single position measurement $\mathbf{q}$ is included. No explicit position is provided for the slew joint, as its state is implicitly represented through the target waypoint definition.  
The configuration inertia $I_z$ is the same as in \cref{eq:NN_formulation_vel}.  
The waypoint targets are expressed in cylindrical coordinates $\mathbf{w} = [r, \phi, z]^\top$.  
Assuming rotational symmetry and position invariance, the angle \(\phi\) is defined as the angular error between the current slew position and the waypoint. The index of the next waypoint is denoted by the subscript \(\cdot_m\), and $M=5$ denotes the total number of waypoints as well as the index of the last one in each episode. The policy observes the next three waypoints, which are updated in a receding-horizon manner. Each of them is considered reached when the gripper reduces the distance from it below $r_{tube}$. Choosing a horizon of three waypoints balances the tradeoff between providing the policy with farther lookahead and limiting the complexity of the \ac{RL} training. To emphasize the robot’s approach to the final waypoint, a boolean flag \(\mathbbm{1}{\{m=M\}}\) is introduced. By observing the waypoints through this moving window, the policy is able to handle arbitrarily long sequences during deployment; thus, the flag is required to plan for a smooth stop to minimize gripper oscillations at the last target. 
The observation \(\delta_{\text{tube}}\) represents the signed distance between the gripper and the closest edge of the tube constraint. We consider a tube of radius \(r_{\text{tube}}\) connecting the previous waypoint \(p_1\) and the next waypoint \(p_2\), and extend it by \(r_{\text{tube}}\) beyond both waypoints for numerical stability (see \cref{fig:waypoint_sim}). Let \(E\) denote the position of the tool \ac{CoM}. Then, the distance \(\delta_{\text{tube}}\) from the tube boundary is given by:
\begin{equation}\label{eq:tube_distance}
    \delta_{\text{tube}} = \max\left( -\frac{r_{\text{tube}}}{2}, r_{\text{tube}} - \frac{\left| \overrightarrow{p_1 E} \times \overrightarrow{p_1 p_2} \right|}{\left| \overrightarrow{p_1 p_2} \right|}\  \right),
\end{equation}
thus \(\delta_{\text{tube}}\) has range \(\left[-r_\text{tube}/2, r_{\text{tube}}\right]\). This value allows the agent to correlate the state with the tube constraint reward during training and serves as feedback on the current adherence to the constraint during deployment.

The observations are perturbed with uniform noise proportional to their normalization factors, following literature approaches shown to improve robustness in real-world deployment~\cite{Hwangbo19LearningAgile, Egli22GeneralApproach}.  
This perturbation is not applied to the waypoints, the tube distance, or the boolean last-target flag.

\paragraph{Episode Initialization and Domain Randomization}

At the beginning of each training episode, the environment is initialized by sampling initial joint positions from collision-free configurations and setting the gripper load randomly within the range $[0, 3]\si{\tonne}$. Additionally, the arm controller's time constant $K$, delay $d$, tool lengths $l_{x}, l_{y}$, and friction parameters $b_{fx}, b_{fy}$ are all sampled from uniform distributions.
The target trajectory consists of $M=5$ waypoints, generated according to the following rules:
\begin{enumerate}
    \item In \(2/3\) of the episodes, the same slew rotation direction is used for all waypoints. In the remaining \(1/3\), the rotation direction is resampled for each new waypoint.  
    This setup enables the policy to explore fast-paced motions with consistent slew rotation, while also improving robustness to abrupt directional changes.
    
    \item Each new waypoint is sampled such that its cylindrical-coordinate Manhattan distance from the previous one is bounded. This constraint mimics the continuous paths that will be provided during deployment.
    
    \item In \(1/2\) of the episodes, the final waypoint is placed below the second-to-last one, with a height \(z \in [1, 3]\,\si{\meter}\).  
    This encourages improved tracking behavior when approaching the bulk material pile.
\end{enumerate}

\paragraph{Termination Conditions}
The episode is terminated with a negative reward if the gripper comes too close to the ground or the cabin.  
This design promotes a safe behavior but does not guarantee collision-free motion, as the policy lacks perception of external obstacles.

\paragraph{Reward}

\begin{figure}
    \centering
    \includegraphics[trim={1cm 0cm 1cm 1cm},clip, width=\columnwidth]{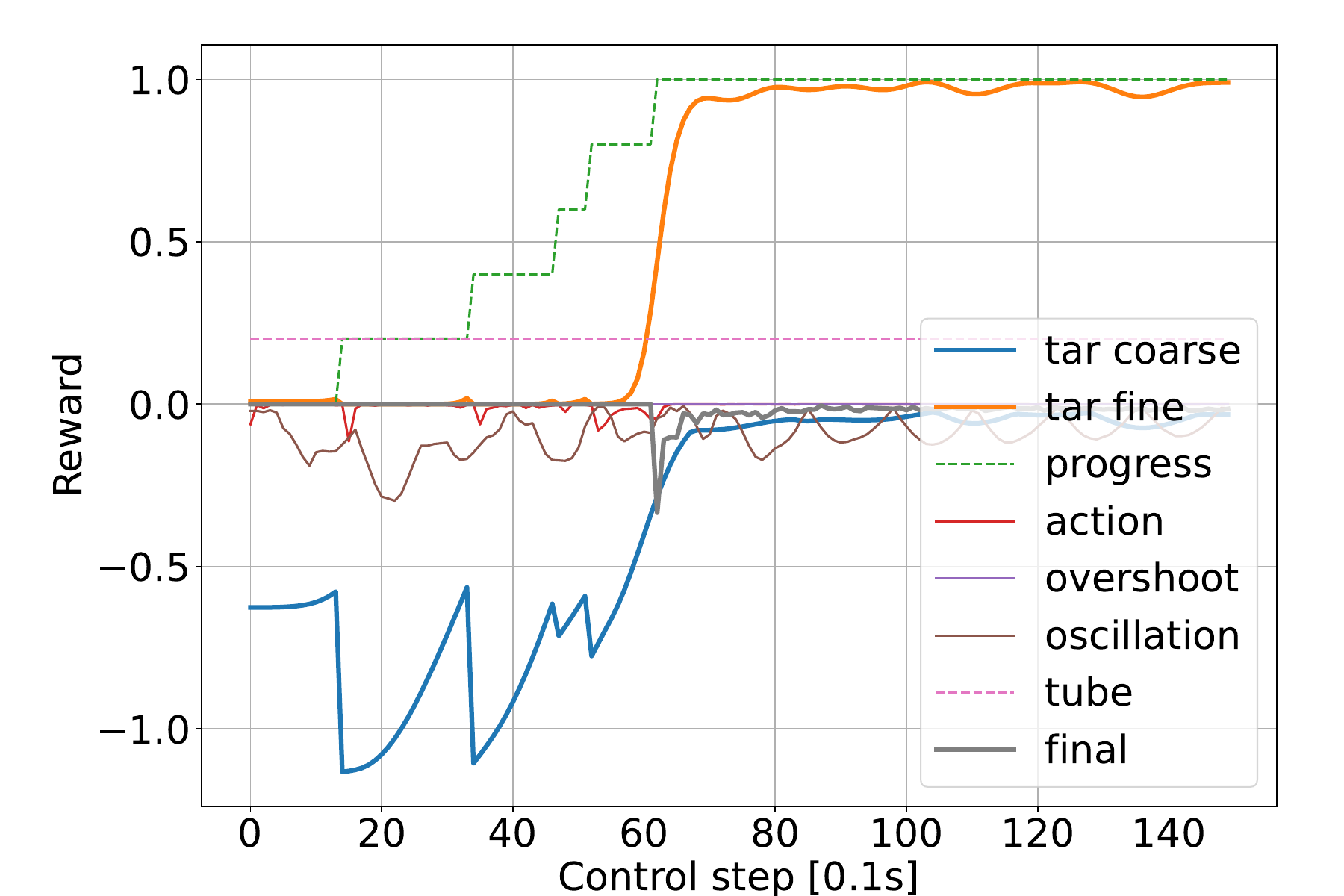}
    \caption{Reward components evolution during a $15\si{\second}$ policy rollout. Upon reaching each waypoint, $r_k^{\text{target\_coarse}}$ (blue) decreases with the updated distance. To encourage progress, we introduce $r_k^{\text{progress}}$ (green), which increases along the sequnce. At the end of the trajectory, where $r_k^{\text{target\_fine}}$ (orange) is maximized, $r_k^{\text{final}}$ (grey) forces the policy to limit any additional actions. Consequently, the agent understands the machine dynamics to command motions leading to high reward states and low oscillations without further corrections.}
    \label{fig:waypoint_reward}
\end{figure}

The reward at each timestep consists of eight components: 
\begin{equation}\label{eq:waypoint_reward}
    \begin{aligned}
        R_k & = r_k^{\text{target\_coarse}} + r_k^{\text{target\_fine}} + r_k^{\text{progress}} + r_k^{\text{action}} \\
        & + r_k^{\text{overshoot}} + r_k^{\text{oscillation}} + r_k^{\text{tube}} + r_k^{\text{final}}. 
    \end{aligned}
\end{equation}

To define the reward terms, we require the following quantities. At timestep $k$ with the next waypoint index $m$, the gripper position is $E=[x_k, y_k, z_k]^\top$. The next waypoint position is $\mathbf{w}_m=[x^{wp}_m, y^{wp}_m, z^{wp}_m]^\top=[r^{wp}_m, \phi^{wp}_{m,k}, z^{wp}_m]^\top$ where $\phi^{wp}_{m,k}$ is time dependent since it represents the angular offset between the current slew angle and the next waypoint target. The total number as well as the last index of waypoints is denoted $M=5$. The detailed reward definition is as follows:

\begin{equation*}
    \begin{aligned}
        & r_k^{\text{target\_coarse}} \propto \Big(\exp\big(-\| \tilde{\varepsilon}_{m,k} \|_1\big) -1\Big), \\ 
        & r_k^{\text{target\_fine}} \propto \exp\Big(-\| \tilde{\varepsilon}_{m,k} \|_2^2\Big), \\
        & r_k^{\text{progress}} \propto m/M, \\
        & r_k^{\text{action}} \propto -\| \Delta \textbf{u}_{k} / \sigma_u \|_2^2, \\
        & r_k^{\text{overshoot}} \propto \Big(\exp\big(-|\phi_{m,k}^{\text{ovs}}|\big)-1\Big), \\
        & r_k^{\text{oscillation}} \propto -\| \tilde{\varphi}_{k} \|_1, \\ 
        & r_k^{\text{tube}} \propto \mathbbm{1} \{ \delta_{\text{tube},k} \geq 0 \wedge \overrightarrow{p_1 E} \cdot \overrightarrow{p_1 p_2} \geq 0 \wedge \overrightarrow{p_2 E} \cdot \overrightarrow{p_1 p_2} \leq 0\}, \\
        & r_k^{\text{final}} \propto -\| \textbf{u}_{k} / \sigma_u \|_2^2 \cdot \mathbbm{1} \{ \| \tilde{\varepsilon}_{M, k} \|_2 < r_{\text{tube}} \}
    \end{aligned}
\end{equation*}
where
\begin{equation*}
    \begin{aligned}
        \tilde{\varepsilon}_{m, k} & = \big[x^{wp}_{m}-x_{k}, y^{wp}_{m}-y_{k}, z^{wp}_{m}-z_{k} \big], \\
        \phi_{m,k}^{\text{ovs}} & = 
        \begin{cases}
            \max\big( 0, \phi_{m,k} \big) & \text{if} \quad \phi_{m,0} < 0 \\
            \min\big( 0, \phi_{m,k} \big) & \text{if} \quad \phi_{m,0} > 0
        \end{cases}, \\
        \tilde{\varphi} & = \big[ \dot{q}_{\text{tool},x}, \dot{q}_{\text{tool},y} \big].
    \end{aligned}
\end{equation*}

The first three are the main rewards for waypoint tracking. \( r_k^{\text{target\_coarse}} \) and \( r_k^{\text{target\_fine}} \) encourage approaching the next waypoint by shaping a dense attraction field in 3D Euclidean space. As illustrated in \cref{fig:waypoint_reward}, \( r_k^{\text{target\_coarse}} \) is prominent when the waypoint error \( \tilde{\varepsilon}_{m, k} \) is large, guiding the agent toward the general vicinity of the target. In contrast, \( r_k^{\text{target\_fine}} \) comes into play when \( \tilde{\varepsilon}_{m, k} \) is small, providing finer rewards for precise positioning near the waypoint, particularly at the final one. 
Two separate rewards facilitate reward shaping and curriculum implementation.
$r_k^{\text{progess}}$ encourages progressing along the waypoints sequence. Without it, the decrease in $r_k^{\text{target\_coarse}}$ at each reached target may mislead the agent to a stall without reaching the following waypoints, hence never exploring the final target state. 

The remaining reward terms are designed to address material handling requirements and ensure safe deployment. The term \( r_k^{\text{action}} \), with \( \sigma_u \) the specific action scaling constant, promotes smooth actions to support safe operation on real hardware. To discourage overshooting the target due to incorrect braking of the slew joint, \( r_k^{\text{overshoot}} \) penalizes rotational overshoot. Together with \( r_k^{\text{tube}} \), which encourages motion within the tube connecting consecutive waypoints, and \( r_k^{\text{oscillation}} \), which penalizes excessive tool joint velocities, these terms constrain the tool's trajectory to the intended path, enhancing safety. Finally, \( r_k^{\text{final}} \) facilitates the next transition by minimizing tool oscillations and unnecessary movements at the final waypoint (which corresponds to the attack point), prioritizing a behavior which applies minimal corrections before transitioning to the grasping phase.

\paragraph{Training Details}
We train the agent using \ac{PPO}~\cite{Schulman17ProximalPolicy} for 5000 updates (\( N_{\text{upd}} \)). The policy and value functions are parameterized by neural networks with layer dimensions \([256, 128, 128]\), employing \(\tanh\) activations and a linear output layer. To encourage exploration, we initialize the action distribution with a covariance of $1.5$ and apply entropy regularization with a coefficient \(\beta = 0.01\). Inspired by curriculum learning in \ac{RL}~\cite{Bengio09CurriculumLearning, Soviany22CurriculumLearning}, we gradually introduce the \( r_k^{\text{final}} \) reward term using an arctangent scaling profile centered at \( N_{\text{upd}} /2\). This scheduling delays the full effect of the terminal penalty until the policy has learned the basic motion, reducing the probability of divergence during early training.

Simulation is implemented using RaiSim~\cite{Hwangbo18PerContactIteration}, running 512 environments in parallel. Each environment operates at a simulation frequency of \SI{50}{\hertz} and a control frequency of \SI{10}{\hertz}. Episodes last \SI{15}{\second}, after which environments are reset.

\subsubsection{RL Throwing Controller}\label{sec:throw_rl}
The \ac{RL} throwing controller extends the \ac{RL} waypoint following training in \cref{sec:waypoint_rl} by incorporating the gripper actuation as an additional action. Consequently, this section focuses on the elements needed to train the new skill.
 
\begin{figure}
    \centering
    \includegraphics[trim={10cm 0cm 0cm 0cm},clip, width=\columnwidth]{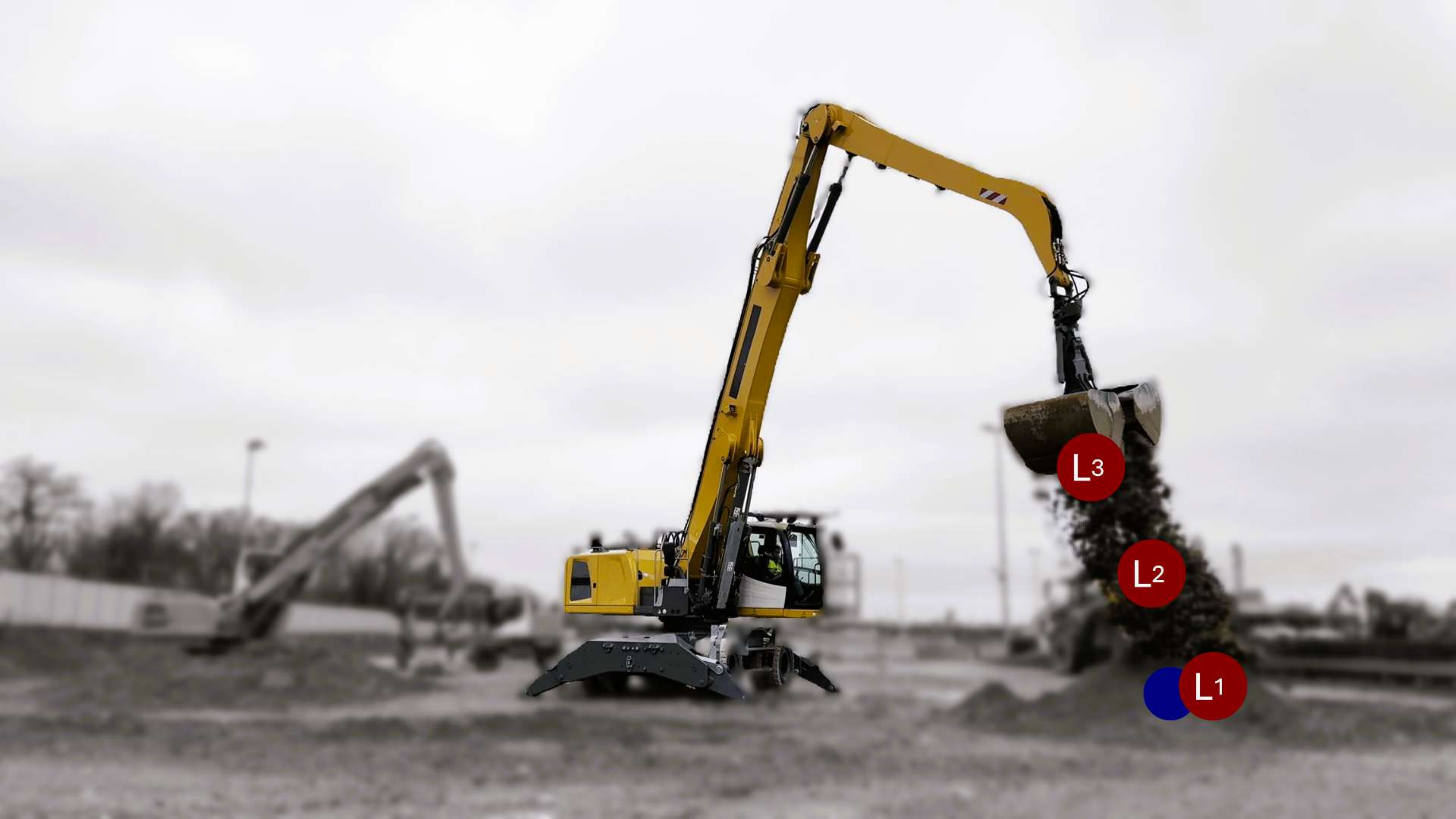}
    \caption{To simulate granular material in the training environment, we include three rigid bodies (red) released sequentially with initial velocity and position of the gripper \ac{CoM}. The target is visualized in blue and is always positioned on the ground.}
    \label{fig:throw_sim}
\end{figure}

\paragraph{Task Description}
Starting with a loaded gripper, the agent follows a randomly generated waypoint trajectory and throws all material at a designated target location on the ground. The trajectory is subject to the same waypoint-following constraints described in \cref{sec:waypoint_rl}. To compensate for the granular nature of the material, the agent must modulate the gripper’s velocity profile throughout the release phase. As described in \cref{sec:sim_dynamics}, the material is approximated by three sequentially released discrete loads $L_{1:3}$ (\cref{fig:throw_sim}), emulating continuous flow and capturing the spread distribution induced by tool oscillations.
Following material release, the policy reduces oscillations and control effort, resulting in a stable configuration suitable for initiating subsequent motions.

\paragraph{Actions and Observations}

The controller is represented by an \ac{MLP} $\pi_{th}$ as follows:
\begin{equation}\label{eq:throw_policy}
    \begin{aligned}
        \mathbf{u}_k & = \Big[u_{\text{slew},k},\dot{q}^{\text{~ref}}_{\text{~boom},k},\dot{q}^{\text{~ref}}_{\text{~stick},k},u_{\text{gripper},k}
        \Big]^\top \\
        & = \pi_{th} \Big( \mathbf{u}_{[k-H: k-1]}, \dot{\mathbf{q}}_{[k-H: k-1]}, \mathbf{q}_{k-1}, I_{z, k-1}, \mathbf{w}_{[m:m+2]},\\ & \mathbbm{1}{\{m=M\}}, \delta_{\text{tube}, k-1}, \chi_{k-1} \Big). 
    \end{aligned}
\end{equation}
 
The agent actuates all four active joints: slew, boom, stick, and gripper opening. The command $u_{\text{gripper}}\in[-1,1]$ is compared against an opening threshold of $0.5$. If $u_{\text{gripper}, k} > 0.5$, then a joystick signal for opening the gripper shells is provided until the episode ends. In this state, any additional $u_{\text{gripper}}$ has no effect. 

The policy uses all observations from the \ac{RL} waypoint-following controller and introduces two additional observations specific to the throwing task. First, a boolean flag \(\chi\) indicates whether the material has already been released. This flag is set to true once the first simulated discrete load \(L_1\) is released. Second, five past gripper actions \(u_{\text{gripper}}\) are included as part of the action history \(\mathbf{u}_{[k-H:k-1]}\). These additions enable estimating the delay between gripper actuation and actual material free fall.
No observations of the released material are provided, as such information would not be available during real-world deployment. The agent is expected to infer the parabolic trajectory based on the actuation delay, the estimated gripper velocity, and the reward signal collected throughout training.

\paragraph{Episode Initialization and Domain Randomization}

At the beginning of each episode, we randomize the delay between the gripper opening command and the release of the first discrete load \(L_1\), sampling it from a uniform distribution over \([0.24, 0.36]\,\si{\second}\), as identified through real-world experiments. The second and third loads, \(L_2\) and \(L_3\), are subsequently released at fixed intervals after \(L_1\), maintaining a constant total time to empty the bucket. To support learning of both path-following and throwing behaviors, we introduce a curriculum on the generated trajectory length. Specifically, the total number of waypoints $M$ is sampled from the range \(\left[2, \min\left(2+6\frac{n_{\text{upd}}}{N_{\text{upd}}}, 5\right)\right]\), where \(n_{\text{upd}}\) is the current training iteration and \(N_{\text{upd}}\) is the total number of updates. This approach enables the agent to acquire the throwing skill early in training, then progressively increase trajectory complexity while continuing to explore relevant throwing states. Regardless of the sampled path length, the final waypoint is always placed on the ground to represent the target piling location.

\paragraph{Termination Conditions}
Unlike \cite{Werner24dynamicThrowing}, which terminates the episode with a positive reward immediately after the throw, our approach accumulates rewards and penalties over the full episode duration without early termination. This enables simultaneous optimization of material dumping accuracy and minimization of tool oscillations following release.

\paragraph{Reward}

The proposed reward at timestep $k$ and next waypoint $\mathbf{w}_m=[x^{wp}_m, y^{wp}_m, z^{wp}_m]^\top$ in an episode with a total of $M$ waypoints consists of nine components: 
\begin{equation}\label{eq:throw_reward}
    \begin{aligned}
        R_k & = r_k^{\text{target\_coarse}} + r_k^{\text{target\_fine}} + r_k^{\text{throw}} + r_k^{\text{progress}} + r_k^{\text{action}} \\
        & + r_k^{\text{overshoot}} + r_k^{\text{oscillation}} + r_k^{\text{tube}} + r_k^{\text{final}}. 
    \end{aligned}
\end{equation}
Only four are different from the definitions in \cref{eq:waypoint_reward}:
\begin{equation*}
    \begin{aligned}
        & r_k^{\text{target\_coarse}} \propto \Big(\exp\big(-\| \tilde{\varepsilon}_{m,k} \|_1\big) -1\Big), \\ 
        & r_k^{\text{target\_fine}} \propto \begin{cases}
            \exp\Big(-\| \tilde{\varepsilon}_{m,k} \|_2^2\Big) & \text{if} \quad m \neq M   \\
            \exp\Big(-\| \tilde{\mathcal{L}}_{k} \|_2^2\Big)\cdot \chi_{k} & \text{if} \quad m = M
        \end{cases}, \\
        & r_k^{\text{throw}} \propto \begin{cases} 
            r_k^{\text{target\_fine}} \quad & \text{if} \quad m=M,\\
            0 & \text{otherwise}
        \end{cases}, \\
        &\begin{aligned}
            r_k^{\text{final}} &\propto -\| \textbf{u}_{k} / \sigma_u \|_2^2 \cdot \mathbbm{1} \big\{ \| \tilde{\varepsilon}_{M,k} \|_2 < r_{\text{tube}} \vee \mathbbm{1}{\{m=M\}} \cdot \chi_{k} \big\},
        \end{aligned}
    \end{aligned}
\end{equation*}
with quantities:
\begin{equation*}
    \begin{aligned}
        \tilde{\varepsilon}_{m,k} & = \big[x^{wp}_{m}-x_{k}, y^{wp}_{m}-y_{k}, z^{wp}_{m}-z_{k}+h \cdot \mathbbm{1}{\{m=M\}} ] \big], \\
        \tilde{\mathcal{L}}_{k} & = \frac{1}{3} \sum_{l\in\{L_1, L_2, L_3\}} \big|x^{wp}_{M}-l_{x,k}, y^{wp}_{M}-l_{y,k}, z^{wp}_{M}-l_{z,k} \big|.
    \end{aligned}
\end{equation*}

Since the final waypoint now corresponds to the throwing target on the ground, we modify the waypoint-tracking error $\tilde{\varepsilon}_{m,k}$ in \( r_k^{\text{target\_coarse}} \) to include a vertical offset of \( h = 5\,\si{\meter} \) when approaching the last waypoint. This adjustment causes the gripper to hover above the target location after releasing the material.
The reward term \( r_k^{\text{target\_fine}} \) is modified such that, when approaching the throwing target, the error is no longer based on the end-effector tracking error \( \tilde{\varepsilon}_{m,k} \), but instead on the mean absolute position error of the discrete loads relative to the target, denoted \( \tilde{\mathcal{L}}_k \), and conditioned on a successful release via the indicator \( \chi_k \). The throwing reward \( r_k^{\text{throw}} \) is made proportional to \( r_k^{\text{target\_fine}} \) when operating on the final waypoint, further encouraging accurate material release. To accommodate the new terminal condition, the criteria for triggering \( r_k^{\text{final}} \) are slightly adjusted; the penalty now activates either when the end-effector nears the target or when a throwing action has occurred.

\paragraph{Training Details}
We train the policy using \ac{PPO} with \( N_{\text{upd}} = 10{,}000 \). The combination of trajectory length curriculum, progressive activation of \( r_k^{\text{final}} \), and entropy regularization with coefficient \( \beta = 0.01 \) ensures smooth and stable learning. Simulation is performed with 512 parallel environments, each running at a simulation frequency of \SI{50}{\hertz} and a control frequency of \SI{10}{\hertz}. Each episode lasts \SI{15}{\second} and is reset thereafter.

\subsection{Grasping Controller}
\label{sec:grasp_control}
Since the interaction between the gripper and the source pile is not modeled in the simulation, we design a simple controller to perform the grasping routine. After the waypoint-following policy has stabilized the gripper at the intended attack point, the grasping controller executes three sequential phases: (1) the gripper descends until full contact with the pile is made, (2) the gripper shells close, and (3) the boom retracts upward, establishing a favorable initial configuration for the throwing policy.

This design choice simplifies simulation by avoiding the need for a computationally intensive and difficult-to-generalize particle-based environment. It also reduces the complexity of the \ac{RL} training process. However, the resulting grasping routine is not optimized for performance and therefore leads to longer grasping times than ideally achievable.

\subsection{State Machine}
\label{sec:state_machine}
Transitions between the waypoint-following, grasping, and throwing controllers are managed by a high-level \ac{SM}. It sets the current state (\texttt{WAYPOINT}, \texttt{GRASP}, \texttt{THROW}) and communicates it to a control selector; this forwards only the corresponding control commands to the low-level joint controller operating on the machine. Transitions are triggered based on feedback signals (\texttt{SUCCESS}, \texttt{FAILED}) returned from each state. To ensure seamless switching, the buffers of all \ac{RL} policies are continuously updated, regardless of whether the policy is currently active. The \ac{SM} follows a fixed execution sequence for each defined task and is therefore not adaptive to unexpected conditions. If a \texttt{FAILED} signal is received, it transitions to the \texttt{STOPPED} state, halting all control commands to the low-level interface until a manual reset is issued.

\section{Simulation Results}

We run extensive analysis in simulation to systematically investigate the response of the proposed approach across a wide number of tests, and to highlight the advantages over more traditional solutions.

\subsection{RL Attack-Point Planner} \label{sec:rl_att_pt_sim}
To evaluate the \ac{RL}-based attack point planner, we compare it against a highest-point heuristic baseline, commonly adopted in prior works~\cite{Wang21HierarchicalPlanning, Jud17PlanningControl, zhao2021TaskNet, Zhang21AutonomousExcavator}. Both planners receive the same grid-based height observation, and the heuristic simply selects the 3D coordinates of the highest point as next attack point. Since the simulation environment (\cref{sec:attack_sim_env}) introduces observation noise that degrades heuristic performance, we also run tests using ground-truth observations to approximate the best-case scenario achievable with ideal mapping and state-estimation modules.

We assess both planners on the two types of soil geometries introduced in \cref{fig:soil_model}, the unconstrained pile and the material-in-container, with results reported in \cref{tab:sim_att}. Across both geometries, the \ac{RL} policy operating under noisy observations consistently outperforms the heuristic planner, even when the latter operates with ideal noise-free conditions, both in terms of average gripper fill ratio and average number of steps required to reach positive termination. These findings indicate that the highest-point heuristic does not guarantee optimal efficiency, independently on the available terrain information. Note that the positive termination condition described in \cref{sec:rl_att_planner} requires the average material height to fall below \SI{0.15}{\meter}; consequently, the shovel fill ratio naturally decreases as the pile diminishes, and full scoops become infeasible.

\begin{table}
    % \vspace{-0.3cm}
    \begin{center}
         \caption{Attack point planner evaluation over 100 simulated episodes for the pile and container scenarios. The proposed \ac{RL} approach is compared against the baseline under varying noise conditions.}
         \label{tab:sim_att}
         \begin{tabular}{@{}clcc@{}}
             \toprule
             \textbf{Geometry} & \textbf{Policy} & \textbf{Avg. Fill} [\%]~($\uparrow$) & \textbf{Avg. Steps~($\downarrow$)}\\ \midrule
             & RL w/ noise  & \textbf{62.7} & \textbf{16.0}\\
             Pile & Heuristic w/ noise & \textit{14.6} & \textit{60.0} \\ 
             & Heuristic w/o noise & 57.7 & 17.8\\
             \midrule
             & RL w/ noise  & \textbf{62.9} & \textbf{21.6}\\
             Container & Heuristic w/ noise & \textit{33.5} & \textit{35.6} \\ 
             & Heuristic w/o noise & 61.2 & 23.0 \\
             \bottomrule
         \end{tabular}
    \end{center}
    % \vspace{-0.5cm}
\end{table}

\subsection{RL Waypoint Following \& Throwing Controllers} \label{sec:rl_way_throw_sim}

\begin{figure}
    \centering
    \includegraphics[trim={0.0cm 0cm 1cm 0.5cm},clip, width=\columnwidth]{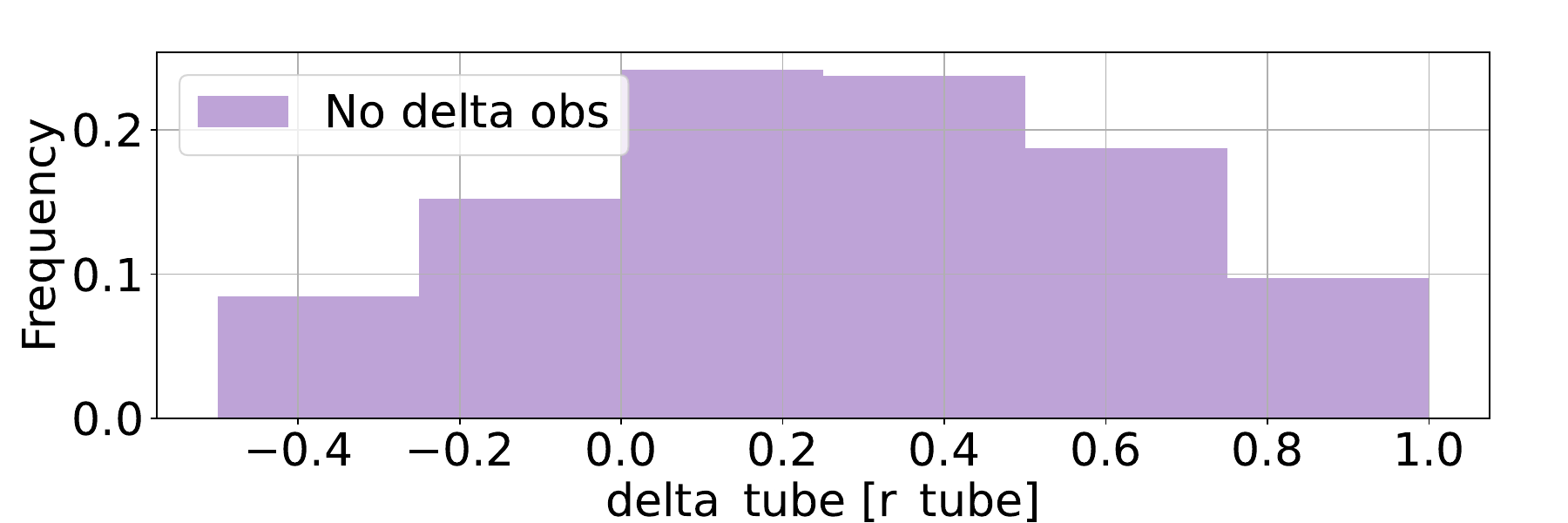}
    \includegraphics[trim={0.0cm 0cm 1cm 0.5cm},clip, width=\columnwidth]{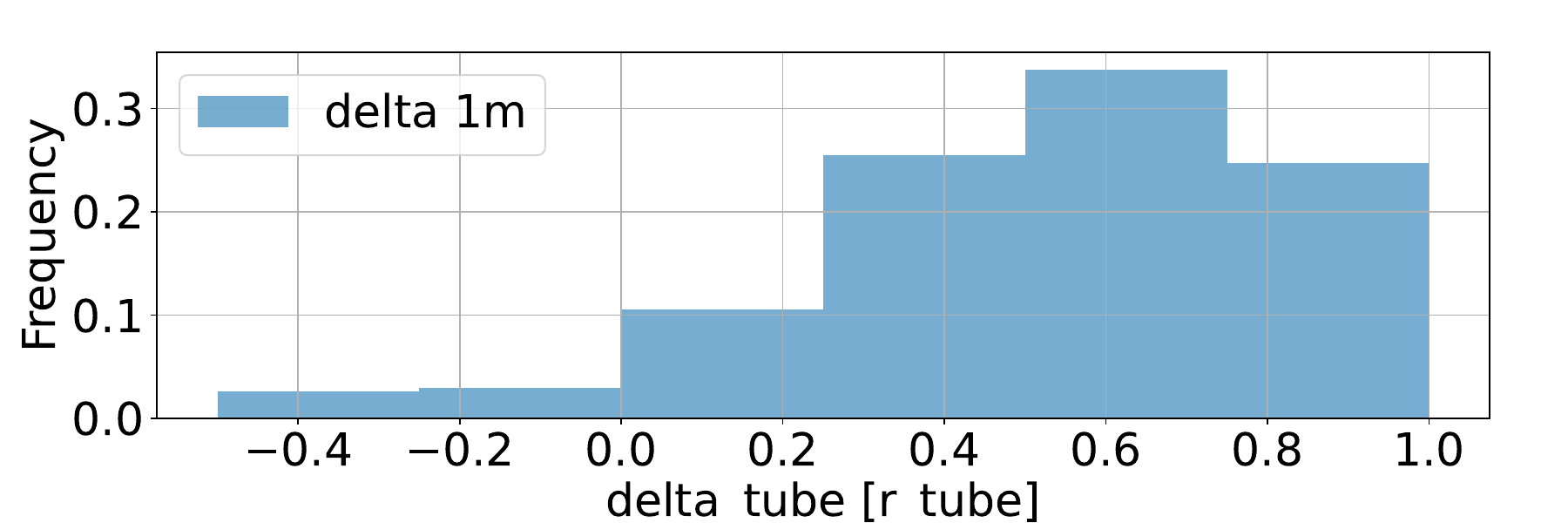}
    \includegraphics[trim={0.0cm 0cm 1cm 0.5cm},clip, width=\columnwidth]{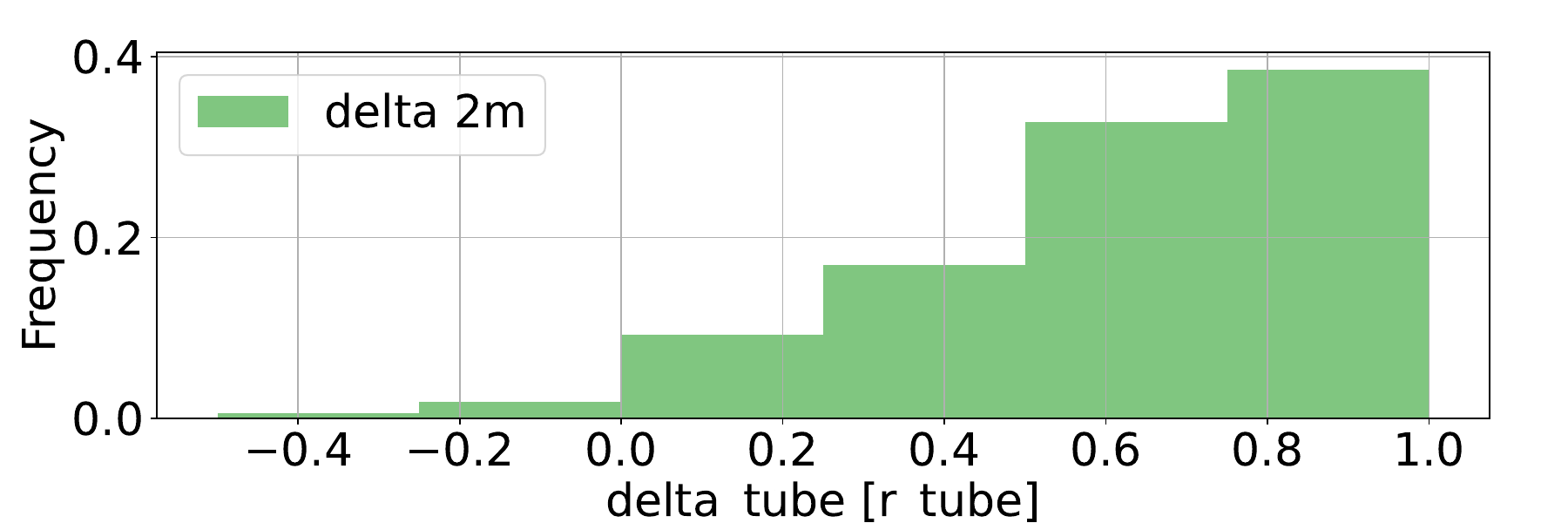}
    \caption{Distribution of the normalized distance from the tube boundary $\delta_{\text{tube}}$ during simulation rollouts for a policy without access to the $\delta_{\text{tube}}$ observation (purple), and for the \SI{1}{\meter}-tube (blue) and \SI{2}{\meter}-tube (green) policies presented in this work. The agent traverses 100 randomly sampled waypoints, ensuring constraint satisfaction ($\delta > 0$) for approximately \SI{72}{\percent}, \SI{94}{\percent} and \SI{98}{\percent} of the control steps respectively.}
    \label{fig:waypoint_bound}
\end{figure}

The simulation environment introduced in \cref{sec:sim_dynamics} is used to evaluate the controllers in multiple tasks.
First, we focus on safety guarantees.
Our two-stage planning and control pipeline (\cref{fig:ros2_overview}) requires \ac{RL} policies to track waypoints with bounded deviation to guarantee a collision-free motion. To facilitate this, we introduce the observation \( \delta_{\text{tube},k} \), representing the remaining safety margin at timestep \(k\), and tune the environment to maximize the frequency of the reward \( r_k^{\text{tube}} > 0 \). Simulation results (\cref{fig:waypoint_bound}) show that the learned policy behaves conservatively, trying to maintain \( \delta_{\text{tube}} \approx r_{\text{tube}} \) and therefore steering the tool near the tube center.  
Compared to the purple histogram—generated by a policy trained with a \SI{1}{\meter} tube but without  \( \delta_{\text{tube}} \) observation—our agent effectively adapts its motion to satisfy safety requirements. Some slack is yet tolerated to accommodate sharp trajectory segments or regulate tool velocities, which may reduce constraint satisfaction under tighter \( r_{\text{tube}} \) settings.  
These findings validate the observation vector design and environment setup, demonstrating that a well-shaped reward suffices to guide policy behavior without requiring termination conditions. Moreover, the agent’s ability to adjust control based on the observed safety margin opens the possibility of training policies with variable tube geometries, enabling more reactive behaviors.

\begin{figure}
    \centering
    \includegraphics[trim={2.75cm 0cm 2cm 1cm},clip, width=0.95\columnwidth]{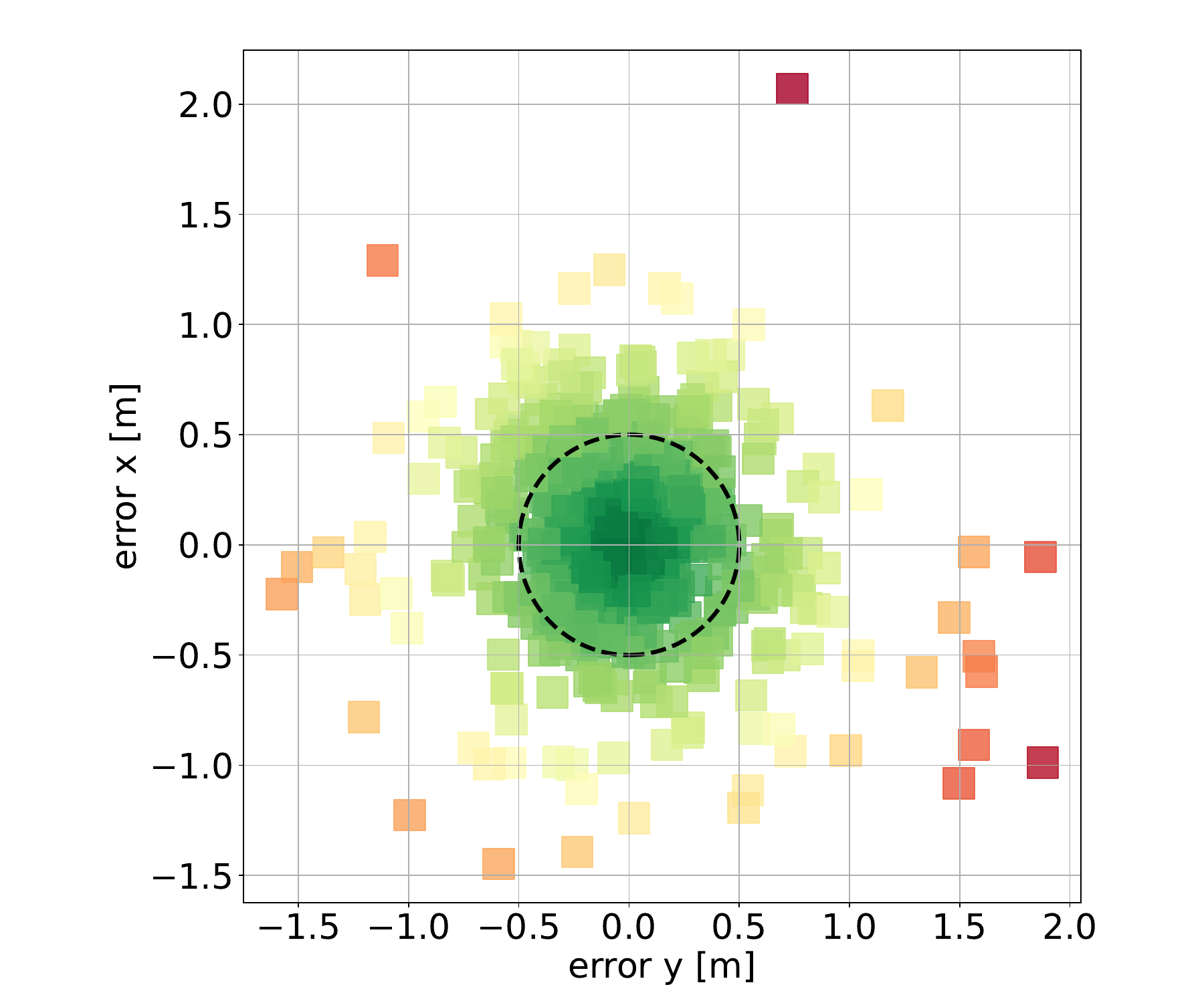}
    
    \caption{Distribution of each load throwing error after 200 policy rollouts, with randomly sampled approaching waypoints along a fixed rotation direction.   
    Each load landing position is represented by a square to simulate the granular material spread. For reference, the black circle of radius \SI{0.5}{\meter} corresponds to the ideal Gaussian reward return.}
    \label{fig:throw_error}
\end{figure}

A similar analysis is run for the throwing policy. Despite the increased action space dimensionality required to incorporate the throwing skill, the augmentation preserves the waypoint-following safety constraints. Notably, the throwing controller successfully follows randomly sampled trajectories while satisfying the \SI{2}{\meter} tube constraint in approximately \SI{98}{\percent} of cases, ruling out any performance degradation.

To validate the environment and rewards designed to learn the active throwing action, we run 200 policy rollouts with fixed rotation direction, and collect the result in \cref{fig:throw_error}. A 2D Gaussian fitted to the throwing error yields a mean of  
\[
\mu_{yx} = \begin{bmatrix} -0.014 & 0.031 \end{bmatrix} \, \si{\meter},
\]  
with covariance  
\[
\Sigma_{yx} = \begin{bmatrix} 0.225 & -0.026 \\ -0.026 & 0.216 \end{bmatrix}.
\]
We can thus assess an unbiased error, and an achieved precision exceeding the physical limits of the real robot, which is equipped with a gripper measuring $2.0 \si{\meter} \times 1.5 \si{\meter}$. Consequently, the real-hardware errors reported in \cref{tab:throw_stat} stem not from incorrect execution of the throwing skill, but from sim-to-real mismatches and the uncontrollable flow of material after release.

\begin{table*}
    % \vspace{-0.3cm}
    \begin{center}
         \caption{Two \ac{MPC} reward tunings are provided (slow and fast), and \textit{vanilla} variants not penalizing tool oscillations. These configurations, along with the \ac{RL} controllers, are evaluated under identical scenarios with progressively increasing actuation delays. Performance is compared across several metrics: average slew speed during transitions, average end-effector error at final trajectory targets, average tool angular velocity, tube constraint satisfaction rate, average normalized safety margin, maximum constraint violation, and average cycle time (excluding the grasping phase, which is identical across controllers). The \textbf{best} and \textit{worst} values for each category are highlighted.}
         \label{tab:mpc_stat}
         \begin{tabular}{@{}cccccccc@{}}
             \toprule
             \textbf{Controller} & \textbf{Speed} [\si{\degree \per \second}] & \textbf{Pos Error} [\si{\meter}] & \textbf{Oscillations} [\si{\degree \per \second}] & $\delta_\text{tube}>0$ [\%] & \textbf{Norm.} $\delta_\text{tube}$ & \textbf{Violation} [\si{\meter}] & \textbf{Time w/o Grasp} [\si{\second}]  \\ \midrule
             MPC vanilla slow & \textit{8.78} & 0.85 & 37.21 & \textbf{98} & \textbf{0.64} & \textbf{0.55} & \textit{36.51} \\
             MPC vanilla fast & \textbf{13.52} & \textit{1.31} & \textit{63.18} & 73 & \textit{0.29} & \textit{2.88} & \textbf{26.27} \\
             MPC delay \SI{0.0}{\second} & 12.90 & 1.24 & 54.51 & \textit{70} & \textit{0.29} & 1.95 & 27.67 \\
             MPC delay \SI{0.1}{\second} & 11.57 & 1.20 & 55.19 & 74 & 0.32 & 2.69 & 29.98 \\
             MPC delay \SI{0.2}{\second} & FAILED & - & - & - & - & - & - \\ \midrule
             RL delay \SI{0.0}{\second} & 12.10 & 0.28 & 14.66 & 95 & 0.61 & 0.80 & 29.47 \\
             RL delay \SI{0.1}{\second} & 12.32 & 0.32 & 14.00 & 94 & 0.60 & 0.85 & 28.96 \\
             RL delay \SI{0.2}{\second} & 12.87 & 0.35 & \textbf{13.09} & 94 & 0.60 & 0.83 & 29.66 \\
             RL delay \SI{0.3}{\second} & 11.76 & \textbf{0.27} & 13.72 & 94 & 0.62 & 0.84 & 31.97 \\
             \bottomrule
         \end{tabular}
    \end{center}
    % \vspace{-0.5cm}
\end{table*}

Finally, we test the complete manipulation routine.
To highlight the advantages of the \ac{RL}-based approach proposed in this work, we use as baseline an \ac{MPC} controller inspired by~\cite{kalmari2014nonlinear}. A detailed implementation is available in Appendix \hyperref[apx:mpc]{A}.
Tests are run in simulation, first assuming ideal conditions with no actuation delay, then progressively increasing it to approximate real hardware behavior.
The results, summarized in \cref{tab:mpc_stat}, show that \ac{MPC} controllers can be tuned to favor either efficiency or accuracy. In contrast, the \ac{RL} policy consistently achieves top-tier performance across all metrics. In the remaining of the section, we select the \ac{MPC} \textit{fast} as best candidate, being it aligned to \ac{RL} in terms of average cycle time.  
While the \textit{vanilla} variant doesn't account for oscillation minimization, the \ac{MPC} can optimize over an approximate tool model to improve control and safety metrics. The \ac{RL} approach still outperforms it in terms of end-effector error, oscillation suppression, and tube constraint satisfaction. The simplified slew dynamics model used by the \ac{MPC} fails to capture inertia-driven acceleration effects, resulting in significant overshoot. Similarly, the tool dynamics limited to the prediction horizon do not allow to effectively leverage long-term couplings between actuated and unactuated joints, limiting the controller’s ability to damp oscillations. 
Differently from the \ac{MPC} framework, relying on an accurate predictive model to optimize control actions over a finite horizon, the \ac{RL} training paradigm incorporates complex temporal correlations directly into the policy, leveraging an history of observed states to infer the underlying system dynamics. This distinction becomes particularly evident when increasing the actuation delays: whereas the performance of \ac{MPC} progressively deteriorates and ultimately becomes unstable, the proposed \ac{RL} controller dynamically adjusts its behavior, achieving a robust and consistent output.

Although more sophisticated \ac{MPC} schemes could be designed by optimizing over augmented state histories, accounting for delays and nonlinear dynamics, we argue that our task is better suited for \ac{RL}:
\begin{itemize}
    \item Slew dynamics are challenging to model from first principles. Adopting partial models in receding-horizon methods leads to overshoot and instability. In contrast, an \ac{RL} policy can learn through data-driven training.
    \item Hydraulic actuation dynamics are inherently unpredictable. While adaptive \ac{MPC} methods exist, \ac{RL} provides a more direct and effective mechanism for handling adaptation to model mismatches.
    \item Despite the derived mathematical model, tool oscillations are load-dependent. The gripper material load is typically poorly measured; consequently, model-based approaches suffer reduced accuracy, whereas our \ac{RL} policies demonstrate robustness across varying loading conditions.
\end{itemize}
In summary, \ac{RL} is better suited for tasks involving significant model uncertainty and complex dynamics. Ongoing research~\cite{oh2025discovering} is also aiming to reduce manual design and tuning efforts, addressing one of the primary limitations of the approach.

\section{Real-World Experimental Results}

Our work is validated with a thorough set of experiments on the \SI{40}{\tonne} research autonomous material handler presented in \cref{fig:machine_frame}. We analyze the performance of each proposed module: the efficiency of the attack-point planner, the adherence to safety requirements of the path planner and waypoint-following controller, and the precision of the throwing policy.
Then, we demonstrate the integrated pipeline of \cref{fig:ros2_overview}, implemented using ROS 2, for two representative tasks: bulk material pile management and dump truck loading. We quantitatively evaluate the overall performance and compare it with that of human operators. Notably, even if focusing on different aspects, all the reported experiments are collected by running the full pipeline in similar conditions. Consistent results are achieved across all trials.

\subsection{Attack-Point Planning}

\begin{figure}[]
    \centering
    \subfigure[RL planner]{
        \includegraphics[trim={1.325cm 1.1cm 4.75cm 2.3cm}, clip, width=0.45\columnwidth]{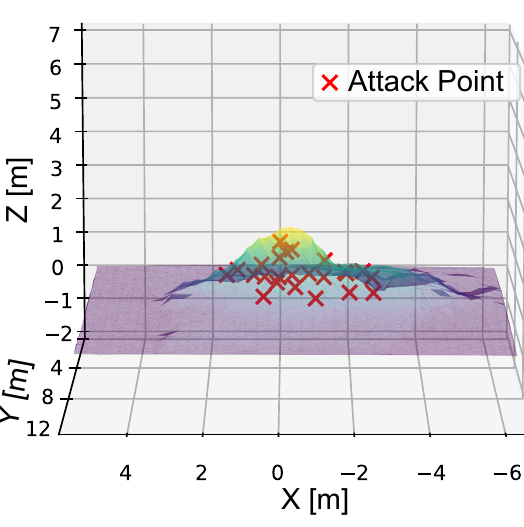}
        \label{fig:grasp_rl}
    }
    \subfigure[Heuristic planner]{
        \includegraphics[trim={1.325cm 1.1cm 4.75cm 2.3cm}, clip, width=0.45\columnwidth]{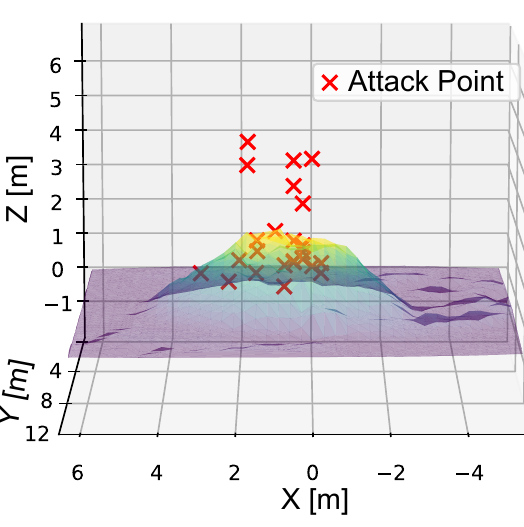}
        \label{fig:grasp_heuristic}
    }

    \caption{Visualization of attack points proposed by our RL planner and the baseline heuristic planner, over the initial pile profile. The heuristic planner directly outputs the highest point in the point cloud observation, making it more susceptible to noise.}
    \label{fig:grasp_points}
\end{figure}

\begin{table}
    % \vspace{-0.3cm}
    \begin{center}
         \caption{The performance achieved using the attack points planned by the \ac{RL} policy compared with the heuristic-based approach in terms of grasping success and average bucket fill ratio per cycle.}
         \label{tab:grasp_stat}
         \begin{tabular}{@{}cccc@{}}
             \toprule
             \textbf{Policy} & \textbf{Success} [\%] & \textbf{Avg. Fill} [\%] & \textbf{Avg. Success Fill} [\%]\\ \midrule
             RL Planner  & \textbf{100} & \textbf{67.3} & \textbf{67.3}\\
             Heuristics & 76 & 47.9 & 62.9\\ 
             \bottomrule
         \end{tabular}
    \end{center}
    % \vspace{-0.5cm}
\end{table}

First, we analyze the \ac{RL} attack point planner. We compare its impact on material moving efficiency against that of the heuristics-based planner presented in \cref{sec:rl_att_pt_sim}.
To ensure a fair comparison under consistent initial pile conditions, each test begins with a manually constructed pile within a $6\times6$~\si{\meter} grid, with volumes ranging from \SIrange{36}{41}{\meter^3} and peak heights from \SIrange{2.3}{2.6}{\meter}. Both planners are evaluated under identical scenarios, with their proposed attack points executed within the full pipeline in the bulk pile management setting.
We record all grasps performed with each method, extracting how many of them successfully picked up material and computing the average bucket fill ratio per cycle.
The experimental results in \cref{tab:grasp_stat} demonstrate that the \ac{RL} attack point planner outperforms heuristics on all metrics.
This experiment highlights the major advantage of the \ac{RL} method, which is its robustness to observation noise.
As the pointcloud input used by both planners is affected by noisy observations above the pile generated from material dropping, the heuristic planner is often misled to propose attack points in the air, leading to unsuccessful grasps~(\cref{fig:grasp_points}).
The \ac{RL} planner, on the other hand, is robust to such noise and provides consistent attack points.
Hence, our method is able to grasp \SI{41}{\percent} more volume of material on average.
When considering only successful grasps, our method still improves over the baseline, as it accounts for the shape of the gripper and the profile of the pile, and selects attack points that result in a more optimal filling. This result agrees with the finding from simulated experiments in \cref{tab:sim_att}, showing that even under optimal conditions, the heuristic planner performs worse than the \ac{RL}-based planner.

Additionally, our experiments show that expert human operators, much like the \ac{RL} planner, do not necessarily target the highest point of the pile. In many cases, attacking the slope rather than the peak yields a fuller scoop. Notable, the highest recorded average fill ratio across all of our tests is \SI{76}{\percent} (\cref{tab:bulk_stat}), result of a lack of ideal conditions: flat soil surface, full gripper contact, sufficient depth and, crucially, loose material. As the pile is depleted, its geometry becomes increasingly irregular and shallow, rendering a full scoop infeasible.

\subsection{Waypoint Following and Obstacle Avoidance}

\begin{figure*}
    \centering
    \begin{minipage}{0.5\columnwidth}
        \begin{overpic}[trim={11cm 10cm 10cm 2cm},clip, width=\textwidth]{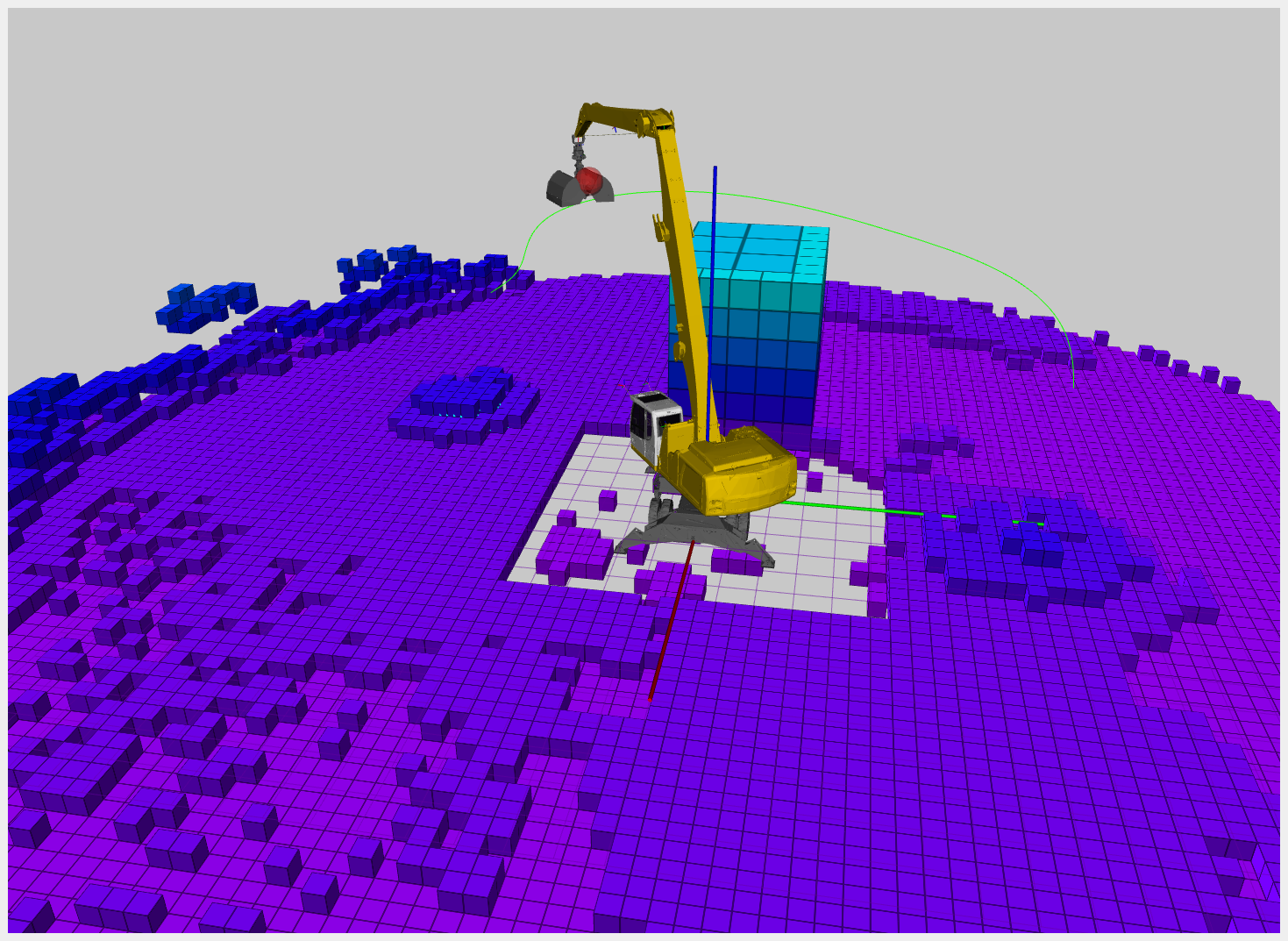}
            \put(90,85){\textcolor{black}{\textbf{A}}}
        \end{overpic}
    \end{minipage}
    \begin{minipage}{1.49\columnwidth}
        \begin{overpic}[trim={3cm 5cm 2cm 7cm},clip, width=\textwidth]{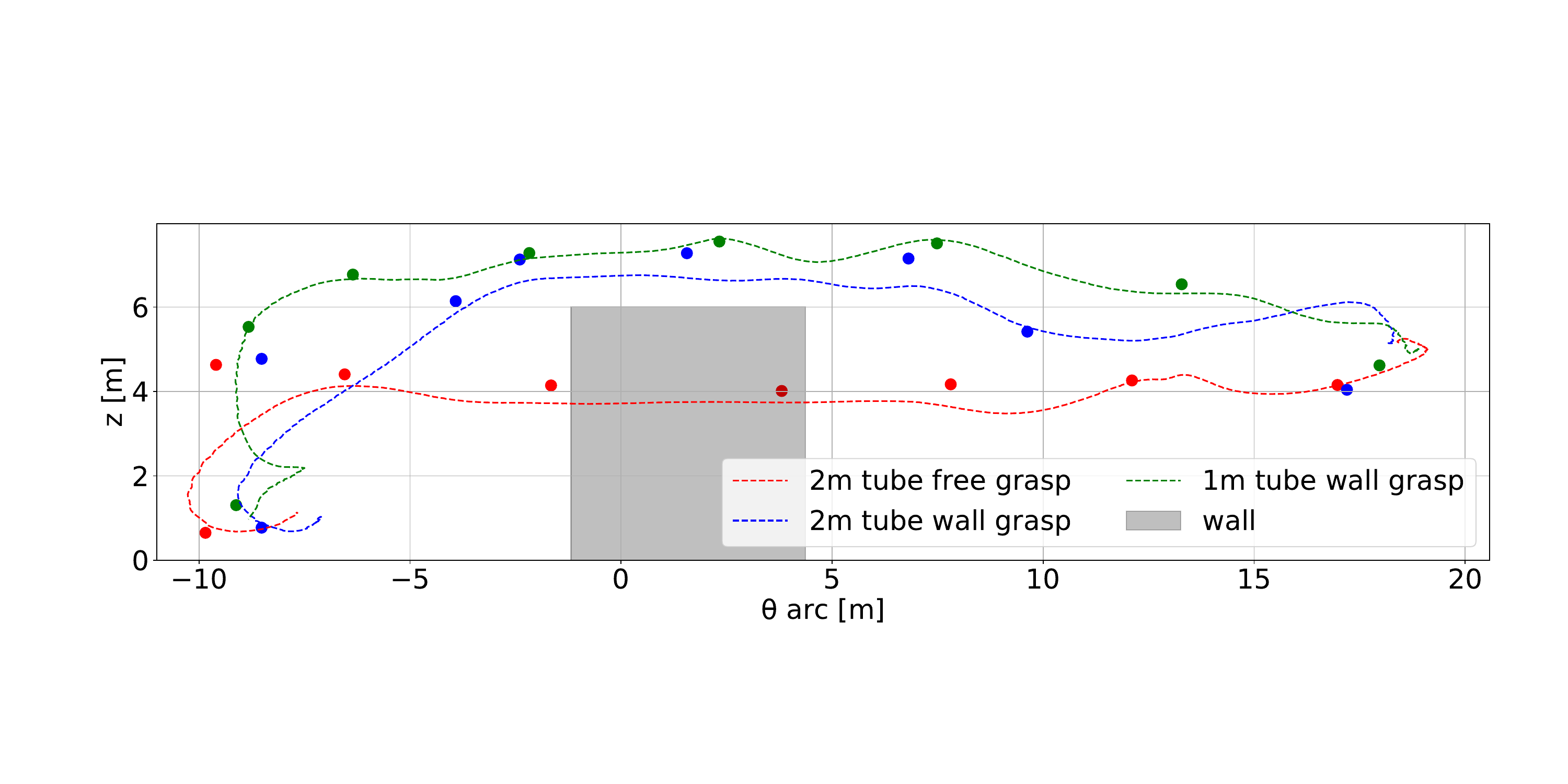}
            \put(95,26){\textcolor{black}{\textbf{B}}}
        \end{overpic}
    \end{minipage}
    \begin{overpic}[trim={4cm 0cm 2cm 3cm},clip, width=\textwidth]{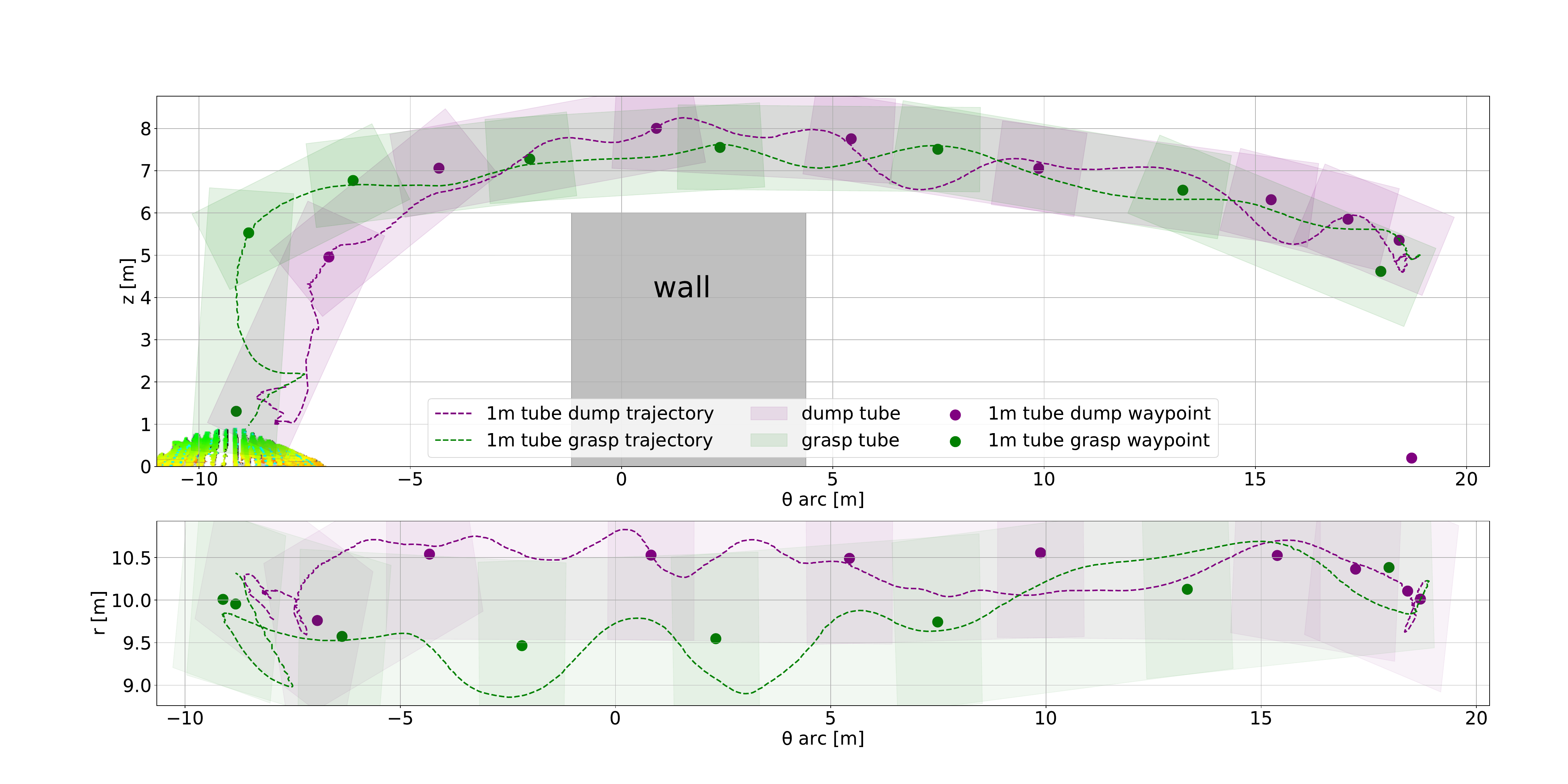}
        \put(94,46){\textcolor{black}{\textbf{C}}}
        \put(95,7){\textcolor{black}{\textbf{D}}}
    \end{overpic}
    
    \caption{Experimental results of the proposed framework with a virtual wall obstacle, evaluated with different tube settings. (A) shows a visualization of the OctoMap used by the path planner, including the grasping pile and the virtual obstacle. (B) shows from the cabin view the grasping trajectories of a controller when operating in free space (red) and when in the presence of a \SI{6}{\meter} virtual wall. Distinct behaviors emerge for policies trained with \SI{2}{\meter} (blue) and \SI{1}{\meter} (green) tube constraints. (C) and (D) deeply analyze the behavior of the \SI{1}{\meter} tube policy, from cabin view and top-down view respectively. They illustrate the commanded waypoints and tube boundaries in light colors. Grasping and dumping trajectories are shown separately, along with an approximate point cloud of the pile for reference.}
    \label{fig:planning_around}
\end{figure*}

\begin{table}
    % \vspace{-0.3cm}
    \begin{center}
         \caption{Evaluation of the waypoint-following policy for multiple consecutive cycles. We report the tube constraint satisfaction percentage, the average normalized safety margin, and the maximum constraint violation for two different policies trained with $r_{\text{tube}}$ of $1\si{\meter}$ and $2\si{\meter}$ respectively. Simulated, obstacle-free, and cluttered scenarios are tested.}
         \label{tab:tube_stat}
         \begin{tabular}{@{}ccccc@{}}
             \toprule
             \textbf{Policy} & \textbf{Scenario} & $\delta_\text{tube}>0$ [\%] & \textbf{Norm.} $\delta_\text{tube}$ & \textbf{Violation} [\si{\meter}]\\ \midrule
              & Simulation & 94 & 0.52 & 0.50 \\ 
             $1\si{\meter}$ Tube & Free Space & 95 & 0.51 & 0.49 \\
              & Obstacle & \textit{91} & \textit{0.41} & 0.54 \\ \midrule 
              & Simulation & 98 & 0.62 & \textit{0.86} \\
             $2\si{\meter}$ Tube & Free Space & \textbf{99} & 0.64 & \textbf{0.11} \\
              & Obstacle & 97 & \textbf{0.65} & 0.61 \\
             \bottomrule
         \end{tabular}
    \end{center}
    % \vspace{-0.5cm}
\end{table}

We evaluate the proposed arm path planner and RL waypoint-following controller in a material pile management setting, where a virtual wall is manually added as voxels in the OctoMap (\cref{fig:planning_around} (A)). The framework must perform grasp-and-dump cycles while avoiding the virtual wall obstacle.
First, the task is performed three times with the same start and goal points under different conditions: (1) free space between the starting point and the goal, executed by the \texttt{2m Tube} policy, (2) shortest path blocked by a virtual wall obstacle, executed by the same \texttt{2m Tube} policy, and (3) shortest path blocked by a virtual wall, executed by the \texttt{1m Tube} policy.
The resulting grasping trajectories are plotted in~\cref{fig:planning_around} (B), showing the tracking behavior of the controllers across the waypoints and the influence of the tube radius selected at training time.
These experiments indicate that the framework can handle realistic obstacles, smoothly adapting the trajectory to move over the wall.

The statistics of tube constraint satisfaction of the path-following controller are summarized in \cref{tab:tube_stat}. Moreover, \cref{fig:planning_around} (C) and (D) visualize the grasping and dumping trajectories in cylindrical coordinates with superimposed \SI{1}{\meter}-tube. Since the sampling-based path planner is queried every time before any phase change, to react to changes in the workspace, different waypoints for the two trajectories are provided.
The \texttt{1m Tube} policy satisfies the constraint in \SI{95}{\percent} of free-space samples and \SI{91}{\percent} when a \SI{6}{\meter} wall is present, incurring a single worst violation of \SI{0.54}{\meter}. These values are aligned with the ones obtained in simulation, certifying successful sim-to-real transfer.
The trajectories in \cref{fig:planning_around} (C) confirm that most of the time the gripper remains inside the rendered safety tube for both dumping and grasping segments. This guarantees safety upon deployment in cluttered environments.
Increasing the tube radius for the \texttt{2m Tube} policy raises the constraints satisfaction rate to \SI{99}{\percent}-\SI{97}{\percent} and trims the maximum violation to \SI{0.11}{\meter} in free space. The controller, however, tends to shortcut corners more often: for $\theta_{\mathrm{arc}}\in[-10,-5]$ \si{\meter} the end-effector departs by up to \SI{0.61}{\meter} from the interpolated waypoints, shown by the blue trace in \cref{fig:planning_around} (A).
This occasional divergence, identified as an optimal solution during training and already visible in the simulation results, speeds up the grasp approach yet reduces obstacle clearance.
Overall, these two radii span a spectrum of control strategies between fast path-tracking and maximal safety margin. When addressing complete tasks (\cref{sec:pile_exp}, \cref{sec:truck_exp}), the appropriate policy has to be selected according to the specific requirements. 

Our results confirm the framework's ability to safely handle environments with obstacles, hence enabling \ac{RL} policies to perform large-scale manipulation tasks in complex worksites.

\subsection{Granular Material Throwing}

\begin{figure*}[]
    \centering
    \subfigure[RL single throwing policy]{
        \includegraphics[trim={1.1cm 1.25cm 2.4cm 2.2cm}, clip, width=0.28\textwidth]{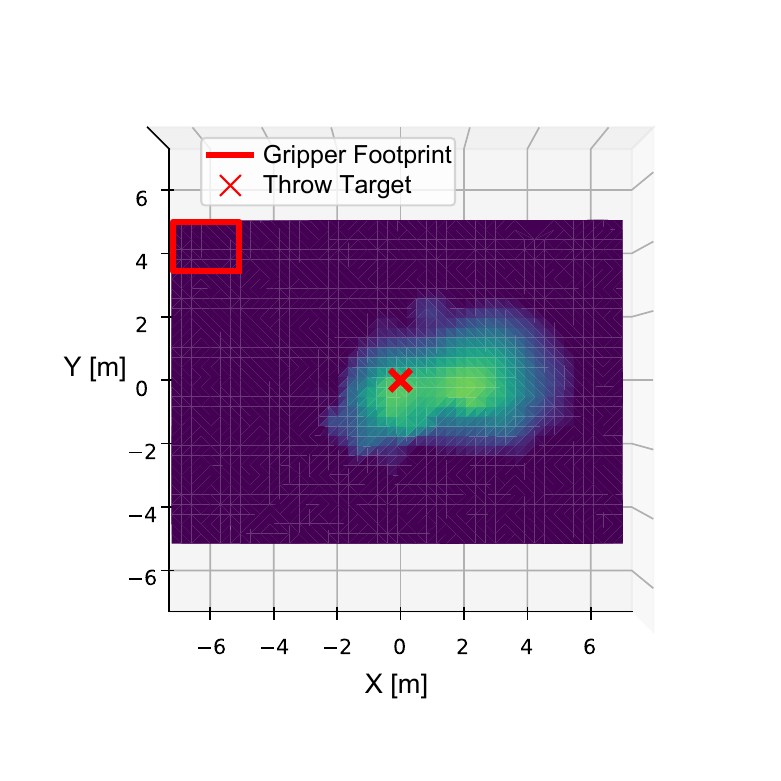}
        \label{fig:throw_rl_single}
    }
    \hfill
    \subfigure[RL multi throwing policy]{
        \includegraphics[trim={1.1cm 1.25cm 2.4cm 2.2cm}, clip, width=0.28\textwidth]{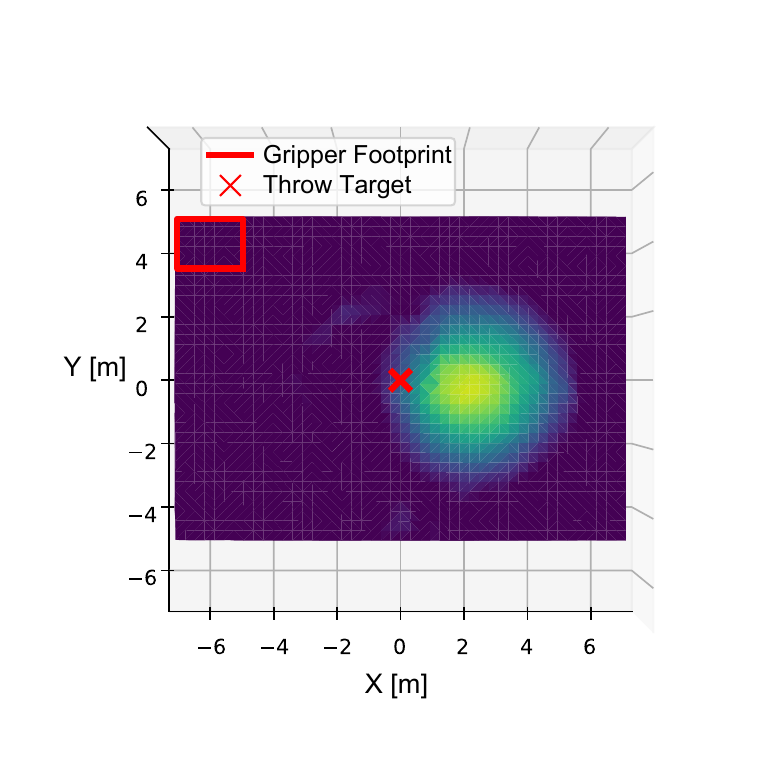}
        \label{fig:throw_rl_multi}
    }
    \hfill
    \subfigure[\SI{1}{\meter} Tube throwing policy]{
        \includegraphics[trim={1.1cm 1.25cm 2.4cm 2.2cm}, clip, width=0.28\textwidth]{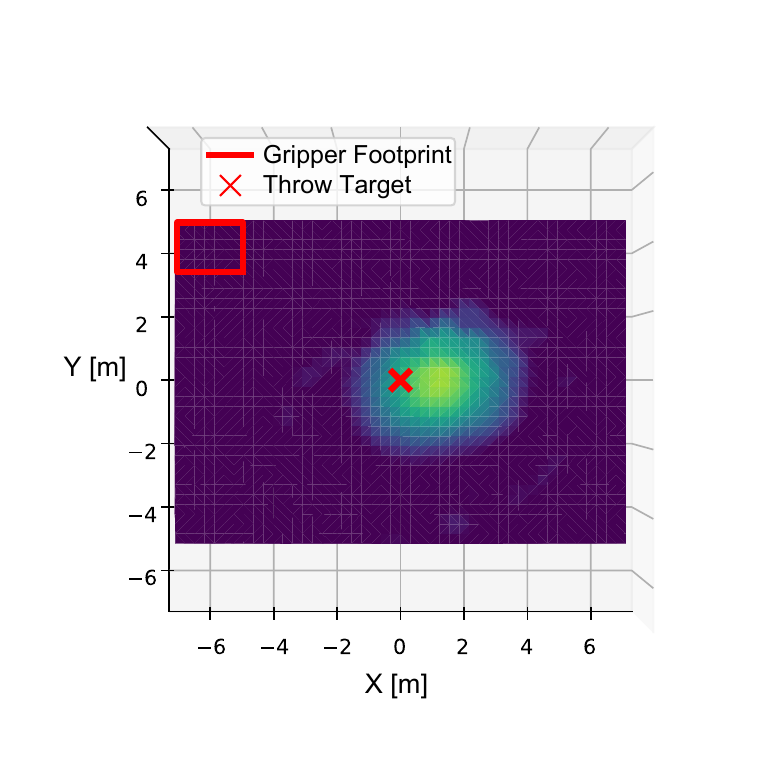}
        \label{fig:throw_heuristic}
    }
    \hfill
    \includegraphics[trim={12.5cm 1.8cm 0.25cm 2.35cm}, clip, width=0.075\textwidth]{images/grasp_points_vis/heuristic_edited.pdf}

    \caption{Final pile heights resulting from different throwing policies. The single and multi throwing policies are trained with one and three simulated discrete material bodies, respectively. The \SI{1}{\meter} tube policy is also trained with three bodies, but under an additional \SI{1}{\meter} tube constraint. While all policies exhibit some degree of overshoot relative to the target (machine rotation direction from left to right), incorporating multi-body dynamics and the tube constraint leads to improved precision and more uniform pile distribution. For reference, the red box in the top left corner indicates the dimensions of the open gripper ($2\si{\meter} \times 1.5\si{\meter}$).}
    \label{fig:throw_meshes}
\end{figure*}

\begin{table}
    % \vspace{-0.3cm}
    \begin{center}
         \caption{Resulting pile distribution for each policy, after the robot has moved approximately  \SI{24}{\cubic \meter} of granular material (about 20 scoops) between the same two locations. The final pile is fitted with a 2D Gaussian distribution from which the mean and determinant of the covariance $\Sigma$ are computed.}
         \label{tab:throw_stat}
         \begin{tabular}{@{}ccc@{}}
             \toprule
             \textbf{Policy} & \textbf{Mean Error} [\si{\meter}] & \textbf{Det($\Sigma$)} [\si{\meter^4}] \\ \midrule
             Simulation & 0.034 & 0.048 \\ \cmidrule(lr){2-3}
             Single & 1.415 & \textit{2.613} \\
             Multi & \textit{2.298} & 2.449\\
             \SI{2}{\meter} Tube & 2.061 & 2.132 \\
             \SI{1}{\meter} Tube & \textbf{1.151} & \textbf{1.878} \\
             \bottomrule
         \end{tabular}
    \end{center}
    % \vspace{-0.5cm}
\end{table}

To investigate the performance of the throwing controller, we perform a series of experiments in which the robot is required to move a pile of soil to a designated location, making use of the throwing skill. The same experiment is repeated with different policies:
\begin{enumerate}
    \item The \texttt{Single} policy, where the throwing skill is trained based on the first discrete load $L_1$ only, without waypoint-tracking capabilities, similarly as done in~\cite{Werner24dynamicThrowing}. This serves as a baseline.
    \item The \texttt{Multi} policy, where the throwing skill is trained to optimize the throwing accuracy of the distribution of all discrete particles $L_{1:3}$, without waypoint-tracking capabilities.
    \item The \texttt{2m} and \texttt{1m Tube} policies, where the controller is trained to follow a path with a tube constraint leading to the throwing target, as proposed in \cref{sec:throw_rl}.
\end{enumerate}

We scan the resulting pile formed after all material has been thrown at the target location and visualize the distributions in \cref{fig:throw_meshes}. Each pile is fitted with a 2D Gaussian, and the corresponding statistics are summarized in \cref{tab:throw_stat}. The inclusion of multiple discrete particles during training of the \texttt{Multi} policy helps improving precision compared to the \texttt{Single} policy, as confirmed by both the pile dispersion metrics in \cref{tab:throw_stat} and the qualitative distributions in \cref{fig:throw_meshes}. This indicates that our simulation environment, tuned for granular material behavior, effectively enhances policy performance in terms of pile compactness compared to previously proposed solutions~\cite{Werner24dynamicThrowing}.
When combined with waypoint-following policies, the throwing skill demonstrates improved accuracy and precision in the resulting dump pile. Without intermediate waypoints, the throwing policy tends to operate faster, resulting in fast and dynamic throwing motions. In contrast, incorporating waypoint-following leads to more conservative and stable behavior, producing a less dynamic but better-aligned throwing phase. 
These different motions are exemplified by the \texttt{Fast Throw} policy and the \texttt{1m Tube} policy in \cref{tab:bulk_stat}, and are analyzed later.

Nevertheless, the mean error observed in real-world trials remains two orders of magnitude higher than in simulation, highlighting a large sim-to-real gap. The accuracy trends in \cref{tab:throw_stat} are influenced by control performance and therefore do not follow a consistent pattern. In particular, all policies exhibit overshoot in the rotational direction, a phenomenon previously reported in~\cite{Spinelli24ReinforcementLearning}. This behavior, likely caused by the sim-to-real mismatch in the dynamics of the slew joint, prevents the controller from accurately timing the release when operating at high speed. Consistent with this claim, the slowest controller (\texttt{1m Tube}) achieves the lowest final error.

\subsection{Bulk Material Pile Transfer}\label{sec:pile_exp}

\begin{figure*}
    \centering
    \begin{minipage}{\columnwidth}
        \begin{overpic}[trim={0cm 10cm 0cm 0cm},clip, width=\textwidth]{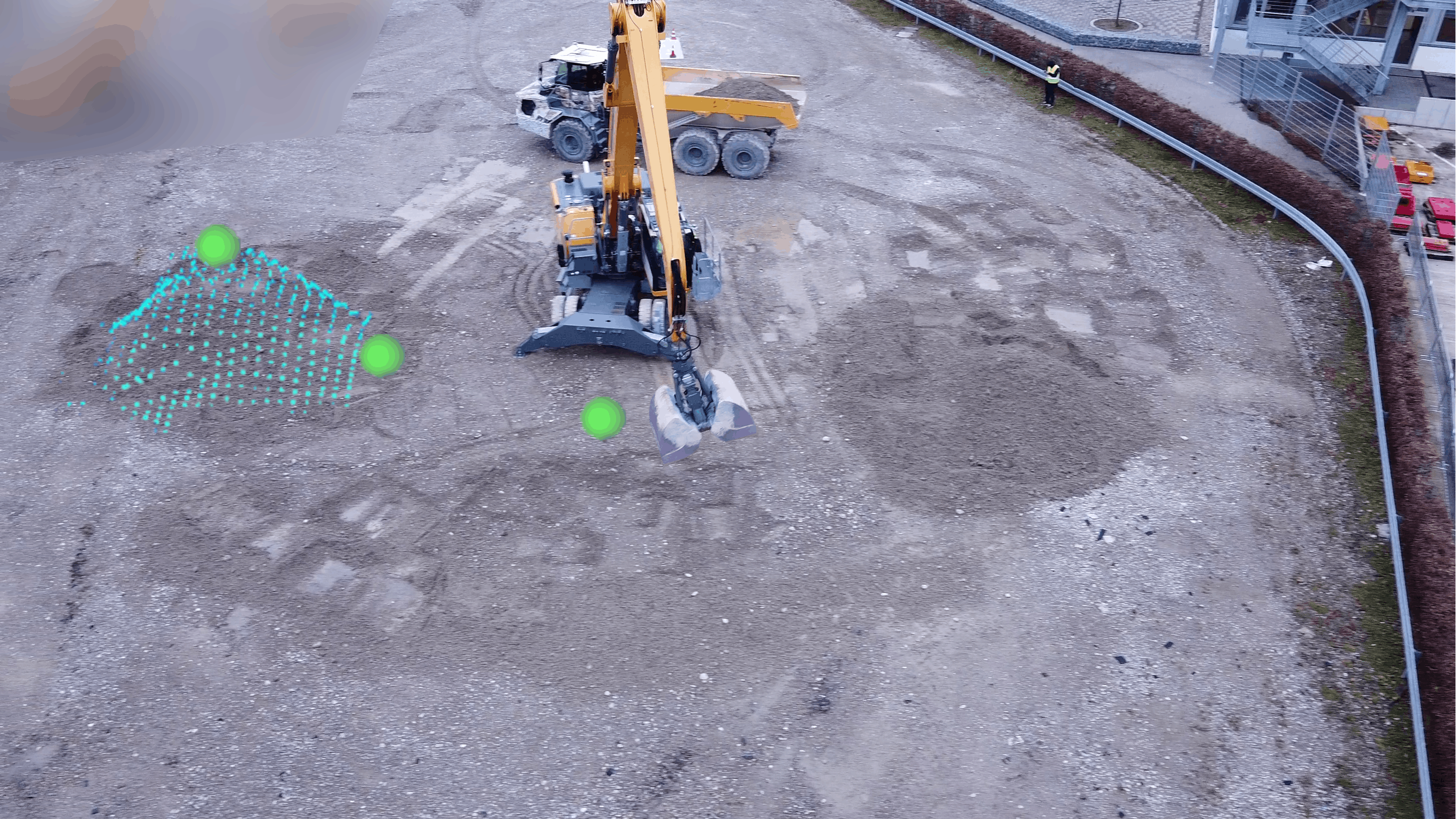}
            \put(5,5){\textcolor{white}{\textbf{A}}}
        \end{overpic}
        \par\vspace{0.5\columnsep}
        \begin{overpic}[trim={0cm 10cm 0cm 0cm},clip, width=\textwidth]{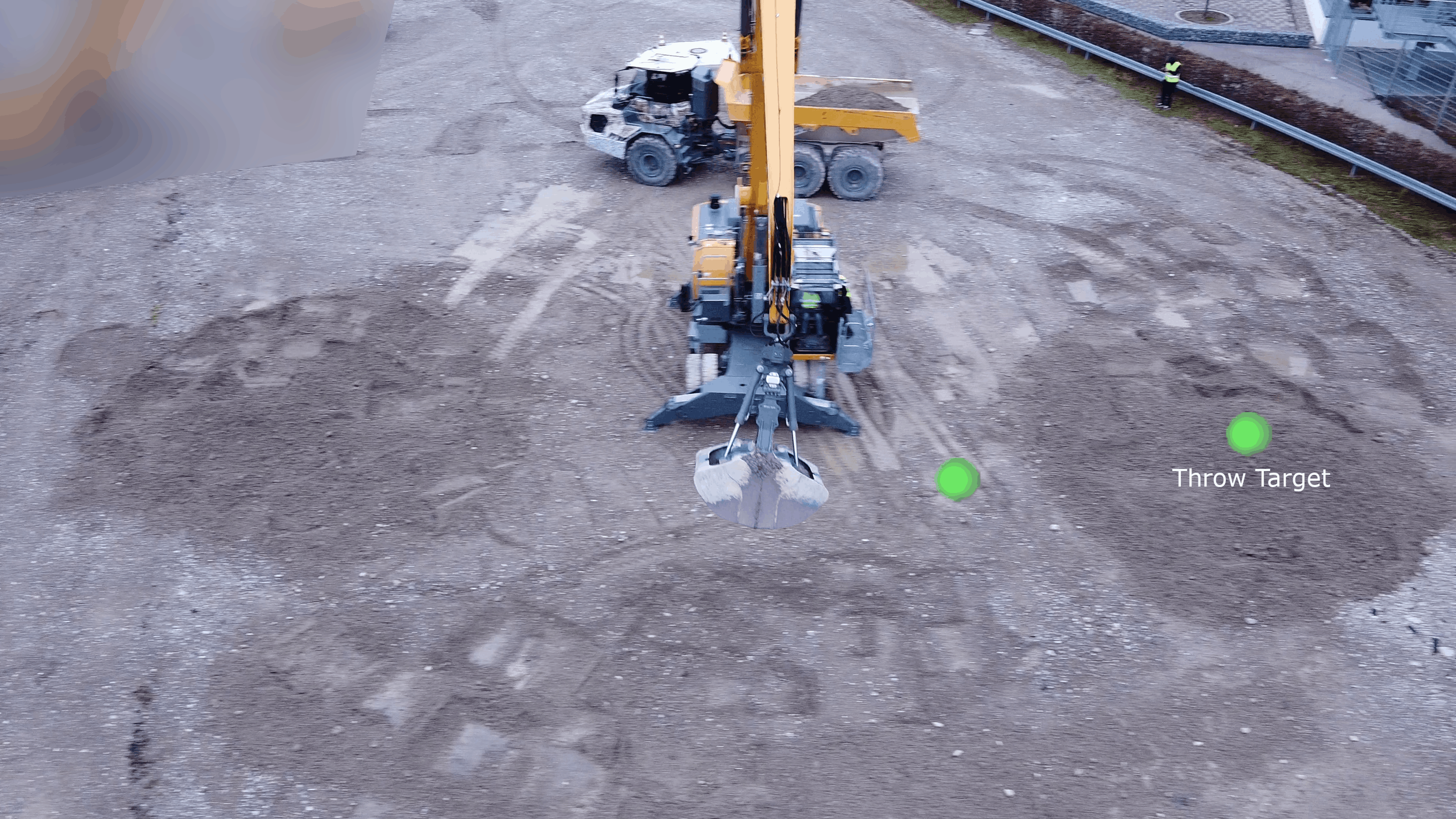}
            \put(5,5){\textcolor{white}{\textbf{C}}}
        \end{overpic}
    \end{minipage}
    % \hspace{-0.6\columnsep}
    \begin{minipage}{\columnwidth}
        \begin{overpic}[trim={0cm 10cm 0cm 0cm},clip, width=\textwidth]{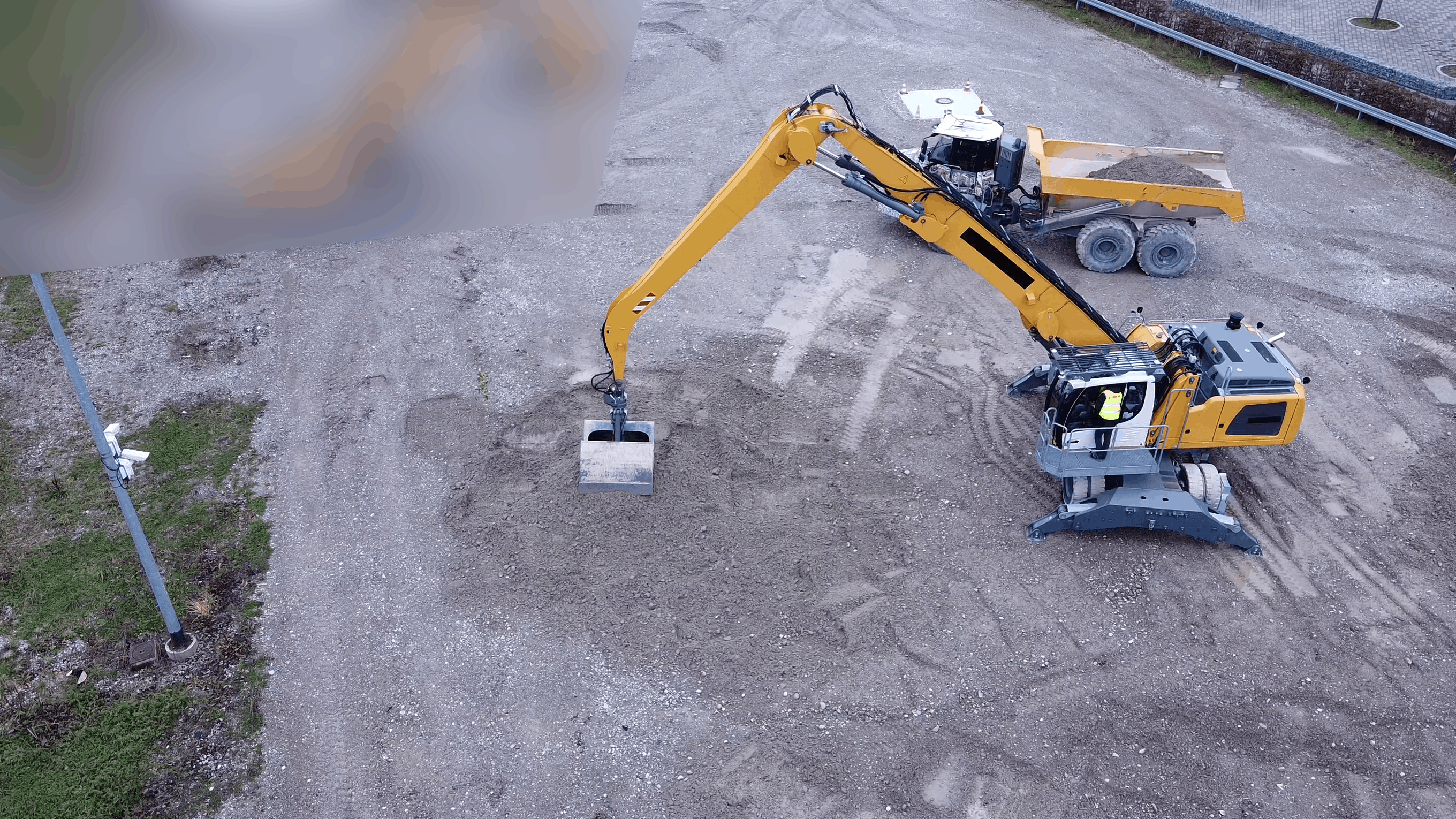}
            \put(5,5){\textcolor{white}{\textbf{B}}}
        \end{overpic}
        \par\vspace{0.5\columnsep}
        \begin{overpic}[trim={0cm 10cm 0cm 0cm},clip, width=\textwidth]{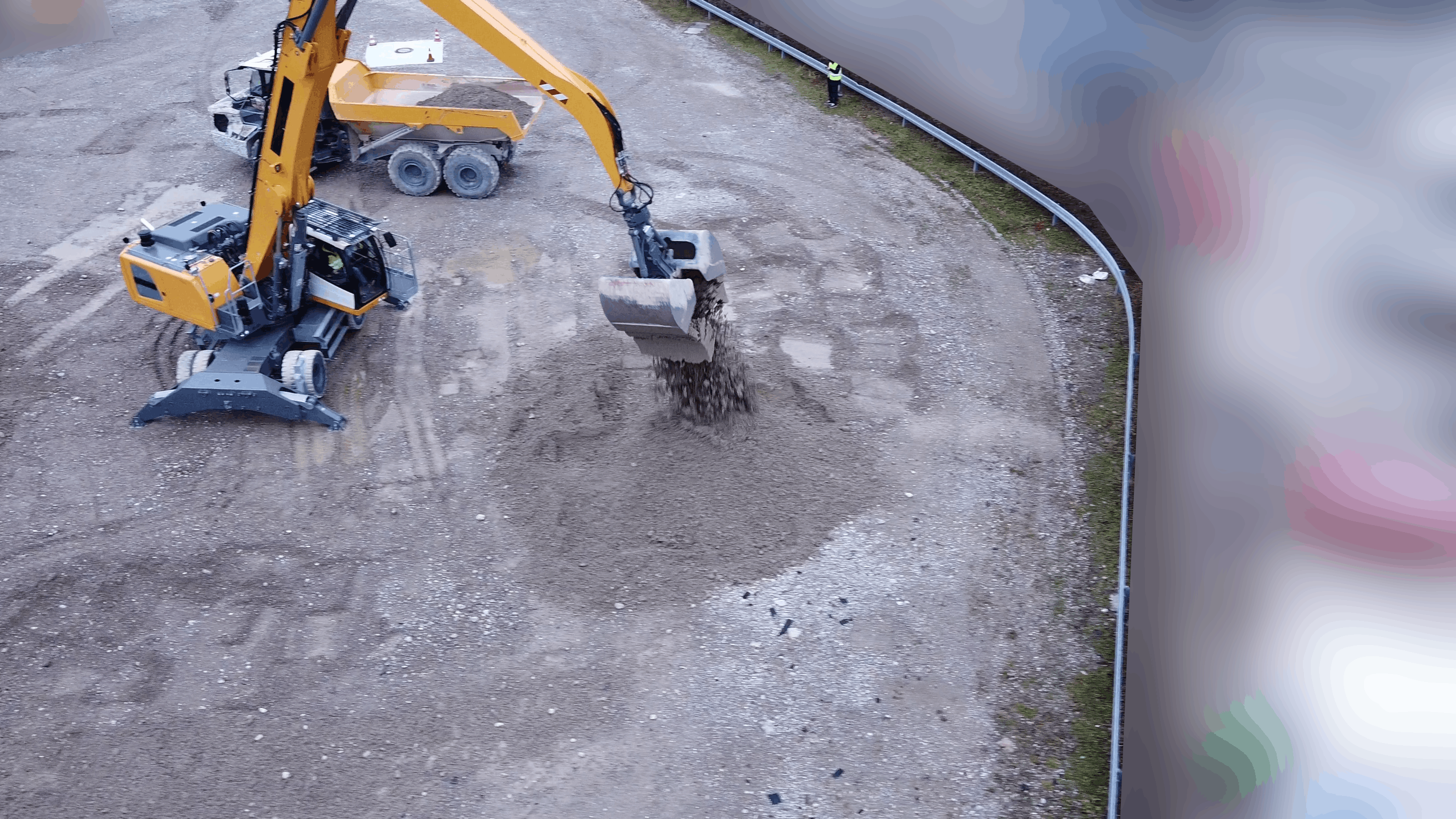}
            \put(5,5){\textcolor{white}{\textbf{D}}}
        \end{overpic}
    \end{minipage}
    \caption{Our test scenario involves a bulk of granular material piled at a given location (to the left of the robot) and a fixed dumping point (to the right). Using our framework, the robot autonomously transfers the material from the original pile to the target location, demonstrating consistent performance over more than 20 consecutive cycles. The task is executed in three phases, with transitions managed internally by a state machine.  (A) The gripper is moved to the planned grasping position by the waypoint-following policy. (B) The grasping routine is executed. (C) The \ac{RL} throwing policy is used to rapidly reach the dumping location and (D) release the material, leveraging passive tool oscillations to complete the throw more efficiently.}
    \label{fig:bulk_material_drone}
\end{figure*}

\begin{table*}
    % \vspace{-0.3cm}
    \begin{center}
         \caption{Our framework is tested for 20 consecutive transitions on the bulk pile transfer task in free space. We use different \ac{RL} policies focusing on either speed or precision. As a benchmark, three drivers with different levels of expertise perform the same task under similar conditions. The results are compared in terms of average slew speed during the transitions, average end-effector error at the final trajectory targets, average tool angular velocity, spread of the material pile, average volume of material moved each cycle, average cycle time, and average cycle time without accounting for the grasping phase. The \textbf{best} and \textit{worst} values for each category are highlighted. As a comparison for human driver control error, we include results from~\cite{Spinelli24ReinforcementLearning}. In that work, however, the performed task had a higher focus on precision than speed and efficiency.}
         \label{tab:bulk_stat}
         \begin{tabular}{@{}cccccccc@{}}
             \toprule
             \textbf{Agent} & \textbf{Speed} [\si{\degree \per \second}] & \textbf{Pos Error} [\si{\meter}] & \textbf{Oscillations} [\si{\degree \per \second}] & \textbf{$\det(\Sigma)$} [\si{\meter^4}]  & \textbf{Avg. Fill Ratio} [\%] & \textbf{Cycle Time} [\si{\second}] & \textbf{Time w/o Grasp} [\si{\second}]  \\ \midrule
             Fast Throw & 14.57 & 0.59 & 23.39 & 2.52 & 67.3 & 32.55 & 20.56 \\
             $2\si{\meter}$ Tube & 12.68 & 0.64 & 20.40 & 2.13 & 66.7 & 33.53 & 25.35 \\
             $1\si{\meter}$ Tube & \textit{10.83} & \textbf{0.48} & \textbf{16.82} & \textbf{1.88} & 68.0 & \textit{41.29} & \textit{31.62} \\ \midrule
             Expert Fast & \textbf{22.51} & - & \textit{38.43} & \textit{3.92} & 66.7 & \textbf{17.44} & \textbf{13.76} \\
             Practiced Fast & 17.12 & - & 29.31 & 2.56 & 73.3 & 24.19 & 18.89 \\
             Novice Fast & 21.37 & - & 33.29 & \textit{5.24} & 65.3 & 21.86 & 16.96 \\ \midrule 
             Expert Slow & 19.65 & - & 31.71 & 2.02 & \textbf{76.0} & 21.62 & 17.29 \\
             Novice Slow & 15.23 & - & 22.73 & 2.37 & \textit{58.0} & 34.09 & 26.38 \\ \midrule
             Expert \cite{Spinelli24ReinforcementLearning} & 11.12 & 0.59 & 20.23 & - & - & - & - \\
             Practiced \cite{Spinelli24ReinforcementLearning} & 11.46 & \textit{1.15} & 21.83 & - & - & - & - \\
             \bottomrule
         \end{tabular}
    \end{center}
    % \vspace{-0.5cm}
\end{table*}

As the first full-scale experiment, we consider the bulk material pile transfer task, where the material handler operates primarily in free space, with user-supplied source pile location and the desired dumping point. We deploy the full pipeline described in~\cref{sec:method_mh_pipeline}: the attack point planner and the path planner optimize the reference trajectory; the \ac{RL} policies control the robot during the grasping and dumping motions; a state machine manages the transitions between phases. This setup enables fully autonomous execution of the material transfer task, as illustrated in~\cref{fig:bulk_material_drone}.

\begin{figure}
    \centering
    \includegraphics[trim={0cm 1cm 2cm 2cm},clip,width=\columnwidth]{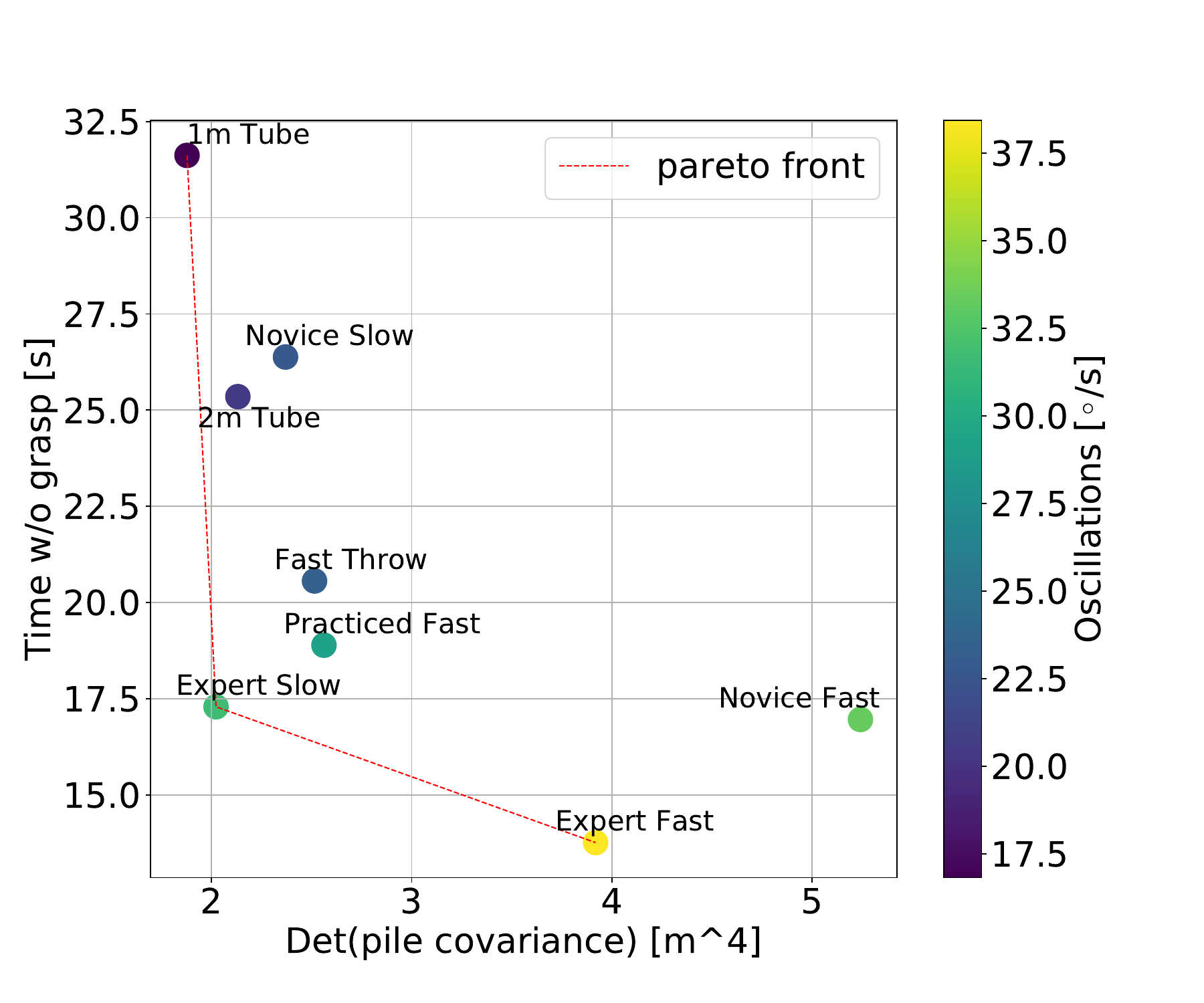}
    \caption{Pareto optimal analysis showing the trade-off between the spread covariance determinant \(\det(\Sigma)\) of the final pile and the cycle time excluding the grasping phase (\textit{Time w/o Grasp}). We rank the eight experiments reported in \cref{tab:bulk_stat}. The RL policies operate close to the Pareto-optimal front and outperform novice human drivers across all metrics. Reduced oscillations, indicated by the color coding, can be prioritized in safety-critical scenarios, albeit at the cost of longer cycle times.}
    \label{fig:pareto_points}
\end{figure}

To evaluate our framework, we perform the same task using different policy combinations: a \texttt{Fast Throw} option combining a \SI{2}{\meter}-tube waypoint-following policy with the \texttt{Multi} throwing policy from \cref{tab:throw_stat} (which does not use waypoint-following to guide the motion approaching the throwing target), and two complete setups employing \texttt{1m Tube} policy and \texttt{2m Tube} policy for both waypoint-following and throwing.
We compare their performances against three human drivers: an \texttt{Expert} operator with over 10 years of experience, a \texttt{Practiced} user trained for material handling machines with 2–3 years of experience, and a \texttt{Novice} newly introduced to material handler operations, with limited experience operating other construction machinery. Given the task’s competing objectives of speed versus precision, human drivers were instructed to prioritize either rapid execution or precise throwing. This setup aims to reveal the natural trade-offs operators make when balancing these metrics. To ensure a fair comparison, all tests are conducted using the same pile and dump locations, with manually assembled bulks of similar size and shape, and under comparable weather conditions. Each test is terminated as soon as the initial pile volume falls below a user-selected threshold where ground contact is still negligible.
Results presented in \cref{tab:bulk_stat} are evaluated along three key dimensions: control authority, material manipulation capability, and operational efficiency. To quantify these, we report the average slew rotational speed during transitions, the final end-effector position error, the mean oscillation amplitude of the tool, the determinant of the pile spread covariance matrix $\Sigma$ at the end of the experiment, the average volume of material grasped per cycle, and the average cycle time for approximately 20 execution cycles.
A visual summary of all experiments is shown in \cref{fig:pareto_points}, highlighting the correlation between cycle time, throw precision, and oscillation minimization.

To analyze the agent's control performance, we focus on two key metrics: the slew rotation speed, which can be increased by accurately learning the hydraulic motor dynamics (modeled by our \ac{NN} in~\cref{sec:sim_dynamics}), and the tool oscillations, which should be minimized for safety. While higher slew speed improves operational efficiency, they also generate increased oscillations which contribute to a more dispersed and uncontrolled pile distribution. These two aspects are therefore inversely related, as illustrated in~\cref{fig:pareto_points}. To ensure safe deployment under varying conditions, the \ac{RL} training setup addresses the trade-off by prioritizing oscillation minimization, which generally results in slower performance compared to human operators. Supporting this observation, we find that more relaxed tube constraints lead to increased rotation speed. The emphasis on safety is a deliberate environment design choice, but can be tuned for different needs by adjusting the reward coefficients in \cref{eq:waypoint_reward}.

Unlike the \ac{RL} controllers, human operators do not explicitly follow waypoints, making it difficult to quantify their tracking accuracy. However, experiments from~\cite{Spinelli24ReinforcementLearning}, which used visually marked end-effector targets, report tracking errors in the range of $[0.5, 1.0]\si{\meter}$, depending on the operator's level of experience. All of our policies achieve errors in the lower half of this range while maintaining comparable slew speeds to the ones recorded for the end-effector tracking task. This demonstrates that the \ac{RL} controllers attain accuracy on par with expert human operators, while significantly outperforming model-based approaches (\cref{tab:mpc_stat}) and the previously proposed solution~\cite{Spinelli24ReinforcementLearning}. The accuracy and safety requirements needed for operating in real construction sites are therefore met.

Granular material manipulation is evaluated based on grasping efficiency and throwing precision. As shown in~\cref{tab:grasp_stat} and~\cref{tab:throw_stat}, the proposed solutions outperform the identified baselines in both aspects. While the \ac{RL}-based planner performs comparably to \texttt{Expert Fast} in terms of average material volume per cycle, the expert achieves higher bucket fills when allowed more time to plan and execute fine-grained maneuvers (\texttt{Expert Slow}). In these cases, they use fine-grained arm adjustments and a slower descent, leveraging compliance to ensure all four gripper corners get in contact with the pile.
In contrast, our \ac{RL} planner (like \texttt{Expert Fast}) lowers the boom directly, often causing one edge to catch, preventing full gripper contact, and leading to incomplete scoops.

Data from~\cref{tab:bulk_stat} show that all policy combinations achieve throwing precision comparable to that of \texttt{Expert Slow}, who defines the Pareto front in~\cref{fig:pareto_points}. By learning and predicting the free-fall dynamics of granular material, the \ac{RL} throwing controller consistently delivers accurate throws across diverse real-world conditions, demonstrating the suitability of \ac{RL} for dynamic manipulation tasks.

Finally, we evaluate the overall task efficiency. To guarantee a focused comparison between the learning-based approach and human drivers, we use the Time w/o Grasp metric, which measures the average cycle time excluding the grasping routine. This allows isolating the impact of the \ac{RL} components from the unoptimized grasping procedure. For completeness, the full cycle time is also reported in \cref{tab:bulk_stat}. By tuning the strictness of the end-effector motion constraints, our method achieves higher throughput than novice users (\texttt{Novice Slow}) and performs on par with intermediately experienced operators (\texttt{Practiced Fast}) while reducing the target pile dispersion. The efficiency of expert users remains the upper bound, owing to their ability to fluidly transition between task phases. When instead optimizing for the material spread, our autonomous pipeline closely approaches even the most experienced users.

\subsection{Dump Truck Loading}\label{sec:truck_exp}

\begin{table}
    % \vspace{-0.3cm}
    \begin{center}
         \caption{The \SI{1}{\meter} tube policy is compared with two drivers for the challenging dump truck loading task. Relevant metrics for this task in constrained space are rotation speed, tool oscillations and cycle time, allowing us identify efficient yet safe operations.}
         \label{tab:truck_stat}
         \begin{tabular}{@{}cccc@{}}
             \toprule
             \textbf{Agent} & \textbf{Speed} [\si{\degree \per \second}] & \textbf{Oscillations} [\si{\degree \per \second}] & \textbf{Time w/o Grasp} [\si{\second}]  \\ \midrule
             $1\si{\meter}$ Tube & \textit{8.18} & 16.20 & 26.88 \\ \midrule
             Expert & \textbf{11.52} & \textit{18.98} & \textbf{21.55} \\
             Novice & 10.42 & \textbf{16.08} & \textit{30.06} \\
             \bottomrule
         \end{tabular}
    \end{center}
    % \vspace{-0.5cm}
\end{table}

\begin{figure*}
    \centering
    \includegraphics[trim={4cm 1cm 3cm 2cm},clip,width=2\columnwidth]{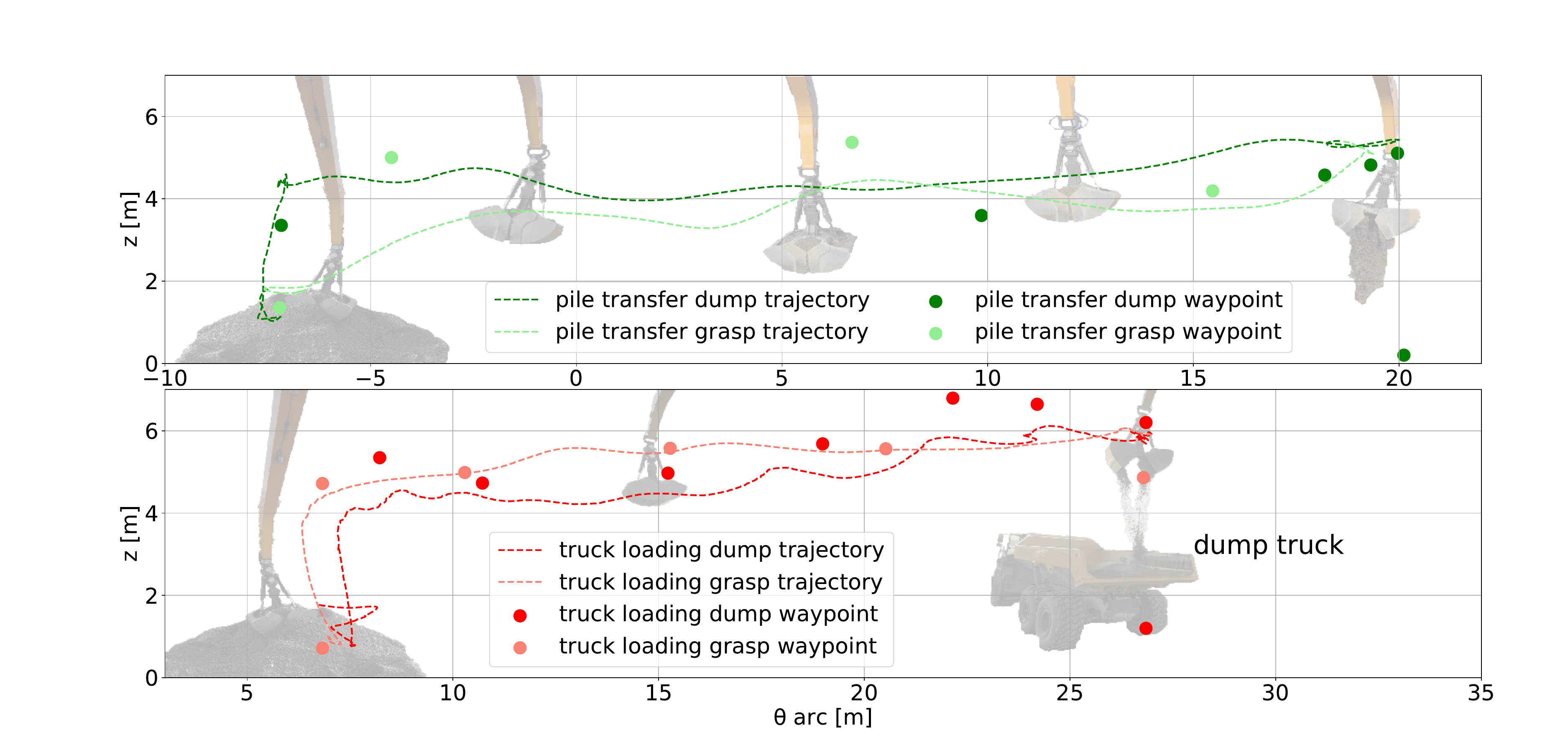}
    \caption{Visualization of two consecutive grasping and dumping trajectories in the bulk material pile transfer task (green, \SI{2}{\meter} tube policy) and the dump truck loading task (red, \SI{1}{\meter} tube policy). In the pile transfer task, a sparser waypoint sampling allows for faster, less constrained motions. The policy is also able to shortcut sharp path segments, for instance, during grasping at $\theta_{\text{arc}} \in [-10, 0]\si{\meter}$. In contrast, the dump truck loading task requires a denser path and a more restrictive tube constraint to ensure safe motion around obstacles. To improve dumping precision, three closely spaced target points are sampled at the end of each dumping trajectory.
    }
    \label{fig:pile_truck}
\end{figure*}

Our second full-pipeline experiment is high-precision dump truck loading, represented in \cref{fig:dump_truck_drone}.
Such a task presents higher complexity than material handling in free space due to the collision risk with the vehicle, and the high precision required to dump the material in the truck bed. For reference, the truck-bed has an approximate dimension of $5.5\si{\meter}\times2.5\si{\meter}$ while the open gripper $2\si{\meter}\times1.5\si{\meter}$.
The pile approach and grasping phase of our framework remain unchanged. By manually specifying the dump location inside the truck bed, collisions between the two machines are inherently handled as the path planner observes the dump truck as an obstacle. To ensure all the material gets correctly dumped into the truck, we adapt the throwing policy at deployment-time to reduce the tool oscillations before opening the gripper.
This is achieved by \textit{i)} policy input modulation by sampling waypoints more densely towards the end of the throwing trajectory (as shown in \cref{fig:pile_truck}), which results in slower and safer tracking facilitating oscillation damping, and \textit{ii)} action gating by suppressing the gripper open action until the tool oscillations fall below \SI{0.1}{\radian \per \second}. 
These adaptations are applied to the safest and most accurate \texttt{1m Tube} throwing policy without retraining. 

The performance metrics of the proposed pipeline are compared with those of two different human drivers in \cref{tab:truck_stat}.
In such a constrained task, human operators are extremely careful: the \texttt{Expert} driver reduces the slew rotation speed by \SI{41}{\percent} compared to free-space operations, the \texttt{Novice} by \SI{32}{\percent}.
Our policy, on the other hand, only reduces the speed by \SI{24}{\percent} as the \ac{RL} controllers are trained with tracking precision and oscillation damping rewards regardless of the tasks. The slowdown of the slew rotation speed is purely due to the adaptations for operating on a less smooth trajectory.
Under these conditions, the efficiency of our autonomous pipeline is closer to human operators, with performance falling between that of an expert and a novice.
\section{Discussion}

This section discusses the performance and limitations of our framework across four key aspects: granular material manipulation, robust and safe control, low-level actuation, and efficiency relative to human operators.

\subsection{Granular Material Manipulation}
Our first contribution is the \ac{RL}-based attack point planner, which outperforms a commonly used heuristic by \SI{41}{\percent} when operating with raw point cloud observations, as shown in \cref{tab:grasp_stat}. While the performance of the highest-point heuristic can be significantly improved by a more precise state estimator~\cite{khattak2025compslam,Jelavic22RoboticPrecision}, our simulated experiments show that even under ideal conditions, such a logic does not necessarily yield the optimal result (\cref{tab:sim_att}). This observation is supported by human operators behavior, who also do not always target the highest point. On the other hand, our planner achieves near human-level performance, integrates seamlessly with LiDAR-based observations, and can be retrained for different task geometries in under an hour. Despite the strong performance of our \ac{RL}-based method, the expert human operator still achieves superior bucket fill ratios. When provided with additional time, they can plan more effectively and execute precise maneuvers to collect more soil. We believe this performance gap could be narrowed by replacing the current scripted grasping behavior with a more accurate \ac{RL}-based grasping controller, and by integrating it into a hierarchical \ac{RL} framework for planner training. Realizing this improvement requires simulating the compliance of the passive gripper joints in response to the soil profile, thereby further bridging the sim-to-real gap.

Another key contribution of this work is a new \ac{RL}-based throwing policy. By incorporating simulated granular material dynamics during training, our policy achieves up to a \SI{39}{\percent} improvement in throwing precision over previous \ac{RL} controllers, as shown in \cref{tab:throw_stat}. As illustrated in \cref{fig:throw_meshes}, the resulting material spread is significantly more compact, enabling more demanding tasks such as precise loading into dump trucks. This level of precision aligns with that of expert human operators (\cref{fig:pareto_points}). However, our real-world results remain less accurate than those observed in simulation. Despite the higher mean error shown in \cref{fig:throw_meshes} compared to \cref{fig:throw_error}, the resulting piles exhibit spread aligned to human baselines, suggesting a systematic bias leading to a consistent offset. We attribute this to the sim-to-real gap, particularly in modeling the material release dynamics and/or the robot's joint dynamics. The initial conditions of the material at release, affected by friction and its heterogeneous distribution, are inherently hard to predict. While our simulation approximates the material as discrete load particles and captures the correlation between tool oscillation and release timing, it falls short in modeling the complex flow dynamics of freely falling granular material from an open gripper. These phenomena could be better captured by a particle-based simulator, albeit with significantly higher computational demands that may compromise the feasibility of model-free \ac{RL}.
Finally, discrepancies in slew joint dynamics and tracking errors of other joints in the low-level controller may also contribute to the observed inaccuracies in the real system.

\subsection{Safe Control of Hydraulic Machines}
\label{sec:discuss_safe_control}

Compared to prior work~\cite{Spinelli24ReinforcementLearning}, we propose a pipeline with clear safety guarantees, crucial for handling free-hanging, underactuated grippers. Our \ac{RL} policies track a safe reference trajectory within bounded deviation, allowing reliable operation when paired with an obstacle-aware planner. However, the introduced tube constraint relies on restrictive assumptions that may limit flexibility. In particular, we observe the following issues:
\begin{enumerate}
    \item The agent's planning horizon is limited to the next three waypoints. With this design, the spacing of these waypoints directly influence the agent’s speed and can be used as a tuning parameter to address different tasks (\cref{fig:pile_truck}). For instance, when the three visible waypoints are closely spaced, the controller tends to move more conservatively to avoid violating tube constraints beyond its limited view, and to overshoot the terminal waypoint. 
    This cautious behavior arises because the agent cannot anticipate upcoming turns or required maneuvers, and is not aware of the final target position. In contrast, providing a longer or full sequence of waypoints would allow the policy to better anticipate the global path and plan more optimally. 
    However, this approach introduces challenges: the expanded observation space complicates training, and requiring a fixed-length waypoint sequence reduces the deployment flexibility achieved with the receding-horizon approach.

    \item The tube radius, which constrains the allowable deviation from the planned path, is fixed across the entire trajectory and conservatively chosen based on the most constrained segment (e.g., near obstacles). This design limits the policy’s ability to adapt its behavior to varying path difficulty. In tight areas, a small tube ensures safety, but the same constraint applies in open space, where greater flexibility could enable faster and more efficient motion. As a result, the agent often reduces joint velocities to remain within bounds and suppress oscillations, leading to overly cautious behavior even when more aggressive motion would be safe. Our ablation study in \cref{sec:rl_way_throw_sim} lays a clear path for improvements: the agent learns to correlate the observed safety margin with the need for more or less cautious behavior. Therefore, new unified policies could be trained to operate with an adaptive tube radius. However, integrating such policies into the current framework would necessitate a more sophisticated path planner that supplies both waypoints and segment-specific error margins. Such an extension is considered beyond the scope of this work.
\end{enumerate}

In conclusion, given these two limitations, our tube-constrained \ac{RL} policies still lag behind human operators in terms of global planning and adaptive decision-making. Having a complete knowledge of the task specifics, human drivers can commit to more informed long term decisions. Furthermore, aware of the differently dangerous portions of the environment, they can command considerably faster motions and neglect tool stabilization when not essential for safety.

\subsection{Low Level Joint Control}
\label{sec:discuss_low_level}

\begin{figure}
    \centering
    \includegraphics[trim={2cm 0cm 3cm 1cm},clip,width=\columnwidth]{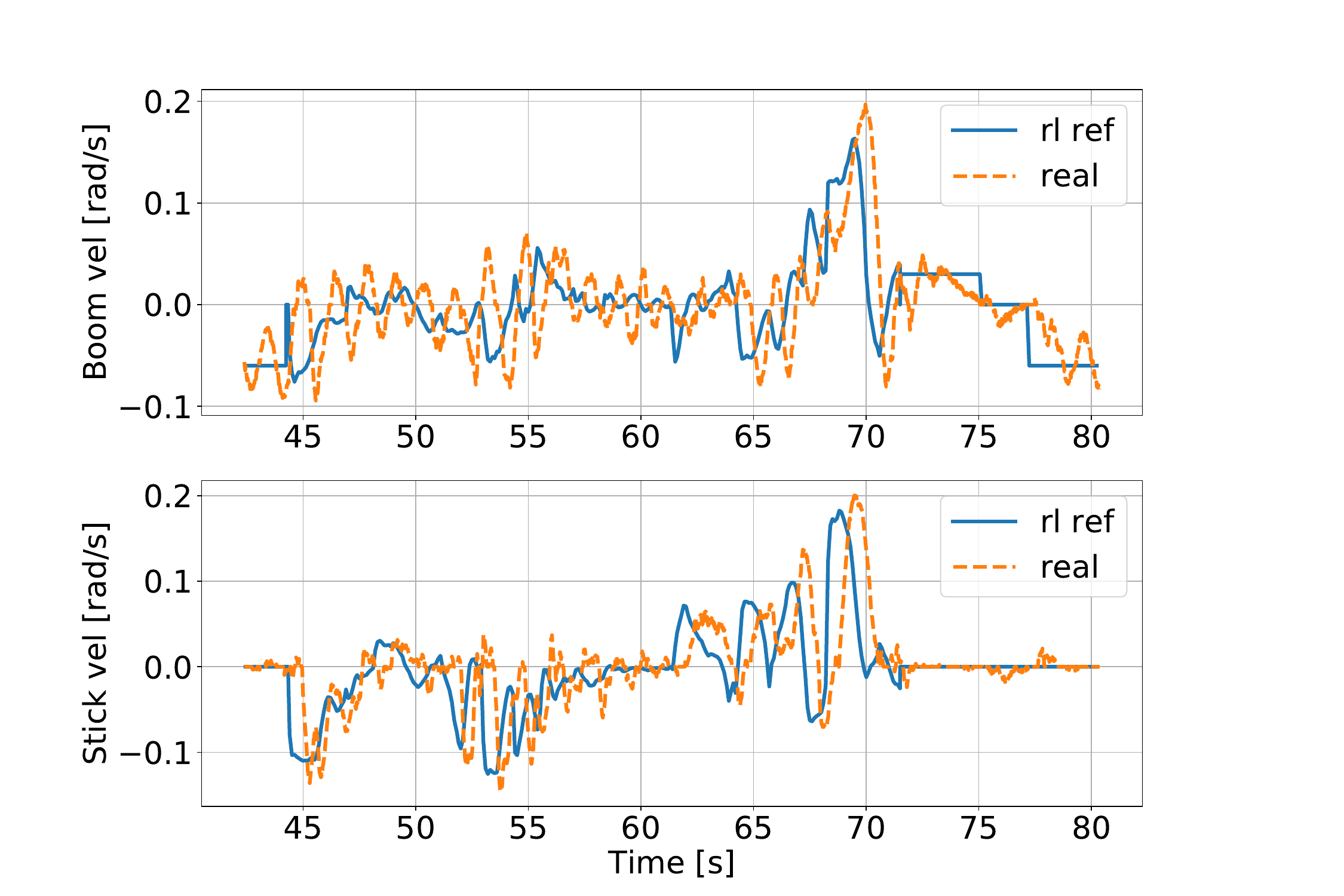}
    \caption{Low-level velocity controller tracking performance for boom and stick joint, in the dump truck loading task. During the first half of the trajectory (dumping phase, until \SI{\sim 60}{\second}), both joints move up (negative velocity) to avoid collision with the truck, and in the second half (grasping phase, until \SI{\sim 70}{\second}), the gripper is lowered to approach the pile. From $70\si{\second}$ to $80\si{\second}$, the grasping routine is triggered, providing step velocity references for the boom joint only.}
    \label{fig:vel_control}
\end{figure}

The low-level control approach for tracking boom and stick velocity references builds on prior works~\cite{Egli22SoilAdaptiveExcavation, Jud21RoboticEmbankment}, which primarily target slow and steady joint motions.

Unlike the \acp{NN} that capture the full dynamics of the slew joint, the \ac{LUT}-based controller relies solely on a feedback PI component to manage transients and unmodeled dynamics. This method achieves high accuracy primarily for simple hydraulic cylinders moving along slow trajectories with minimal inertial effects. In material handling tasks, these conditions typically apply to the arm joints, allowing for a hybrid control strategy rather than training a \ac{NN} modeling the entire machine as in~\cite{Egli22GeneralApproach}, which demands more extensive, joint-specific data collection.

Nevertheless, the low-level joint velocity controllers demonstrate unreliable performance. As shown in \cref{fig:vel_control}, they induce oscillations that negatively impact the \ac{RL} policy’s closed-loop stability in subsequent steps. We attribute this instability mainly to the feedforward control’s inability to account for transient dynamics, particularly during high-load operations on the boom joint. Importantly, the \ac{RL} policies, trained with domain randomization and observation noise, can compensate for these limitations to some extent and still provide effective end-effector control.

\subsection{Comparison with Human Operators}
The proposed \ac{RL} solution demonstrates a clear improvement over model-based approaches (\cref{tab:mpc_stat}), offering fast and accurate motions while maintaining safe gripper configurations. Its behavior aligns with that of less experienced human operators. However, highly trained experts still outperform the current system, highlighting an upper bound yet to be reached.
As discussed in \cref{sec:discuss_safe_control}, one key limitation of the \ac{RL} controller is its use of a fixed tube constraint size. This restricts its ability to move quickly, even in scenarios where damping gripper oscillations is not strictly necessary. For instance, in the open-space pile transfer task, where intermediate waypoints are not required for safety, the policy still adheres to the tube constraint, thus compromising efficiency. It is \SI{83}{\percent} slower than \texttt{Expert Slow} when optimizing for accuracy (\texttt{1m Tube}), and \SI{49}{\percent} slower than \texttt{Expert Fast} when prioritizing speed (\texttt{Fast Throw}), as shown in \cref{fig:pareto_points}. This disparity stems from the human operators’ strategic choice to allow more gripper oscillation, tolerating oscillations \SI{64}{\percent} to \SI{89}{\percent} larger than the \ac{RL} agent, in order to maximize throughput. While this results in less precise dumping, as indicated by the pile metrics in \cref{tab:bulk_stat}, high precision is typically not critical for general pile management tasks.

In contrast, the dump truck loading experiment presents a scenario where our framework excels. When collision avoidance and dumping accuracy become primary requirements, the proposed design proves essential for successful execution. As shown in \cref{fig:pile_truck}, the planned safe trajectory is less smooth and must be followed closely to avoid collisions with the truck, which stands approximately \SI{3.5}{\meter} tall. In this context, minimizing oscillations is crucial both for safety and for meeting accuracy demands. The autonomous system performs only \SI{25}{\percent} slower than an expert operator, and notably, \SI{12}{\percent} faster than a novice operator. This validates the effectiveness of our control strategy in safety-critical applications.

Overall, our framework enables large-scale material handling systems to operate autonomously with a level of performance comparable to that of intermediate-skilled human operators, delivering safe, precise, and reliable robotic execution in complex real-world scenarios.
\section{Conclusion}

In this work, we presented a novel framework for autonomous material handling that leverages reinforcement learning to tackle the complex dynamics of hydraulic systems and the mechanical flow of granular materials. Our approach integrates a perceptive \ac{RL} attack point planner, a sampling-based path planner, and a set of \ac{RL} controllers to achieve safe and efficient manipulation, successfully addressing challenges posed by underactuated kinematics and uncertainties in hydraulic actuation.

The autonomous material handler demonstrates robust performance in selecting optimal attack points, performing collision-free trajectories in unstructured environments, and executing precise material dumping operations, closely matching and in some cases surpassing the capabilities of human operators.
Moreover, the framework’s modular design enables seamless integration of environmental perception and control, allowing reliable adaptation to diverse and unstructured application scenarios. Extensive experimental validation on a \SI{40}{\tonne} research machine confirms the effectiveness of our approach under real-world conditions, marking a significant step forward in safe and efficient construction automation.

Despite these results, several limitations highlight opportunities to further improve the framework’s robustness and efficiency. First, the low-level controller struggles to track the \ac{RL} policy’s references during fast motions due to limited modeling of hydraulic dynamics; this can be improved by using a controller that estimate and adapts to the full cylinder dynamics online, as proposed by Nan and Hutter~\cite{Nan24LearningAdaptive}. Second, the fixed tube constraint in the waypoint-following policy is unnecessarily restrictive in open-space scenarios; combining perception, planning, and control in an end-to-end fashion~\cite{Hoeller24ANYmalParkour, Miki22LearningRobust} would enable a perceptive \ac{RL} policy to trade off sway damping and motion speed based on the semantic understanding of the workspace. Third, we do not model the full granular material dynamics during grasping and instead use an unoptimized grasping controller; integrating a particle-based simulation, such as~\cite{servin2021multiscale}, would enable replacing the scripted routine with an efficient \ac{RL}-based controller. Installing a gripper yaw joint encoder and incorporating such joint in attack point planning could further also benefit the grasping phase, improve the fill ratio and the overall efficiency. Fourth, the framework currently relies on a manually tuned state machine to coordinate the three separate controllers; imitation learning~\cite{Brohan23RT1Robotics, rudin2025parkour} could be used to distill them into a unified agent capable of inferring transitions and sequencing tasks autonomously. Finally, our current framework is limited to fixed-base settings. Incorporating a base pose optimization that accounts for the pile geometry and dumping location could further enhance system performance and applicability. 

Overall, this study highlights the potential of reinforcement learning to transform conventional material handling systems, paving the way toward more autonomous, resilient, and high-performance heavy machinery for industrial applications.

\section*{APPENDIX}
% Appendixes, if needed, appear before the acknowledgment.
\subsection{MPC Baseline for Arm Control}\label{apx:mpc}  
The \ac{MPC} controller is used to benchmark the proposed \ac{RL} policies on motion control tasks. The goal of this analysis is to identify potential weaknesses inherent to the \ac{MPC} framework, rather than obtaining the best absolute performance. Accordingly, few simplifications have been introduced.
The state vector is defined as:
\begin{equation}
    X = \big[q_{\text{slew}}, q_{\text{boom}}, q_{\text{stick}}, q_{\text{tool},x}, q_{\text{tool},y} \big]^\top,
\end{equation}
with same control inputs  as the \ac{RL} policies (\cref{eq:waypoint_policy}, \cref{eq:throw_policy}):
\begin{equation}
    U = \big[u_{\text{slew}},\dot{q}^{\text{~ref}}_{\text{~boom}},\dot{q}^{\text{~ref}}_{\text{~stick}} \big]^\top.
\end{equation}
The optimization minimizes:
\begin{equation}\label{eq:mpc_cost}
    J = \sum_{k=1}^N \Big( J_{\text{traj}}(X_{[:3]}, \mathbf{q}^{\text{ref}}) + J_{\text{osc}}(X_{[3:]}) + J_{\text{act}}(U) \Big),
\end{equation}
where $\mathbf{q}^{\text{ref}}$ is the \ac{IK} solution for the next waypoint (accounting for active joints only). The horizon $N=5$ with $\Delta t = 0.1\si{\second}$ ensures the same time-length access as the \ac{RL} policies. $J_{\text{traj}}$ promotes waypoint tracking, $J_{\text{osc}}$ penalizes tool oscillations, and $J_{\text{act}}$ regularizes control effort. Even without explicit tube constraints, the optimal trajectory minimizing the cost remains within bounds.  
The dynamics are modeled as
\begin{equation}\label{eq:mpc_model}
    X_{k+1} = X_k + \dot{X} \cdot \Delta t + \tfrac{1}{2} \Ddot{X} \cdot \Delta t^2,
\end{equation}
with
\begin{equation*}
\begin{aligned}
     \dot{X} &= \big[ f_{\text{slew}}(U_{[0]}), U_{[1]}, U_{[2]}, 0, 0 \big]^\top, \\
     \Ddot{X} &= \big[ 0, 0, 0, f_x(X_{[3:]}, U_{[0]}), f_y(X_{[4]}, U_{[:]}) \big]^\top.
\end{aligned}
\end{equation*}
A first-order model of the known dynamics is used, neglecting actuation delays. While usually sufficient, this approximation fails to simulate an accurate slew joint behavior, which however lacks a precise analytical formulation in existing literature.

The \ac{MPC} controller is tested in the simulation environment introduced in \cref{sec:sim_dynamics}. The robot receives a slew joystick input and boom and stick velocity references. The slew dynamics are approximated by a \ac{NN} model (\cref{eq:NN_formulation_vel}), while arm joints follow target velocities through a low-pass filter (\cref{eq:arm_simulated}). The environment parameters are sampled from the training distribution.  
For evaluation, we provide as input a pointcloud recorded on the real machine, including obstacles and noise, and run the full pipeline of \cref{fig:ros2_overview} to compute the attack point and a collision-free path. The \ac{RL} components and the \ac{MPC} controller are interchanged within the same state-machine coordination, operating together with the grasping controller.  
This setup highlights the performance gap between a traditional \ac{MPC} formulation and the proposed \ac{RL} solution, and how well each method integrates within the overall pipeline. 

\section*{ACKNOWLEDGMENT}
We extend our gratitude to Leon Erzberger for developing the first iteration of the waypoint policy and to Jonas El-Zehairy for investigating and evaluating alternative options for the path planner. 
Special thanks to Patrick Goegler and Stephan Brockes for their invaluable support in the prototype platform development. 
Finally, we would like to acknowledge Hongyi Yang, Antoine Fontaine, Youssef Ahmed, Nicolas Pfitzer, and Riccardo Maggioni for their contributions in addressing some of the challenges encountered during this project.

\bibliographystyle{bibliography/IEEEtran}
\bibliography{bibliography/references_new}

\vfill\pagebreak

\end{document}